%% file: main.tex
\documentclass[12pt, oneside]{report}
\usepackage{epsf}
\usepackage{latexsym}
\usepackage{amsmath}
\usepackage{xspace}
\usepackage{amssymb}
\usepackage{amsmath}
\usepackage{multirow}
\usepackage{graphicx}
\usepackage{subfigure}
\usepackage{makeidx}
\usepackage{multirow}
\usepackage{url}
\usepackage{alltt}
\usepackage{setspace}
\usepackage{algpseudocode}
\usepackage{algorithm}
\usepackage{tabularx}
\usepackage[hidelinks]{hyperref}
\usepackage{xcolor}
\usepackage{cleveref}
\usepackage[margin=1in]{geometry}

\newcommand\ud[1]{\texttt{#1$^{\texttt{UD}}$}}
\newcommand\xpos[1]{\texttt{#1$^{\texttt{X}}$}}
\newcommand\upos[1]{\texttt{#1$^{\texttt{U}}$}}
\newcommand\KALMF{{KALM\textsuperscript{FL}}\xspace}
\newcommand\KALMR{{KALM\textsuperscript{RA}}\xspace}
\newcommand\Stanza{{\textsc{Stanza}}\xspace}
\newcommand\MS{{m\textsc{Stanza}}\xspace}
\definecolor{highlight}{RGB}{255,255,0}

\newcounter{property}
\newenvironment{property}{
    \refstepcounter{property}
    \noindent\textbf{Property \theproperty}}

\newcounter{prompt}
\newenvironment{prompt}{
    \refstepcounter{prompt}
    \noindent\textbf{Prompt \theprompt}}

\newcounter{promptinstantiation}
\newenvironment{promptinstantiation}{
    \refstepcounter{promptinstantiation}
    \noindent\textbf{Instantiation of Prompt 3, No. \thepromptinstantiation}}

\newcounter{example}[chapter]
\renewcommand{\theexample}{\thechapter.\arabic{example}}
\newenvironment{example}{
    \refstepcounter{example}
    \noindent\textbf{Example \theexample}}

\newcounter{definition}[chapter]
\renewcommand{\thedefinition}{\thechapter.\arabic{definition}}
\newenvironment{definition}{
    \refstepcounter{definition}
    \noindent\textbf{Definition \thedefinition}}

\begin{document}

\input{0_1_cover/0_main}

{\setstretch{1.3}

\input{0_2_abstract/0_main}

\clearpage
\pagenumbering{arabic}

\input{1_intro/0_main}

\input{2_related/0_main}
\input{3_kalmfl/0_main}

\input{4_kalmra/0_main}
\input{5_disambiguation/0_main}
\input{6_conclusion/0_main}

\newpage
\addcontentsline{toc}{chapter}{Bibliography}
\bibliographystyle{plain}
\chaptermark{Bibliography}
\bibliography{main}

\input{7_appendix/0_main}

}

\end{document}

%% file: 0_1_cover/0_main.tex
\input{0_1_cover/1_cover}
\input{0_1_cover/2_committee}
\input{0_1_cover/3_abstract_title}

%% file: 0_1_cover/1_cover.tex
\pagenumbering{gobble}
\title{\bf{Knowledge Authoring with Factual English, Rules, and Actions}}
\vspace*{3\baselineskip}
\centerline{\bf{Knowledge Authoring with Factual English, Rules, and Actions}}
\vspace*{1\baselineskip}
\centerline{A Dissertation presented}
\vspace*{1\baselineskip}
\centerline{by}
\vspace*{1\baselineskip}
\centerline{\bf{Yuheng Wang}}
\vspace*{1\baselineskip}
\centerline{to}
\vspace*{1\baselineskip}
\centerline{The Graduate School}
\vspace*{1\baselineskip}
\centerline{in Partial Fulfillment of the}
\vspace*{1\baselineskip}
\centerline{Requirements}
\vspace*{1\baselineskip}
\centerline{for the Degree of}
\vspace*{1\baselineskip}
\centerline{\bf{Doctor of Philosophy}}
\vspace*{1\baselineskip}
\centerline{in}
\vspace*{1\baselineskip}
\centerline{\bf{Computer Science}}
\vspace*{1\baselineskip}
\centerline{Stony Brook University}
\vspace*{2\baselineskip}
\centerline{\bf{December 2023}}

%% file: 0_1_cover/2_committee.tex
\newpage
\pagenumbering{roman}
\setcounter{page}{2}
\centerline{\bf{Stony Brook University}}
\vspace*{1\baselineskip}
\centerline{The Graduate School}
\vspace*{2\baselineskip}
\centerline{Yuheng Wang} \vspace*{2\baselineskip}
\centerline{We, the dissertation committee for the above candidate for the}
\vspace*{1\baselineskip}
\centerline{Doctor of Philosophy degree, hereby recommend}
\vspace*{1\baselineskip}
\centerline{acceptance of this dissertation}
\vspace*{1\baselineskip}
\centerline{\bf{Advisor: Michael Kifer}}
\centerline{\bf{Professor, Department of Computer Science, Stony Brook University}}
\vspace*{1\baselineskip}
\centerline{\bf{Co-advisor: Paul Fodor}} 
\centerline{\bf{Associate Professor of Practice,}}
\centerline{\bf{Department of Computer Science, Stony Brook University}}
\vspace*{1\baselineskip}
\centerline{\bf{Committee Chair: Andrew Schwartz}} 
\centerline{\bf{Associate Professor, Department of Computer Science, Stony Brook University}}
\vspace*{1\baselineskip}
\centerline{\bf{Outside Committee Member: David Warren, }} 
\centerline{\bf{Professor Emeritus, Department of Computer Science, Stony Brook University}}
\vspace*{1\baselineskip}
\centerline{\bf{Outside Committee Member: Daniela Inclezan}} 
\centerline{\bf{Associate Professor,}}
\centerline{\bf{Department of Computer Science and Software Engineering, Miami University}}
\vspace*{2\baselineskip}
\centerline{This dissertation is accepted by the Graduate School}
\vspace*{1\baselineskip}
\centerline{Celia Marshik}
\vspace*{1\baselineskip}
\centerline{Dean of the Graduate School}

%% file: 0_1_cover/3_abstract_title.tex
\centerline{Abstract of the Dissertation}
\vspace*{1\baselineskip}
\centerline{\bf{Knowledge Authoring with Factual English, Rules, and Actions}}
\vspace*{1\baselineskip}
\centerline{by}
\vspace*{1\baselineskip}
\centerline{\bf{Yuheng Wang}}
\vspace*{1\baselineskip}
\centerline{\bf{Doctor of Philosophy}}
\vspace*{1\baselineskip}
\centerline{in}
\vspace*{1\baselineskip}
\centerline{\bf{Computer Science}}
\vspace*{1\baselineskip}
\centerline{Stony Brook University}
\vspace*{1\baselineskip}
\centerline{\bf{2023}}
\vspace*{1\baselineskip}

%% file: 0_2_abstract/0_main.tex
\input{0_2_abstract/1_abstract}

\input{0_2_abstract/2_misc}

%% file: 0_2_abstract/1_abstract.tex
Knowledge representation and reasoning (KRR) systems represent knowledge as collections of facts and rules.
Like databases, KRR systems contain information about domains of human activities like industrial enterprises, science, and business.
KRRs can represent complex concepts and relations, and they can query and manipulate information in sophisticated ways.
Unfortunately, the KRR technology has been hindered by the fact that specifying the requisite knowledge requires skills that most domain experts do not have, and professional  knowledge engineers are hard to find.
One solution could be to extract knowledge from English text, and a number of works have attempted to do so (OpenSesame, Google's Sling, etc.).
Unfortunately, at present, extraction of logical facts from unrestricted natural language is still too inaccurate to be used for reasoning, while restricted grammars of the language (so-called controlled natural languages, or CNL) are hard for the users to learn and use.
Nevertheless, some recent CNL-based approaches, such as the Knowledge Authoring Logic Machine (KALM),
have shown to have very high accuracy compared to others, and a natural question is to what extent the CNL restrictions can be lifted.
Besides the CNL restrictions, KALM has limitations in terms of the types of knowledge it can represent.
For example, KALM users cannot author rules to support multi-step reasoning, nor can they author actions associated with occurrences of events, which hinders its ability to do time-related reasoning.
Apart from the aforementioned shortcomings, the system's speed was insufficient to adequately support the overall knowledge authoring process.

To address these issues, we propose an extension of KALM called KALM for Factual Language (\KALMF).
\KALMF uses a neural parser for natural language, \MS, to parse what we call \textit{factual} English sentences, which require little grammar training to use.
Building upon \KALMF, we propose KALM for Rules and Actions (\KALMR), to represent and reason with rules and actions.
Furthermore, we identify the reasons behind the slow speed of KALM and make optimizations to address this issue.
Our evaluation using multiple benchmarks shows that our approaches achieve a high level of correctness on fact and query authoring (95\%) and on rule authoring (100\%). 
When used for authoring and reasoning with actions, our approach achieves more than 99.3\% correctness, demonstrating its effectiveness in enabling more sophisticated knowledge representation and reasoning. 
We also illustrate the logical reasoning capabilities of our approach by drawing attention to the problems faced by the famous AI, ChatGPT.
Finally, the evaluation of the newly proposed speed optimization points not only to a 68\% runtime improvement but also yields better accuracy of the overall system.

%% file: 0_2_abstract/2_misc.tex
\tableofcontents 

\addcontentsline{toc}{chapter}{List of Figures}
\listoffigures

\addcontentsline{toc}{chapter}{List of Tables}
\listoftables

\newpage
\addcontentsline{toc}{chapter}{Acknowledgements}
\centerline{\bf{Acknowledgements}}
\vspace*{4\baselineskip}
I would like to express my sincere gratitude to my co-advisors, Michael Kifer and Paul Fodor, whose expertise, guidance, and unwavering support have been instrumental in the successful completion of this dissertation.
Their insightful feedback and encouragement have shaped my research and contributed significantly to my growth.
My Ph.D. journey was predominantly spent during the COVID-19 pandemic.
During this period, it was challenging to meet with Michael and Paul in person, but our collaborative efforts ensured that our research consistently remained on an efficient track.
If there's any regret during my doctoral studies, it is the limited face-to-face interactions with Michael and Paul.
However, I firmly believe that there will be more opportunities for us to reunite in the future.

My heartfelt appreciation goes to my parents, Mingchang Wang and Hongmei Li, for their unconditional love, encouragement, and sacrifices.
Their unwavering belief in my abilities has been a constant source of inspiration, motivating me to overcome challenges and pursue goals.

I would also like to acknowledge the faculty and staff of Department of Computer Science at Stony Brook University for providing valuable resources and a collaborative research environment.
The interactions with fellow students, researchers, and friends, including but not limited to Xingzhi Guo, Anand Aiyer, and Khiem Phi, have played a crucial role in shaping my research ideas and fostering a sense of community.

This dissertation would not have been possible without the collective support of these individuals and institutions.
I am truly grateful for the opportunities and guidance that have enriched my academic journey.

%% file: 1_intro/0_main.tex
\chapter{Introduction}

\input{1_intro/1_intro}

%% file: 1_intro/1_intro.tex
\section{Motivation}

Much of human knowledge can be captured in knowledge representation and reasoning (KRR) systems that are based on logical facts and rules. Unfortunately, translating human knowledge into the logical form that can be used by KRR systems requires well-trained domain experts who are scarce.

One popular idea is to use natural language (NL) to represent knowledge, but current technology (e.g. OpenSesame~\cite{swayamdipta2017frame}, SLING~\cite{ringgaard2017sling}) for converting such statements into logic has rather low accuracy.
A possible fix to this problem is to author knowledge via sentences in
controlled natural languages (CNLs), such as ACE used in Attempto~\cite{fuchs1996attempto}. These CNLs are fairly rich and algorithms exist for converting CNL sentences into logical facts.
Unfortunately, CNLs are also very restrictive, hard to extend, and require significant training to use.
Furthermore, both CNLs and the more general NLP systems cannot recognize sentences with identical meanings but different syntactic forms. For example, ``\textit{Mary buys a car}'' and ``\textit{Mary makes a purchase of a car}'' would be translated into totally different logical representations by most systems, which renders logical inference mechanisms unreliable at best. This problem is known as \emph{semantic mismatch}~\cite{gao2018high}.
The Knowledge Authoring Logic Machine (KALM) \cite{gao2018knowledge} was then introduced to solve this problem, but KALM was based on Attempto's ACE and therefore inherited all the aforesaid issues with CNLs.
Meanwhile, KALM can only represent a limited range of knowledge, such as facts and queries, leaving other important types of human knowledge, such as rules and actions, unsolved.
Moreover, although KALM adopts many optimizations, some steps are still inefficient.

In this paper, firstly, we propose KALM for Factual (English) Language (\KALMF) to address the problems associated with CNL.
To this end, we transplant the KALM framework to
a neural parser for natural language (NL), \MS, which is a modified \Stanza \cite{qi2020stanza} version with multiple, ranked outputs. 
Of course, to turn English into an authoring tool for KR one still needs to impose some restrictions on the language.
For instance, ``\textit{Go fetch more water}'' is a command that does not convey any factual information that can be recorded in a knowledge base (except, perhaps, those based on rather esoteric logics).
Here we mainly focus on a type of English sentence suitable for expressing facts and queries which are identified as one class of English sentences called \emph{factual sentences}.
These sentences can be translated into logic and the aforesaid semantic mismatch problem is solved for such sentences.
Unlike CNLs, factual sentences need little training as long as users keep focus on knowledge representation rather than fine letters.

Then, to tackle the issues associated with knowledge coverage, we empower users to create rules and actions using factual sentences.
Our approach incorporates F-logic \cite{kifer1989f} to enable multi-step frame-based reasoning with authored rules.
In addition, we use a formalism called Simplified Event Calculus (SEC) \cite{sadri1995variants} to represent and reason about actions and their effects.
We call this extension KALM for Rules and Actions (\KALMR).
In terms of implementation, we found a Prolog-like system is more suitable for frame-based parsing, so we implemented \KALMR in XSB \cite{swift2012xsb}.
However, the knowledge produced by \KALMR contains disjunctive knowledge and function symbols, so we chose the answer set programming system DLV \cite{leone2006dlv} as the reasoner for generated knowledge.\footnote{
    Other ASP logic programming systems, such as Potassco \cite{gebser2019multi}, lack the necessary level of support for function symbols and querying.}

Finally, we identify the so-called \textit{role-filler disambiguation step}, defined in Section \ref{sec:deploy}, as one that requires major speed-up.
As a result, we propose two optimization approaches using representation learning to enhance the overall speed and accuracy of KALM.

Evaluation on multiple benchmarks including MetaQA \cite{zhang2017variational}, the UTI guidelines \cite{shiffman2009writing}, and bAbI Tasks \cite{weston2015towards} shows that our approaches achieve 95\% accuracy on fact and query authoring, 100\% accuracy on rule authoring, and 99.3\% on authoring and reasoning about actions.
We also assess the powerful AI,
ChatGPT\footnote{\url{https://chat.openai.com/chat}},
using bAbI Tasks,
and highlight its limitations with respect to logical reasoning compared to our proposed extensions.
Finally,  we revise and re-evaluate the role-filler disambiguation step, which shows  that our new approaches not only reduce the runtime by more than 68\%, but also give KALM higher accuracy.

\section{Thesis Contributions and Outline}

The contributions of this thesis are as follows:

\begin{enumerate}
    \item A new subset of the English language for knowledge authoring, which we call the \emph{factual} (English) language.
    This language has few grammatical restrictions and can be mastered by users without extensive training.
    This language is designed to be sufficiently expressive for specifying database facts.
    An extension of this language supports rules and actions, which we call \textit{factual English with rules and actions}.
    This contribution is discussed in Sections \ref{sec:factual}.
    
    \item  An enhanced natural language processing toolkit, \MS, that can generate multiple promising dependency parses for grammatical checks.
    This contribution is discussed in Section \ref{sec:mstanza}.
    
    \item Two high-accuracy knowledge authoring machines, namely \KALMF and \KALMR.
    The former achieves a significant reduction in grammatical constraints by applying factual English, which allows users to author factual knowledge and common queries with lower learning costs.
    The latter is built upon the former, enabling users to author rules and actions for complex reasoning using factual English with rules and actions.
    This contribution is discussed in Section~\ref{sec:kalmf} and Chapter~\ref{chap:5:kalmra}.
    
    \item Optimization of the critical \emph{role-filler disambiguation} step of the KALM processing pipeline, which significantly improves both the speed and the accuracy of the entire system.
    Role-filler disambiguation is discussed in Chapter \ref{chap:x:disam}.
\end{enumerate}

The rest of this thesis is organized as follows. 
Chapter \ref{chap:2:relwork} provides the background required to understand this thesis.
My own contribution in this thesis is presented in Chapter \ref{chap:4:kalmfl} through \ref{chap:x:disam}.
Specifically, 
Chapter \ref{chap:4:kalmfl} extends KALM to factual English;
Chapter \ref{chap:5:kalmra} handles the authoring of rules and actions;
Chapter \ref{chap:x:disam} presents solutions to faster and more accurate role-filler disambiguation.
Finally,
Chapter \ref{chap:7:conclusion} concludes the thesis and discusses possible future directions.

%% file: 2_related/0_main.tex
\chapter{Background}\label{chap:2:relwork}
This chapter presents the topics related to knowledge representation and reasoning, linguistic resources, the Attempto Project, knowledge authoring, and natural language processing and generation.
The connection between each topic and the contributions of this thesis is explained.

\input{2_related/1_krr}
\input{2_related/2_resources}
\input{2_related/3_attempto}
\input{2_related/4_kalm}

\input{2_related/5_nlp}

%% file: 2_related/1_krr.tex
\section{Knowledge Representation and Reasoning}

Knowledge representation and reasoning (KRR) deals with the creation and manipulation of knowledge representations and the development of reasoning algorithms to perform tasks such as classification, prediction, diagnosis, planning, and decision-making.
Here, we only focus on four major KRR-related techniques that are related to our research: XSB Prolog, DLV, F-logic, and Event Calculus.

\subsection{XSB Prolog}

XSB \cite{sagonas1994xsb,swift2012xsb} is a powerful Prolog implementation that extends the standard Prolog language with a number of advanced features. 
One of the most significant of these features is tabling, which is a memoization technique that can greatly improve the efficiency of many Prolog programs.
Tabling allows for the efficient memoization of recursive predicates, which can avoid redundant computations and greatly speed up program execution.
In addition to tabling, XSB Prolog also provides a number of built-in database management tools.
This includes support for dynamic predicates, which can be used to add or remove clauses from a predicate at runtime.
XSB Prolog also includes transaction management, which allows groups of database operations to be treated as a single transaction with ACID properties.
Finally, XSB Prolog includes support for multi-database access, which allows multiple databases to be accessed and manipulated within a single program.
The KALM system and its succeeding extension \KALMF (detailed in Section \ref{sec:kalm} and Chapter \ref{chap:4:kalmfl}, respectively) use XSB as their frame-based parser and database engine.

\subsection{Disjunctive Logic Programming}

DLV (\textbf{D}ata\textbf{L}og with Disjunction, where the logical disjunction symbol \textbf{V} is used) \cite{leone2006dlv} is a disjunctive version of Datalog that operates under the ASP paradigm.
It extends Datalog by adding support for disjunction in facts and rule heads, thus providing greater expressiveness for disjunctive information than KRR systems based on the well-founded semantics (e.g., XSB \cite{swift2012xsb}).
Furthermore, DLV's support for function symbols and querying makes it more convenient for working with semantic frames \cite{fillmore2006frame} than other ASP systems, such as Potassco \cite{gebser2019multi}.

In DLV, rules have the following form:
\begin{verbatim}
h1 v ... v hn :- b1, ..., bm.
\end{verbatim}
\noindent
\texttt{h1} to \texttt{hn} represent explicitly negated literals, where $\texttt{n} > 0$ (i.e. there must be at least one of those).
The \texttt{:-} sign is the transcription of an implication arrow and \texttt{b1, ..., bm} represent general literals, where $\texttt{m} >= 0$ must hold (i.e. the body may be omitted completely).
The part before the \texttt{:-} sign is referred to as head and the part after \texttt{:-} as body.
Note that disjunction symbols may occur in the head only and negation-as-failure symbols in the body only.
Also note that facts can be viewed as special forms of rules, in which the body is empty.
The neutral element of conjunction is true, so facts must always be true.
Disjunctive facts as a special case are also allowed. For example, when expressing the indefinite knowledge ``an apple is either edible or foul," we have the DLV fact as follows:
\begin{verbatim}
edible(apple) v foul(apple).
\end{verbatim}

In the implementation of one of the KALM system extensions, \KALMR (presented in Chapter \ref{chap:5:kalmra}), DLV is employed
as database and inference engine for storing and reasoning about both definite and indefinite knowledge.

\subsection{Frame Reasoning}

F-logic \cite{flogic-95,kifer1989f} is a knowledge representation and ontology language that combines the benefits of conceptual modeling with object-oriented and frame-based languages.
One of its key features is the ability to use \textit{composite frames} to reduce long conjunctions of roles into more compact forms, matching ideally the structure of FrameBase's frames \cite{rouces2015framebase}.

In this thesis, we depart from the actual syntax of F-logic as it is not supported by the DLV system.
Instead, we implemented a small subset of that logic by casting it directly into the already supported DLV syntax.
Therefore, multi-step reasoning and question answering can be achieved in our frame-based knowledge authoring machines, such as \KALMR, which is presented in Chapter \ref{chap:5:kalmra}.
For example, using F-logic, FrameNet frames can be used to answer the question ``\textit{What does Mary buy?}" given the fact ``\textit{Mary buys a car for Bob}," whose frame-based representations, shown below, are not logically equivalent (the fact has more roles than the query).
The variable \texttt{What} is correctly instantiated as \texttt{car}. 

\begin{verbatim}
frame("Commerce_buy",[role("Buyer","Mary"),role("Goods",car),
                      role("Recipient","Bob")]).
?- frame("Commerce_buy",[role("Buyer","Mary"),
                         role("Goods",What)]).
What=car.
\end{verbatim}
\noindent
where \texttt{frame(FrameName,RoleList)} represents a frame and \texttt{role(RoleName, RoleFiller)} is an instance of a role, including the role's name, and the filler word that plays the role.

\subsection{The Event Calculus}
\label{sec:ec}

The event calculus (EC) \cite{kowalski1989logic} is a well-established representation methodology and background theory for time-related reasoning.
It consists of a set of logical axioms that describe the \textit{law of inertia} for actions. This law states that time-dependent facts, \textit{fluents}, that are not explicitly changed by an action preserve their true/false status in the state produced by that action.
The original EC had a very complex vocabulary and was lacking some axioms, which made it difficult to use.
To address these issues, various variants of EC were introduced, including the simplified event calculus (SEC) \cite{sadri1987three,sadri1995variants}, the ``new'' event calculus (NEC) \cite{sadri1995variants}, discrete event calculus (DEC) \cite{mueller2004event}, and the simplified discrete event calculus (SDEC) \cite{katzouris2015incremental}.
Among these variants, SEC stands out as being equivalent to the original EC when time ranges over integer domains, yet it is much simpler and more tractable.
Previous works \cite{katzouris2015incremental,mitra2016addressing,wu2018learning,arias2022modeling} have demonstrated the applicability of SEC to problems involving sequential events or actions in time-related scenarios.
Therefore, this thesis focuses on SEC and its integration into knowledge authoring.
This integration, discussed in Chapter \ref{chap:5:kalmra}, leverages the benefits of SEC to enhance the understanding of actions, which enables high-accuracy reasoning and question answering in various application domains.

A fluent in SEC is said to hold at a particular timestamp if it is initiated by an earlier action and not subsequently terminated prior to the timestamp.
This is formalized by these DLV rules:

\begin{verbatim}
holdsAt(F,T2) :-
    happensAt(A,T1),initiates(A,F), timestamp(T2), T1 < T2,
    not stoppedIn(T1,F,T2).
stoppedIn(T1,F,T2) :-
    happensAt(A,T), terminates(A,F),
    timestamp(T1), T1 < T, timestamp(T2), T < T2.
\end{verbatim}

\noindent
Here \texttt{happensAt/2} represents a momentary occurrence of action \texttt{A} at a timestamp.
If an action is exogenous insertion of a fluent $f$ at time $t$ then we represent it as \texttt{happensAt(f,t)}, i.e., we use the same predicate.
Example \ref{exmp:ec-fact} demonstrates the use of \texttt{happensAt/2}.

\begin{example}
\label{exmp:ec-fact}
The actions
\begin{itemize}
    \item [A\textsubscript{1}.] \textit{Mary goes to the bedroom.}
    \item [A\textsubscript{2}.] \textit{The bedroom is north of the garden.}
\end{itemize}
have the following frame-based representations:

\begin{verbatim}
happensAt(frame("Travel",[role("Person","Mary"),
                          role("Place",bedroom)]),1).
happensAt(frame("North_of",[role("Entity1",bedroom),
                            role("Entity2",garden)]),2).
person("Mary"). place(bedroom).
entity(bedroom). entity(garden). timestamp(1..2).
\end{verbatim}
\end{example}

\noindent
The first \texttt{happensAt/2} introduces an action of traveling from place to place while the second \texttt{happensAt/2} uses an observed (i.e., exogenously inserted) fluent \texttt{"North\_of"(bedroom,garden)}.
Observable fluents are supposed to be disjoint from action fluents, and we will use a special predicate to recognize them in SEC rules.
Timestamps indicate the temporal relation between the action and the observed fluent.
Predicates \texttt{person/1}, \texttt{place/2}, \texttt{entity/2}, define the domains of various roles, while \texttt{timestamp/1} defines the domain of timestamps.

The predicates \texttt{initiates(Action,Fluent)} and \texttt{terminates(Action, Fluent)} in SEC specify domain-specific axioms that capture the initiation and termination of fluents.
In the scenario of Example \ref{exmp:ec-fact}, the predicates \texttt{initiates/2} and \texttt{terminates/2} are as follows.

\begin{example}
\label{exmp:ec-iandt}
The action ``a person goes to a place" initiates the fluent ``the person is located in the place", and the observed fluent ``an entity is north of another" initiates the fluent ``the other entity is south of the entity"; the action ``a person goes to a place" terminates the fluent ``the person is located in another place", which are represented as follows.
\begin{verbatim}
initiates(frame("Travel",[role("Person",Person),
                          role("Place",Place)]),
          frame("Located",[role("Entity",Person),
                           role("Place",Place)]) :-
    person(Person), place(Place).
initiates(frame("North_of",[role("Entity1",Entity),
                            role("Entity2",Entity2)]),
          frame("North_of",[role("Entity1",Entity),
                            role("Entity2",Entity2)]) :-
    entity(Entity), entity(Entity2).
initiates(frame("North_of",[role("Entity1",Entity),
                            role("Entity2",Entity2)]),
          frame("South_of",[role("Entity1",Entity2),
                            role("Entity2",Entity)]) :-
    entity(Entity), entity(Entity2).
terminates(frame("Travel",[role("Person",Person),
                           role("Place",Place)]),
           frame("Located",[role("Entity",Person),
                            role("Place",Place2)]) :-
    person(Person), place(Place), place(Place2).
\end{verbatim}
\end{example}

Combining the SEC axioms with the actions in Example \ref{exmp:ec-fact},and domain-specific axioms in Example \ref{exmp:ec-iandt}, we can derive the following fluents (assuming the highest timestamp is 3):

\begin{itemize}
    \item [F\textsubscript{1}.] \textit{Mary is located in the bedroom at timestamp 2.}
    \item [F\textsubscript{2}.] \textit{Mary is located in the bedroom at timestamp 3.}
    \item [F\textsubscript{3}.] \textit{The bedroom is north of the garden at timestamp 3.}
    \item [F\textsubscript{4}.] \textit{The garden is south of the bedroom at timestamp 3.}
\end{itemize}
\noindent
which can be further represented as the following DLV facts:
\begin{verbatim}
holdsAt(frame("Located",[role("Person","Mary"),
                         role("Place",bedroom)]),2).
holdsAt(frame("Located",[role("Person","Mary"),
                         role("Place",bedroom)]),3).
holdsAt(frame("North_of",[role("Entity1",bedroom),
                          role("Entity2",garden)]),3).
holdsAt(frame("South_of",[role("Entity1",garden),
                          role("Entity2",bedroom)]),3).
\end{verbatim}
\noindent
and visualized as follows:

\begin{figure}[htbp!]
    \centering
    \includegraphics[scale=0.24]{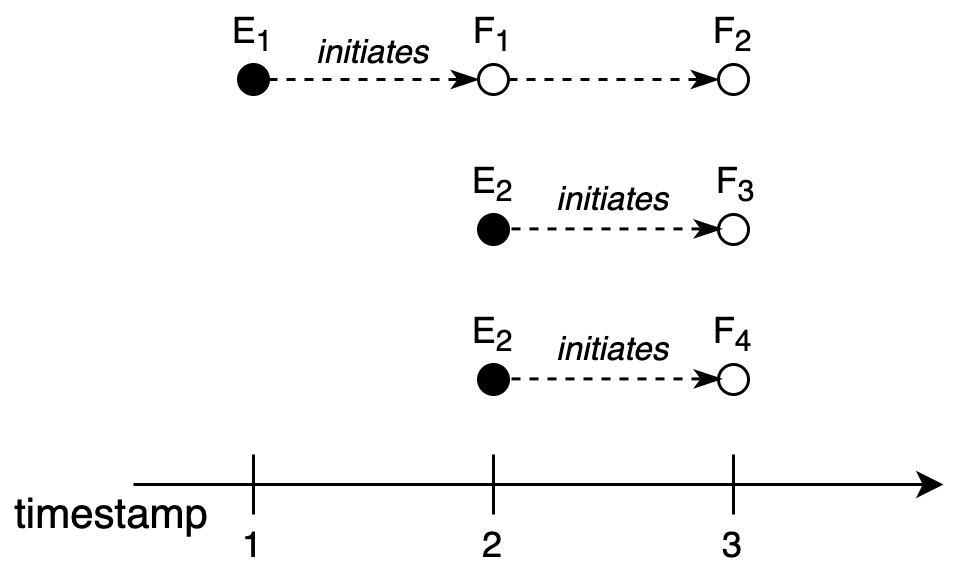}
    \caption{Visualization for SEC-based reasoning}
    \label{fig:ec-flow}
\end{figure}

%% file: 2_related/2_resources.tex
\section{Linguistic Resources}

This section explores two vital linguistic resources, FrameNet and BabelNet, which are extensively employed in the field of KRR to comprehend languages.
Through examples, we illustrate how FrameNet and BabelNet function.

\subsection{FrameNet}

FrameNet\footnote{\url{https://framenet.icsi.berkeley.edu/fndrupal/}} \cite{baker1998berkeley,baker2003framenet} is a semantic framework that aims to capture the meaning of a sentence by identifying the semantic frame(s) that it belongs to and the semantic roles that the words in the sentence play within that frame.

A semantic frame is a conceptual structure that represents a particular scenario or situation, while the semantic roles are the different roles played by entities and events in that scenario.
For example, the semantic frame \texttt{Commerce\_buy} in FrameNet describes a situation in which an individual purchases a good.
This frame includes various semantic roles, such as the \texttt{Buyer}, the \texttt{Goods}, the \texttt{Money}, the \texttt{Seller}, the \texttt{Recipient}, the \texttt{Place}, and the \texttt{Time}.
Each semantic frame is associated with a list of lexical units (LUs) that can trigger the application of that frame.
To account for the different ways in which LUs can be used to express a sentence, FrameNet defines a set of valence patterns for each LU-frame pair.
A valence pattern represents the syntactic context in which a particular LUs and some of the frame roles are used in a sentence.
For example, a valence pattern for the pair \texttt{buy.verb} and \texttt{Commerce\_buy} might specify that the \texttt{Buyer} role is the subject of the \texttt{buy} event and the \texttt{Goods} role is the object of that event.

By applying the appropriate LUs and valence pattern to a sentence, FrameNet can extract the relevant semantic frame and assign the appropriate semantic roles to the words in the sentence. For example,

\begin{example}
\label{exmp:framenet}
FrameNet uses the \texttt{Commerce\_buy} frame to represent the sentence ``\textit{Mary buys a car for Bob for 30,00 dollars.}" as follows:
\begin{center}
\colorbox{red!40}{Mary} \colorbox{black!15}{buys} a \colorbox{green!30}{car} for \colorbox{yellow!60}{Bob} for 
\colorbox{blue!30}{30,000 dollars}.\\
\textcolor{red!50}{\textbf{Buyer}} \xspace
\textcolor{black!25}{\textbf{LU}} \xspace\xspace
\textcolor{green!50}{\textbf{Goods}}
\textcolor{yellow!70}{\textbf{Recipient}} \xspace\xspace\xspace\xspace
\textcolor{blue!40}{\textbf{Money}}\xspace\xspace\xspace\xspace\xspace\xspace\xspace\xspace
\end{center}
\end{example}
\noindent
where the verb ``\textit{buys}" as the LU triggers the \texttt{Commerce\_buy} frame; ``\textit{Mary}" plays the role of \texttt{Buyer}; ``\textit{car}" plays the role of \texttt{Goods}; ``\textit{bob}" plays the role of \texttt{Recipient}; ``\textit{30,000 dollars}" plays the role of \texttt{Money}.

This representation allows for a more nuanced and detailed understanding of the meaning of a sentence and can be useful in a variety of NLP tasks, such as text classification, sentiment analysis, machine translation, and knowledge authoring which we focus on in this thesis. Further details will be provided in Section \ref{sec:kalm}.

\subsection{BabelNet}

BabelNet\footnote{\url{https://babelnet.org/}} \cite{navigli2012babelnet,navigli2021ten} is a multilingual lexical network that integrates information from several different lexical resources, including WordNet \cite{miller1995wordnet}, Wikipedia\footnote{\url{https://www.wikipedia.org/}}, OmegaWiki\footnote{\url{http://www.omegawiki.org/}}, etc.
It provides a comprehensive representation of words and concepts in multiple languages, making it an essential resource for natural language processing tasks that involve cross-lingual information retrieval, word-sense disambiguation, and knowledge-based NLP.

In BabelNet, each word has one or more part-of-speech tags and a glossary for its meanings.
Words with similar meanings are grouped into \textit{synsets} that have unique identifiers (of the form \texttt{bn:dddddddp}, where \texttt{d} is a digit and \texttt{p} is a part of speech symbol like \texttt{v}, \texttt{n}, etc.).
For Example \ref{exmp:framenet}, the word ``\textit{car}" in BabelNet has multiple meanings, which, shown below, belong to different synsets:
\begin{itemize}
    \item \texttt{bn:00007309n}: A motor vehicle with four wheels; usually propelled by an internal combustion engine;
    \item \texttt{bn:00015785n}: A wheeled vehicle adapted to the rails of railroad;
    \item \texttt{bn:00015786n}: The compartment that is suspended from an airship and that carries personnel and the cargo and the power plant;
    \item and more...
\end{itemize}

The BabelNet synset nodes are connected by directed edges representing their semantic relations (i.e., hypernym, hyponym, etc.).
Compared to WordNet, BabelNet has a much larger vocabulary and richer semantic relations.
Specifically, there is an abundance of \textit{named entities}, like famous people, locations, songs, books, etc.
Besides, each edge in the knowledge graph is assigned a weight (in versions earlier than 3.7) that represents the degree of relevance between two connected synset nodes with respect to the type of edge.
Edge weights are a very important feature in BabelNet and are critical to disambiguating role-fillers and eliminating inappropriate candidate parses in the Knowledge Authoring Logic Machine (KALM) \cite{gao2018knowledge,gao2018high,gao2019querying}, which will be introduced in Section \ref{sec:kalm}.
Furthermore, in the discussions in Chapter \ref{chap:x:disam}, we will analyze the advantages and disadvantages of BabelNet-based disambiguation and thus optimize the disambiguation process.

%% file: 2_related/3_attempto.tex
\section{The Attempto Project}
The Attempto Project\footnote{\url{http://attempto.ifi.uzh.ch/site/}}, a research initiative of the University of Zurich, seeks to develop Attempto Controlled English and its accompanying tools.
With the goal of proposing a more effective method for converting human language into structured knowledge, the project has made significant contributions to the field.
These include the development of Attempto Controlled English, the Attempto Parsing Engine, and a Discourse Representation Structure specifically designed for Attempto.

\subsection{Attempto Controlled English}
Attempto Controlled English (ACE) was designed to serve as a knowledge representation language and its usage requires a competent understanding of its construction and interpretation rules.
The vocabulary of Attempto Controlled English consists of built-in function words, (e.g. determiners, articles, pronouns, and quantifiers), built-in phrases (e.g. ``\textit{it is false that ...,}" ``\textit{it is not provable that ...,}") and 100,000 content words, including adverbs, adjectives, nouns, verbs, and prepositions.
Content words are stored in an external lexicon and written in Prolog predicates.
The names of the predicates describe the parts-of-speech of the words. Each predicate has two arguments, between which the first one is the word that the predicate stands for, and the second specifies the base form of the first argument.
For example, the words ``\textit{fast}", ``\textit{faster}", and ``\textit{fastest}" are represented as \texttt{adv(fast,fast)}, \texttt{adv\_comp(faster,fast)}, and \texttt{adv\_sup(fastest,fast)}, respectively, where \texttt{adv} means ``\textit{fast}" is an adverb, \texttt{adv\_comp} means \textit{faster} is a comparative adverb, and \texttt{adv\_sup} means \textit{fastest} is a superlative adverb. All three words use the base form, \textit{fast}.

The \textit{construction rules} of ACE define the format of words, phrases, sentences, and texts:

\begin{itemize}
    \item \textbf{Words:} Built-in function words and built-in phrases are not allowed to be modified by users. Users can create compound words by concatenating two or more content words with hyphens.
    \item \textbf{Phrases:} Phrases include noun phrases, modifying nouns, modifying noun phrases, verb phrases, and modifying verb phrases. Noun phrases in ACE form a subset of noun phrases of regular English plus this includes arithmetic expressions, sets, and lists. Modifying nouns and noun phrases are those that are preceded or followed by adjectives, relative clauses, or possessives. Verb phrases in ACE form a subset of verb phrases in English with specific definitions for negation and modalities. Modifying verb phrases are those that are accompanied by adverbs or preposition phrases.
    \item \textbf{Sentences:} Sentences include declarative, interrogative, and imperative sentences. Declarative sentences include simple sentences, there is/are-sentences, Boolean formulae, and composite sentences. Interrogative sentences include yes/no queries, wh-queries, and how much/many-queries that end with a question mark. Imperative sentences are commands that end with an exclamation mark.
    \item \textbf{Texts:} Texts are sequences of declarative, interrogative, and imperative sentences. 
\end{itemize}

The \textit{interpretation rules} of ACE define how a grammatically correct ACE sentence is translated. Here are three examples of such interpretation rules:

\begin{itemize}
  \item Prepositional phrases modify the verb not the noun: ``\textit{Mary \{enters a card with a code\}."}
  \item Relative clauses modify the immediately preceding noun: ``\textit{Mary enters \{a card that has a code\}."}
  \item Anaphora is resolved using the nearest antecedent noun that agrees in gender and number: ``\textit{Mary was born in Stony-Brook. It's a beautiful place."}
\end{itemize}

\noindent
The first sentence can be interpreted in two ways in English,  ``\textit{A customer uses a code to enter a card.}" and ``\textit{Mary enters a card. The card has a code.}"
In ACE, it is interpreted only in the first way.
In the second sentence, given a relative clause, an ACE sentence refers only to the nearest noun that precedes it.
Hence, the relative clause, ``\textit{that has a code}", modifies the noun, ``\textit{card}". 
In the third sentence, ACE resolves the pronoun, ``\textit{it}", in the second sentence by looking for the nearest antecedent noun that agrees with gender and number, so ACE associates ``\textit{it}" with the word ``\textit{Stony-Brook}".

\subsection{Discourse Representation Structure for Attempto Controlled English}\label{sec:drs}

To parse ACE texts, Attempto Parsing Engine\footnote{\url{https://github.com/Attempto/APE}} (APE) was later implemented.
APE accepts ACE texts as input and produces both paraphrases and Discourse Representation Structures (DRS) in the form of first-order logic terms.
These paraphrases provide the user with a better understanding of how APE interprets ACE sentences.
The concept of DRS was originally introduced in~\cite{kamp1993discourse} and was later redesigned for ACE in~\cite{fuchs2005extended}.
The redesigned DRS, referred to simply as DRS, consists of two parts: a set of discourse \textit{referents} (a.k.a. \textit{identifiers}), called the universe of the DRS, and a set of DRS \textit{conditions}.

A referent in DRS can represent an object introduced by a noun or a predicate introduced by a verb, among others.
A condition in DRS can be either \textit{simple} or \textit{complex}.
A simple condition consists of a logical atom followed by an index, while a complex condition is constructed from other DRS connected by logical operators such as negation, disjunction, and implication.
For each simple condition, the logical atom is limited to one of the 8 predicates, with definitions provided as follows:

\begin{itemize}
    \item \texttt{object(Ref,Noun,Class,Unit,Op,Count)}. The \texttt{object} terms represent objects that are introduced by the various forms of nouns.
    \texttt{Ref} stands for such an object and is used for references. 
    \texttt{Noun} denotes the object that was used to introduce the \texttt{object} functor.
    \texttt{Class} is one of \texttt{\{dom,mass,countable\}} and shows the classification of the noun.
    \texttt{Unit} denotes the measurement noun if the object is introduced together with a measurement noun such as ``\textit{kg}" in ``\textit{2 kg of apples}".
    \texttt{Op} is one of \texttt{\{eq,geq,greater,leq,less,exactly,} \texttt{na\}}, representing quantitative relationships such as ``equal", ``greater or equal", and ``less or equal".
    Together with \texttt{Unit} and \texttt{Op}, \texttt{Count} defines the cardinality or extent.

    \item \texttt{predicate(Ref,Verb,SubjRef)}, \texttt{predicate(Ref,Verb,SubjRef,Obj Ref)}, \\ and \texttt{predicate(Ref,Verb,SubjRef,ObjRef,IndObjRef)} represent relations that are introduced by intransitive, transitive, and ditransitive verbs, respectively.
    \texttt{Ref} stands for such a relation and is used to attach modifiers (i.e. adverbs and prepositional phrases). \texttt{Verb} stands for intransitive, transitive, or ditransitive.
    \texttt{SubjRef} is a reference to the subject, \texttt{ObjRef} points to the direct object, and \texttt{IndObjRef} points to the indirect object in a sentence.

    \item \texttt{property(Ref1,Adjective,Degree)}, \texttt{property(Ref1,Adjective,De gree,Ref2)}, and \texttt{property(Ref1,Adjective,Ref2,Degree,CompTarg et,Ref3)} stand for properties that are introduced by adjectives. \texttt{Ref1} refers to the primary object of the property (i.e. the subject) and \texttt{Ref2} refers to the secondary object of the property.
    \texttt{Ref3} refers to the tertiary object of the property.
    The \texttt{Adjective} argument is intransitive or transitive, representing the type of the adjective.
    \texttt{Degree} is one of \texttt{\{pos,pos\_as,comp,comp\_than,sup\}}; it defines the degree of the adjective.
    Positive and comparative forms can have an additional comparison target (e.g. ``\textit{as rich as ...}", ``\textit{richer than ...}"), and for those cases \texttt{pos\_as} and \texttt{comp\_than} are used.
    \texttt{CompTarget} is one of \texttt{\{subj,obj\}} and defines for transitive adjectives whether the comparison targets the subject (e.g. ``\textit{Mary is more fond-of John than Bob}") or the object (e.g. ``\textit{Mary is more fond-of John than of Bob}").

    \item \texttt{modifier\_pp(Ref1,Preposition,Ref2)}.
    \texttt{Ref1} refers to the modified verb.
    \texttt{Preposition} is the preposition of the prepositional phrase.
    \texttt{Ref2} refers to the object of the prepositional phrase.

    \item \texttt{modifier\_adv(Ref,Adverb,Degree)}.
    \texttt{Ref} refers to the modified verb.
    \texttt{Adverb} is the adverb.
    \texttt{Degree} is one of \texttt{\{pos,comp,sup\}} and defines the adverb's degree.

    \item \texttt{relation(Ref1,of,Ref2)}.
    \texttt{Ref1} refers to the left hand side object of the relation and it is always associated with an \texttt{object}-term. \texttt{Ref2} stands for the right hand side object. The second argument is always \texttt{of} since no other prepositions can attach to nouns.
    
    \item \texttt{has\_part(GroupRef,MemberRef)}.
    \texttt{GroupRef} refers to a group of objects.
    \texttt{MemberRef} stands for the object that is a member of the group.
    
    \item \texttt{query(Ref,QuestionWord)}.
    \texttt{Ref} refers to the object or relation of the query.\\
    \texttt{QuestionWord} is one of \texttt{\{who,what,which,how,where,when\}}.
\end{itemize}

\begin{example}
The ACE sentence ``\textit{Mary buys a beautiful car.}" is parsed by APE and generates the DRS shown below
(note that the indices following the predicates show the location of the input from which the condition is introduced):
\begin{verbatim}
object(A,"Mary",countable,na,eq,1)-1/1
object(B,car,countable,na,eq,1)-1/5
property(B,beautiful,pos)-1/4
predicate(C,buy,A,B)-1/2
\end{verbatim}
\end{example}

\begin{figure}[htbp!]
    \centering
    \includegraphics[scale=0.2]{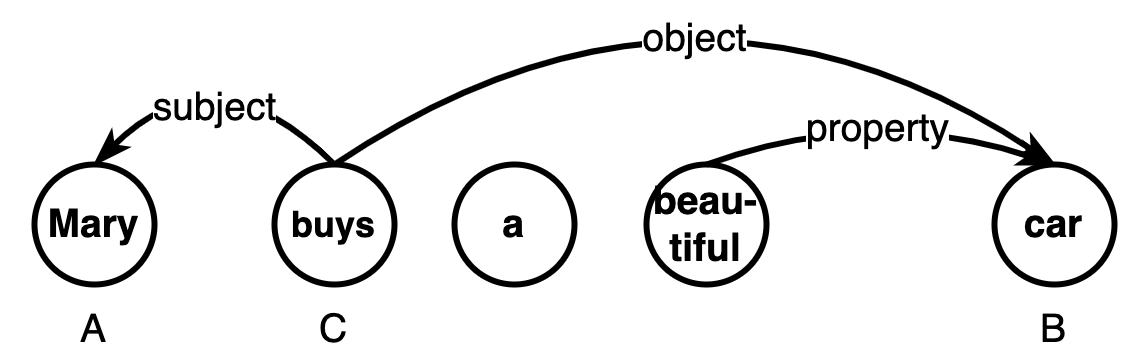}
    \caption{Underlying dependency information in the DRS of ``\textit{Mary buys a beautiful car}"}
    \label{fig:drs-graph2}
\end{figure}

\noindent
By utilizing this DRS, it is possible to extract the word dependencies that underlie the text, as visualized in Figure~\ref{fig:drs-graph2}.
The predicate \texttt{predicate(C,buy, A,B)} indicates that the verb ``\textit{buy}" (referred to as \texttt{C}) links the subject ``\textit{Mary}" (referred to as \texttt{A}) and the object ``\textit{car}" (referred to as \texttt{B}).
Moreover, the object ``\textit{car}" (referred to as \texttt{B}) connects to its adjective ``\textit{beautiful}" since the predicate \texttt{property(B,beautiful,pos)} has \texttt{B} as its 1st argument.

From this example, it can be concluded that to extract the subject of an ACE sentence, one should select the 3rd argument of the \texttt{predicate} term and verify whether this argument refers to an \texttt{object} term.
Likewise, to extract the adjective of the object in an ACE sentence, one should select the 4th argument of the \texttt{predicate} predicate and check if this argument refers to a \texttt{property} predicate.
Generally, every component in a sentence has its own specific method of extraction, and the DRS-implied dependency graphs, such as the one shown in Figure~\ref{fig:drs-graph2}, serve as the foundation for such extraction.

\subsection{Semantic Mismatch}

ACE was developed as a potential way to acquire knowledge, with the expectation that its unambiguous nature could improve the precision of knowledge representation compared to full English.
However, ACE's reliance on the shallow semantics of natural languages makes it challenging to identify sentences with the same meaning but different syntactic structures.
These structures are eventually parsed into different DRS by APE.
The issue is known as \textit{semantic mismatch} \cite{gao2018knowledge, wang2022knowledge}.
Here is an example to illustrate this problem.

\begin{example}
Three synonymous ACE sentences ``\textit{Mary buys a car. Mary is the purchaser of a car. Mary makes a purchase of a car.}" have the following different DRS:
\begin{verbatim}
object(A,"Mary",countable,na,eq,1)-1/1
object(B,car,countable,na,eq,1)-1/4
predicate(C,buy,A,B)-1/2
object(A,"Mary",countable,na,eq,1)-2/1
object(B,purchaser,countable,na,eq,1)-2/4
object(C,car,countable,na,eq,1)-2/7
relation(B,of,C)-2/5
predicate(D,be,A,B)-2/2
object(A,"Mary",countable,na,eq,1)-3/1
object(B,purchase,countable,na,eq,1)-3/4
object(C,car,countable,na,eq,1)-3/7
relation(B,of,C)-3/5
predicate(D,make,A,B)-3/2
\end{verbatim}
\end{example}
\noindent
where all the sentences are translated into different DRS, which does not reveal that all three sentences convey the same meaning.
This occurs because ACE lacks any inherent background knowledge and relies on explicit rules to encode such knowledge.
Unfortunately, constructing these rules is a labor-intensive task that requires the expertise of skilled knowledge engineers.
To address the issue of semantic mismatch, the Knowledge Authoring Logic Machine (KALM) \cite{gao2018knowledge} was introduced, which will be elaborated in Section~\ref{sec:kalm}.

%% file: 2_related/4_kalm.tex
\section{KALM: Knowledge Authoring Logic Machine}\label{sec:kalm}
This section provides a detailed introduction to the Knowledge Authoring Logic Machine (KALM), which forms the foundation of this thesis.
All the approaches proposed in this thesis aim to expand the capabilities of KALM.

\subsection{The KALM Architecture}

\begin{figure}[htbp!]
     \centering
     \subfigure[Training stage]{\includegraphics[width=130mm]{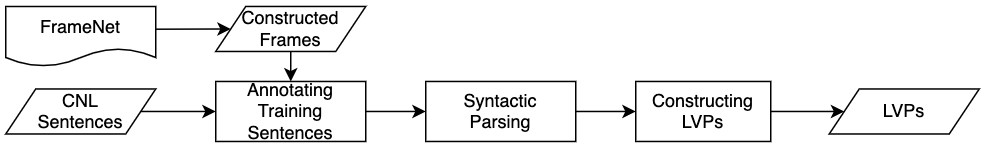}\label{fig:kalm-train}}
     \subfigure[Deployment stage]{\includegraphics[width=135mm]{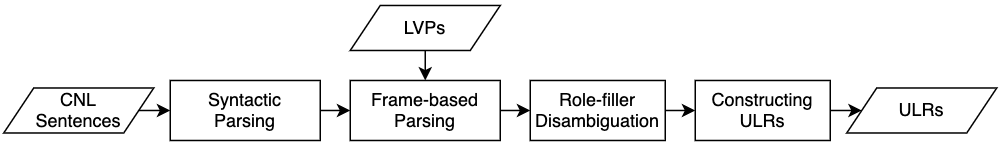}\label{fig:kalm-parse}}
     \caption{The frameworks of the KALM system}
     \label{fig:kalm}
\end{figure}

KALM \cite{gao2018high} is a semantic framework for scalable knowledge authoring.
KALM users author knowledge using CNL sentences (Attempto's ACE, to be specific) and 
KALM ensures that semantically equivalent sentences have identical logical representations through the use of the frame semantics~\cite{fillmore2006frame}.
KALM is a two-stage system following the \textit{structured machine learning} paradigm.
In the first stage, known as the \textit{training stage}, KALM constructs \textit{logical valence patterns} (LVPs) by learning from \textit{training sentences}.
An LVP is a specification that tells how to extract \textit{role fillers} for the concepts represented by the English sentences related to that LVP.
In the second stage, known as the \textit{deployment stage}, the system does frame-based parsing by applying the constructed LVPs to convert CNL sentences into \textit{unique logical representations} (ULRs).
Figure~\ref{fig:kalm} depicts the two stages of KALM.
In terms of engineering implementation, except for the parts related to BabelNet which used Java, all other parts were built with XSB Prolog.

\subsection{Training Stage}
In Figure~\ref{fig:kalm-train}, the training stage of KALM is depicted.
This stage consists of several steps, each of which is explained below.

\subsubsection{Annotating Training Sentences}
To enable semantic understanding of a domain of discourse,
knowledge engineers must first construct the required background knowledge in the form of KALM frames. The overall structure of most of these frames can be adopted from FrameNet and converted into the logic form required by KALM.
Then, knowledge engineers compose training sentences in CNL and annotate them using KALM frames.
For example, the annotated training sentence~(\ref{code:training}), below, indicates that 
the meaning of ``\textit{Mary buys a car for Bob}" is captured by the \texttt{Commerce\_buy} frame;
the LU that triggers this frame is the 2nd word ``\textit{buy}" or its synonym ``\textit{purchase}"; and,
the 1st, the 4th, and the 6th words, i.e., ``\textit{Mary}", ``\textit{car}", and ``\textit{Bob}", play the roles of \texttt{Buyer}, \texttt{Goods}, and \texttt{Recipient} in the frame, respectively.
Additionally, \texttt{required} and \texttt{optnl} say whether this role is required in this frame or is optional.
\begin{equation}
\label{code:training}
\hspace*{-35mm}
\begin{aligned}
\verb|train("Mary buys a car f|&\verb|or Bob","Commerce_buy",|\\
\verb|"LU"=2,[purchase],|\\
\verb|["Buyer"=1+require|&\verb|d,"Goods"=4+required,|\\
\verb|"Recipient"=6+opt|&\verb|nl]).|
\end{aligned}
\end{equation}

\subsubsection{Syntactic Parsing}
\begin{figure}[htbp!]
    \centering
    \includegraphics[scale=0.2]{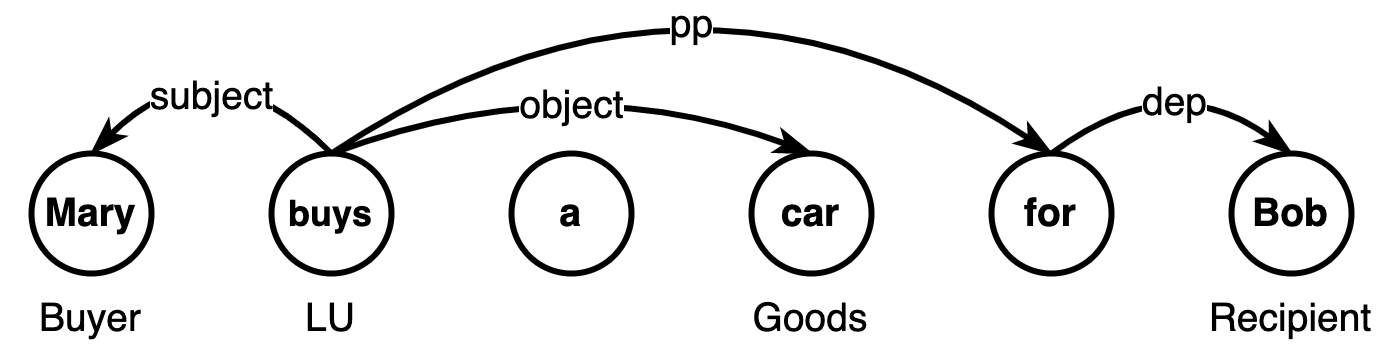}
    \caption{Dependency graph for DRS (\ref{code:drs})}
    \label{fig:drs-graph}
\end{figure}

KALM uses APE to extract the syntactical information from CNL sentences, including POS for each word and the grammatical dependency relations between pairs of words. All extracted information is represented by DRS. Here is the DRS of the training sentence (\ref{code:training}).
\begin{equation}
\label{code:drs}
\hspace*{-52mm}
\begin{aligned}
\verb|object(A,mary,uncountable|&\verb|,na,eq,1)-1/1.|\\
\verb|object(B,car,countable,na|&\verb|,eq,1)-1/4.|\\
\verb|object(C,bob,uncountable,|&\verb|na,eq,1)-1/6.|\\
\verb|predicate(D,buy,A,B)-1/2.|\\
\verb|modifier_pp(E,for,C)-1/5.|
\end{aligned}
\end{equation}

\noindent
where \texttt{A}, \texttt{B}, and \texttt{C} are identifiers for the ``\textit{Mary}", ``\textit{car}", and ``\textit{Bob}", respectively, \texttt{D} is the \textit{buy}-event, and \texttt{E} is the prepositional relation connective ``\textit{for}".
To provide a clearer illustration of the syntactic dependency relations embedded in DRS (\ref{code:drs}), we present a dependency graph in Figure~\ref{fig:drs-graph} that is based on this DRS.
As displayed in Figure~\ref{fig:drs-graph}, every word is depicted as a graph node, while each relationship between them is denoted by an edge that carries a specific edge type.
As an illustration, consider the words ``\textit{Mary}" and ``\textit{buys}" in the sentence.
They are linked by a ``subject" edge in DRS (\ref{code:drs}) because, according to fact \texttt{1/2} of the DRS, the identifier \texttt{A} representing \texttt{mary} appears in the 3rd position of the predicate \texttt{predicate/4}.
This indicates that \texttt{mary} (referred to as \texttt{D}) is the subject of the predicate \texttt{buy}.

\subsubsection{Construction of LVPs}
DRS, along with annotations of CNL sentences, allow KALM to construct LVPs that specify how to fill the roles of a frame triggered by an LU.
As illustrated in Figure~\ref{fig:drs-graph}, by synthesizing the information in the training sentence (\ref{code:training}) and the DRS (\ref{code:drs}), KALM learns that, 
starting from the LU \texttt{buy.verb} (i.e., \texttt{predicate(C,buy,A,B)}), the \texttt{Buyer} ``\textit{Mary}" (i.e., \texttt{object(A,mary, uncountable,na,eq,1)}) can be found by locating the subject (i.e., the 3rd argument in \texttt{predicate/4}) of \texttt{buy.verb}.
As a result, the pattern \texttt{verb->subject} for finding the filler word (i.e., \textit{role-filler}) of \texttt{Buyer} is learned.
Similarly, \texttt{verb-> object} to find \texttt{Goods} can be learned.
Indicators \texttt{required} and \texttt{optnl} are also used in LVPs to decide whether the pattern is required in order to trigger the LVP or is optional.
The learned knowledge about role-filling is formalized as an LVP (\ref{code:lvp-1}) as follows:
\begin{equation}\label{code:lvp-1}
\hspace*{-6mm}
\begin{aligned}
\verb|lvp(buy,"Commerce_buy",|\\
\verb|[pattern("Buyer","v|&\verb|erb->subject",required),|\\
\verb|pattern("Goods","v|&\verb|erb->object",required)]).|\\
\verb|pattern("Recipient|&\verb|","verb->pp[for]->dep",optnl)]).|
\end{aligned}
\end{equation}

\subsection{Deployment Stage}\label{sec:deploy}
The deployment stage of KALM, as depicted in Figure~\ref{fig:kalm-parse}, takes new sentences and generates corresponding logical representations.
The steps in this stage are explained in detail below.

\subsubsection{Frame-based Parsing}
When an unseen CNL sentence comes in, KALM triggers all possible LVPs using the words in the sentence.
Then, the triggered LVPs are applied to this sentence to extract \textit{role-fillers} and thus a \textit{frame parse} of this sentence is generated.
A typical frame parse consists of a frame name, which identifies the meaning of a clause within the sentence, along with a set of named \emph{roles} that specify the key words in the clause and their semantic functions there.
A frame parse identifies the actual words in the clause that ``play'' these roles, i.e., the \emph{role-fillers}. Logically, each role name corresponds to a predicate that represents a particular clauses-to-role-filers relationship.
To illustrate an entire frame-based parsing step, we use the sentence ``\textit{A customer purchases a watch for a friend}" as an example, which has the following DRS generated by APE:

\begin{verbatim}
object(A,customer,countable,na,eq,1)-1/1.
object(B,watch,countable,na,eq,1)-1/4.
object(C,friend,countable,na,eq,1)-1/6.
predicate(D,buy,A,B)-1/2.
modifier_pp(E,for,C)-1/5.
\end{verbatim}

\noindent
The word \texttt{buy.verb} triggers the LVP (\ref{code:lvp-1}).
Following the pattern \texttt{verb-> subject} that extracts the roll-filler of \texttt{Buyer}, KALM starts from the LU \texttt{buy.verb} (i.e., \texttt{predicate(C,buy,A,B)}), and then finds the subject of the LU, which is the 3rd argument of \texttt{predicate(C,buy,A,B)} (i.e., the identifier \texttt{A}).
Finally, the word identified by \texttt{A} (i.e., ``\textit{customer}") is extracted as the role-filler for \texttt{Buyer}.
In this way, KALM applies all patterns and finally generates the following \textit{candidate frame parse} (\textit{candidate parse} for short):
\begin{equation}
\label{code:parse-1}
\hspace*{-39mm}
\begin{aligned}
\verb|p("Commerce_buy",[role("Buyer",|&\verb|customer),|\\
\verb|role("Goods",|&\verb|watch)]).|\\
\verb|role("Recipie|&\verb|nt",friend)]).|
\end{aligned}
\end{equation}

However, in most cases, KALM is able to learn a significant number of LVPs from the training sentences.
For instance, consider the annotated sentence "\textit{Mary buys a car for 5,000 dollars}", as shown below.
\begin{verbatim}
train("Mary buys a car for 5,000 dollars","Commerce_buy",
      "LU"=2,[purchase],
      ["Buyer"=1+required,"Goods"=4+required,
       "Money"=7+optnl]).
\end{verbatim}
\noindent
From this sentence, KALM can learn the following LVP:
\begin{verbatim}
lvp(buy,"Commerce_buy",
    [pattern("Buyer","verb->subject",required),
     pattern("Goods","verb->object",required)]).
     pattern("Money","verb->pp[for]->dep",optnl)]).
\end{verbatim}
\noindent
which parses the test sentence ``\textit{A customer purchases a watch for a friend}" as the following candidate parse different from that of (\ref{code:parse-1}):
\begin{equation}
\label{code:parse-2}
\hspace*{-47mm}
\begin{aligned}
\verb|p("Commerce_buy",[role("Buyer",|&\verb|customer),|\\
\verb|role("Goods",|&\verb|watch)]).|\\
\verb|role("Money",|&\verb|friend)]).|
\end{aligned}
\end{equation}

\subsubsection{Role-Filler Disambiguation}
As indicated in the frame-based parsing step, the frame parser may choose the wrong role-fillers for the frame's roles and thus produce ambiguous candidate parses, if several alternative LVPs apply to the same phrase during parsing.
To address this issue, the role-filler disambiguation step is performed at the level of word \textit{senses}. This has two main steps, as explained below.

The disambiguation starts with the \textit{best role-filler word-sense computation} stage.
For each role-filler from each candidate parse, this stage computes the best word sense of that role-filler from a list of the available word senses.
The algorithm queries BabelNet for each extracted (role-name, role-filler) pair and retrieves two lists of candidate synsets associated with the role-name and the role-filler respectively.
Then, the algorithm performs a breadth-first search with two heuristic scoring functions (\ref{formula:disamb}) and (\ref{formula:disamb-2}) to evaluate all semantic paths that begin at each candidate role-name synset and end at each corresponding role-filler synset or vice versa.

\begin{equation}
\label{formula:disamb}
score^{R}(rn,rf)=\max_{i \in \{1..c\},j \in \{1..d\}}\Big(\big\{score^{S}\big(synset_i(rn),synset_j(rf)\big)\big\}\Big)
\end{equation}
where the superscript $R$ refers to ``role-filler",
$(rn, rf)$ is an arbitrary extracted (role-name, role-filler) pair,
$c$ and $d$ are the numbers of nodes (i.e. synsets in BabelNet)\footnote{
    We use ``nodes" and ``synsets" interchangeably in the context of BabelNet.
}
of the role-name $rn$ and the role-filler $rf$, respectively.
$synset_i(\cdot)$ returns the $i$-th synset of a word, and
$score^{S}(\cdot)$ returns the semantic score of two synsets, which is computed by the following formula:
\begin{equation}
\label{formula:disamb-2}
score^{S}(s_1,s_l)=\frac{ \sum^{l-1}_{k=1}  \Big(\sqrt{f_n(s_k^{s_1,s_l})} \times f_w(e_{k,k+1}^{s_1,s_l})\Big)  }{ 5^{ \sum^{l-1}_{k=1} f_p(e_{k,k+1}^{s_1,s_l}) } }
\end{equation}
\noindent
where the superscript $S$ refers to ``synset",
$s_1$ and $s_l$ denote nodes from BabelNet,
connected by a shortest path of length $l$.
The symbols
$s_k^{s_1,s_l}$ and $e_{k,k+1}^{s_1,s_l}$ denote the $k$-th node and edge, respectively, on the shortest path that begins at $s_1$ and ends at $s_l$. The function
$f_{n}(\cdot)$ returns the number of adjacent nodes in BabelNet given a node as an argument, the function
$f_w(\cdot)$ returns the weight of an edge, and
$f_p(\cdot)$ returns the penalty value for an edge, defined based on the edge type.

With these two formulas, the aforesaid semantic paths can be ranked, and the path with the highest $score^{S}$ is then selected as $score^{R}$.
In this manner, each role-filler is disambiguated to a specific word-sense in the context of the role and the frame that the role-filler corresponds to.

Next, the \textit{best candidate parse computation} stage begins.
For each candidate parse, this stage computes a semantic score to determine which candidate parse gives the optimal role-filler extraction result and, in this way, eliminates unwanted candidate parses.
Scoring is done by the following formula:
\begin{equation}
\label{formula:geo}
score^P(p)=\sqrt[m]{\prod_{i=1}^{m} score^R\Big(rns_i(p),rfs_i(p)\Big)}
\end{equation}
\noindent
where the superscript $P$ refers to ``parse",
$p$ is the index for a specific candidate parse,
$m$ is the number of (role-name, role-filler) pairs in this specific candidate parse,
$rns_i(p)$ and $rfs_i(p)$ return the $i$-th role-name and role-filler of the $p$-th candidate parse, respectively.

\begin{figure}[htbp!]
    \centering
    \includegraphics[scale=0.3]{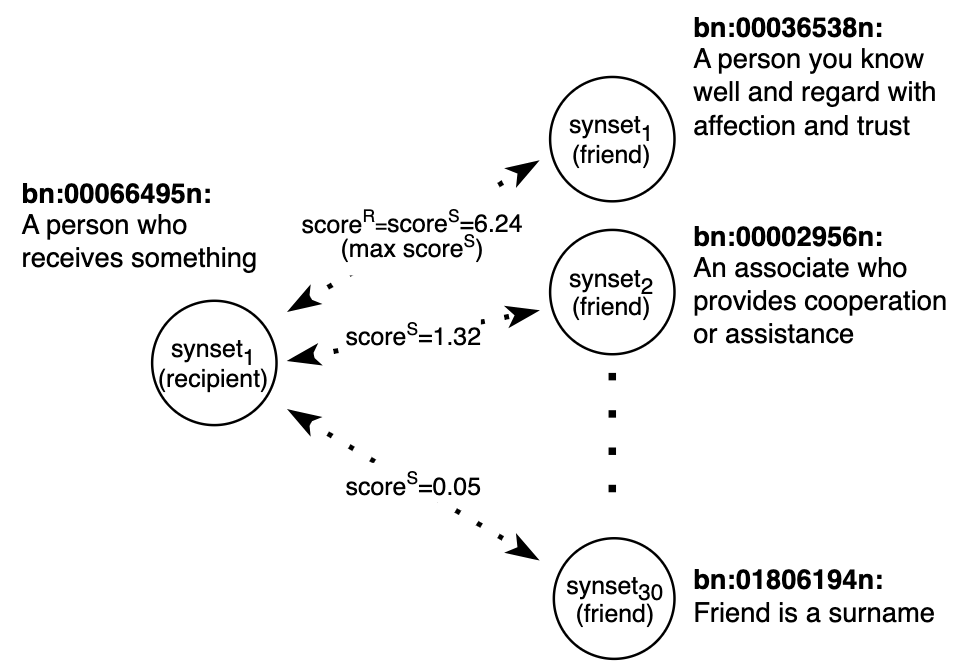}
    \caption{$score^{R}$ computation for role-name/role-filler pair (\texttt{Recipient}, \texttt{friend}) from candidate parse (\ref{code:parse-1})} 
    \label{fig:disamb}
\end{figure}

We illustrate the entire role-filler disambiguation process using two candidate parses, (\ref{code:parse-1}) and (\ref{code:parse-2}).
First, the best role-filler word-sense computation stage is performed.
For the role-name/role-filer pair (\texttt{Recipient}, \texttt{friend}) from candidate parse~(\ref{code:parse-1}),
$score^{R}(\text{``Recipient"},
\text{``friend"})$ is computed as illustrated in Figure \ref{fig:disamb}.
For each combination of $synset_i(\text{``Recipient"})$ and $synset_j(\text{``friend"})$, where $i\in\{1\}$ and $j\in\{1..30\}$,
the corresponding $score^S$ is computed by applying Formula (\ref{formula:disamb-2}).
It turns out that when ``\textit{friend}" is interpreted as the synset \texttt{bn:00036538n:}``\textit{A person you know well and regard with affection and trust,}" \\
$score^S\big(synset_1(\text{``Recipient"}),synset_1(\text{``friend"})\big)$ has the maximum value of $6.24$,
surpassing all other values of $score^S$ for different synset combinations.
Thus, $score^{R}(\text{``Recipient"}, \text{``friend"})$ is computed via Expression (\ref{formula:disamb}):
$$
\begin{aligned}
&\quad \,\, score^{R}(\text{``Recipient"}, \text{``friend"})\\
&=\max_{i \in \{1..1\},j \in \{1..30\}}\Big(\big\{score^{S}\big(synset_i(\text{``Recipient"}),synset_j(\text{``friend"})\big)\big\}\Big)\\
&=\max\big(\{6.24,1.32,...,0.05\}\big)\\
&=6.24
\end{aligned}
$$
Similar computations give us $score^{R}(\text{``Buyer"},\text{``customer"})=8.10$, as well as \\
$score^{R}(\text{``Goods"},\text{``watch"})=3.64$.
Next, using Formula (\ref{formula:geo}), the best candidate parse computation determines $score^P(1)$ and $score^P(2)$ for the candidate parses (\ref{code:parse-1}) and (\ref{code:parse-2}), respectively.
In this case, $score^{P}(1)$ comes to $5.68$, which is greater than $score^P(2)=1.81$.
As a result, KALM ultimately selects the candidate parse (\ref{code:parse-1}) for the sentence ``\textit{A customer purchases a watch for a friend,}" and eliminates the candidate parse (\ref{code:parse-2}).

\subsubsection{Construction of ULRs}
Ultimately, the disambiguated candidate parses are translated into \textit{unique logical representations} (ULRs), which gives the true meaning to the original CNL sentence and is suitable for querying. ULR uses the predicates \texttt{frame/2} and \texttt{role/2} for representing instances of the frames and the roles. The predicates \texttt{synset/2} and \texttt{text/2} are used to specify synset and textual information. For example, ``\textit{A customer buys a watch}'' will be converted into the ULR shown below:

\begin{verbatim}
frame(id_1,"Commerce_buy").
role(id_1,"Buyer",id_2).
text(id_2,customer).
synset(id_2,"bn:00019763n").
role(id_1,"Goods",id_3).
text(id_3,watch).
synset(id_3,"bn:00077172n").
\end{verbatim}

%% file: 2_related/5_nlp.tex
\section{Natural Language Processing and Generation}

This section introduces natural language processing and generation (NLP and NLG), by drawing our attention to a number of neural models that can help us build a more powerful knowledge authoring system.

\subsection{\Stanza}
\Stanza\footnote{\url{http://stanza.run/}} \cite{qi2020stanza} is an open-source NLP toolkit developed by the Stanford NLP group.
It takes 66 natural languages as input and provides a wide range of language analysis capabilities with fast speed and accurate results.
\Stanza is built on top of a powerful neural network-based architecture, which enables it to perform complex language analysis tasks with high accuracy.

The Stanza pipeline is composed of several stages, which are applied in a specific order to raw text to produce structured output. Here we present them as a sequence of steps:
\begin{enumerate}
    \item Tokenization: This stage splits the input text into individual words, punctuations, and other meaningful units, which are known as tokens.
    \item Multi-word token expansion: This stage is optional and is used to expand multi-word tokens into a sequence of single-word tokens.
    \item Part-of-speech (POS) tagging: This stage assigns a grammatical category (e.g., noun, verb, adjective, etc.) to each token.
    \item Lemmatization: This stage reduces each word to its base or dictionary form, a.k.a. lemma, using the POS tagging results.
    \item Dependency parsing: This stage analyzes the syntactic structure of a sentence and identifies the relationships between its words using the POS tagging results. The parser of this stage was refined based on the neural
    dependency parser proposed in \cite{dozat2016deep}.
    \item Named entity recognition (NER): This stage identifies and classifies named entities, such as people, organizations, and locations, in the input text.
    \item Sentiment analysis: This stage is optional and is used to identify the sentiment or emotional tone of the input text.
\end{enumerate}

In this thesis, we will use \Stanza to parse sentences that are very flexible grammatically.
Additionally, we aim to improve \Stanza's parsing ability to support more accurate knowledge authoring with fewer language-induced constraints.
Details are found in Chapter \ref{chap:4:kalmfl}.

\subsection{Representation Learning}\label{sec:rep-learning}
Representation learning is a type of machine learning paradigm where the goal is to automatically discover and learn a meaningful and efficient representation of different types of input data, including text \cite{mikolov2013efficient,pennington2014glove,bojanowski2017enriching,peters2018deep,devlin2018bert,borca2022provable,guo2024evolution,guo2022hierarchies}, image \cite{krizhevsky2012imagenet,he2016deep,he2017mask,dosovitskiy2020image}, 
audio \cite{gillespie2020improving,guo2019inferring}, video \cite{tran2015learning,feichtenhofer2019slowfast,arnab2021vivit,carreira2017quo}, graph \cite{perozzi2014deepwalk,guo2021subset,guo2022verba,guo2022subset,sultan2022low,chen2023accelerating,zhang2023subanom,zhou2024iterative,sultan2022low,guo2023analyzing}, time series \cite{wang2017time,zhang2018measuring,zhang2020developing,zhang2021dynehr,zhang2023semi}, robotics \cite{chiang2014real,guo2019inferring}, recommendation system \cite{lin2023comet}, and so forth.
In this work, we focus on epresentation learning in NLP, which involves converting raw text into meaningful numerical representations that can be understood by machine learning algorithms.
Since the introduction of word embeddings~\cite{mikolov2013distributed}, representation learning in NLP has continued to advance rapidly.
Contextual word embeddings and transformer models, such as Bidirectional Encoder Representations from Transformers (BERT)~\cite{devlin2018bert}, take surroundings of the word into account.
To produce a contextual word embedding for each token in the input sentence,
BERT commonly uses the average of all embeddings of the words in the input sentence to perform downstream prediction tasks.
We call this \textit{sentence embedding}.
These models have achieved state-of-the-art quality on a wide range of NLP tasks.

In this thesis, we use two specializations of BERT, known as Relation BERT and Sentence BERT.
Relation BERT is specializing the learning phase of BERT to produce models that capture relations between words in a sentence.
Sentence BERT is also specializing the learning phase of BERT, with a distinct objective of better reflecting the meaning of sentences.

\subsubsection{Relation BERT}
\textit{Relation} BERT (RelBERT)~\cite{ushio2021distilling} aims to capture the relationships between words in a text.
RelBERT takes pairs of words as input and encodes their relationships into vectors.
For example, the relationships between the input word pair (``\textit{Paris}",``\textit{France}") may include ``capital-of", ``city-of", etc., and RelBERT can represent this set of relationships as an embedding.
To achieve this goal, RelBERT was trained through the following pipeline.
We use the word-pair (``\textit{Paris}",``\textit{France}") to illustrate the pipeline.

\begin{enumerate}
    \item \emph{Prompt generation with word pairs}.
    This step generates complete sentences, called \textit{prompts}, that express
    relationships between the words in the input word-pair $(h,t)$
    A basic prompt generation strategy is to rely on manually created templates such as ``\textit{I am looking for a relation between $[h]$ and $[t]$: $[h]$ is the \textless$mask$\textgreater\ of $[t]$},"
    where \textless$mask$\textgreater\ is a \textless$mask$\textgreater-token representing the missing relationship, and $[h]$ and $[t]$ are slots that are filled with the head word $h$ and tail word $t$ from the given word pair $(h, t)$, respectively.
    Using the word pair (``\textit{Paris}", ``\textit{France}"), RelBERT completes the template and produces a \textit{ground prompt}, such as ``\textit{I am looking for a relation between Paris and France: Paris is the \textless$mask$\textgreater\ of France}."
    
    \item BERT encoding.
    For each ground prompt, BERT encodes it into a contextual embedding.
    In our example, BERT takes the ground prompt ``\textit{I am looking for a relation between Paris and France: Paris is the \textless$mask$\textgreater\ of France}"
    and averages the embeddings for all tokens in the prompt.
    This averaged embedding effectively encapsulates the relationship information associated with the word-pair (``\textit{Paris}", ``\textit{France}").
    We refer to this embedding as the \textit{relationship embedding} of the pair.
    
    \item \textit{RelBERT fine-tuning}
    is an additional learning process that tries to improve the initial BERT model used by RelBERT.
    The objective function of this learning process represents some sort of a distance between pairs of words and is based on the following intuition:
    the distance between the pairs (``\textit{Cannes}", ``\textit{France}") and (``\textit{Berlin}", ``\textit{Germany}") must be fairly small because Cannes and Paris are both cities,
    but the distance between (``\textit{Paris}", ``\textit{France}") and (``\textit{Berlin}", ``\textit{Germany}") must be even smaller because Paris and Berlin are not only cities but also capitals.
    At the same time, the distance between pairs like (``\textit{Paris}", ``\textit{France}") and (``\textit{Apple}", ``\textit{company}") should be fairly large.
    The fine-tuning process is intended to iteratively diminish the first two kinds of distances and increase the third.
\end{enumerate}

Chapter~\ref{chap:x:disam} will discuss how we use RelBERT's relationship embeddings to optimize role-filler disambiguation and thus improve the performance of KALM and its extensions.

\subsubsection{Sentence BERT}\label{sec:sbert}
\textit{Sentence} BERT (SBERT)~\cite{reimers2019sentence} aims to improve BERT's capacity for capturing sentence-level semantics.
For example, consider Sentence 1 ``\textit{Mary buys a car,}" Sentence 2 ``\textit{Bob buys a car,}" and Sentence 3 ``\textit{Mary makes a purchase of a car.}"
Despite the fact that Sentences 1 and 3 are semantically closer 
to each other than to Sentence 2, the distance between the original BERT embeddings of Sentences 1 and 2 is smaller than the distance between Sentences 1 and 3.
The SBERT embedding is designed to better reflect the semantics of these sentences and ensure that the distance between the embeddings of Sentences 1 and 3 is smaller than between Sentences 1 and 2.

SBERT achieves this by fine-tuning the original BERT using a similar objective function as RelBERT, aiming at diminishing the distance between Sentences 1 and 3 and increasing it for Sentences 1 and 2.
Over time, the distance between embeddings becomes more indicative of the semantic similarity between their corresponding sentences, and sentences with similar semantics tend to have embeddings that are close enough to each other.


Chapter~\ref{chap:x:disam} of this thesis uses SBERT to optimize KALM's role-filler disambiguation.

\subsection{Generative Pre-trained Transformer}\label{sec:gpt}

The Generative Pre-trained Transformer (GPT) \cite{radford2018improving,radford2019language,brown2020language} is a series of large language models (LLMs) developed by OpenAI.
This thesis focuses on GPT-3~\cite{brown2020language}, which was the most powerful language model just a few months ago when we conducted our experiments.
In the following discussion, for the sake of brevity, we will use GPT to refer to GPT-3.

Unlike BERT models, GPT directly generates response to a given text instead of encoding the text as an embedding.
This
allows GPT to function as a human-level chatbot, a textual content creator, and most importantly to this work, as an NLP task-solver.
A GPT prompt can be described as a concrete instruction intended to guide text generation.
By utilizing prompts, GPT can perform various NLP tasks, including language translation, text completion, and many more.
\textit{Few-shot learning} is the most common GPT prompting technique to solve NLP tasks.
Compared to the traditional paradigm of few-shot learning, which typically trains models 
to generalize well 
to unseen data, GPT's few-shot learning is achieved
simply by supplying limited amount of labeled data as part of the prompts \emph{without} the need for any kind of training.
For instance, in the named entity recognition (NER) task, the goal is to identify real-world objects such as persons, locations, organizations, products, and more, that can be denoted with a proper name.
Traditional few-shot NER taggers are trained on labels, say, Person, Location, and Organization.
Then, by training with a few examples of the new Product label,
such as ``\textit{\underline{Instant Pot}\textsubscript{[Product]} is very good to use,}"
the taggers become capable of recognizing this new label.
In contrast, GPT's few-shot learning utilizes only a prompt consisting of three parts: a task description, a few-shot demonstration, and prompting.

\begin{example}
GPT is asked to complete a NER task of the product entity recognition.
We prompt GPT with a task description for product entity recognition, three product entity recognition examples, and a sentence with or without product entities to be recognized:
\begin{verbatim}
Perform named entity recognition (NER). Recognize real-world
objects of product that can be denoted with a proper name.
============================================================
Instant Pot is very good to use. ==> Instant Pot
Boeing 747s are manufactured in the US. ==> Boeing 747s
Mary donated 300 dollars to Stony Brook University. ==> None
============================================================
John lost his MacBook today. ==>
\end{verbatim}
GPT response:\\
\verb|MacBook|\\
\end{example}

In this way, GPT's few-shot learning can help to perform the in-context definition generation task, defined in Section~\ref{sec:kalmfl-compsys}.
This requires only a small amount of training data and allows KALM's costly BabelNet-based role-filler disambiguation to be replaced with a much faster GPT-based disambiguation procedure.
Details are in Chapter \ref{chap:x:disam}.

%% file: 3_kalmfl/0_main.tex
\chapter{\texorpdfstring{\KALMF:}\space\space From CNL to Factual (English) Language}\label{chap:4:kalmfl}

In this chapter, we begin by introducing a user-friendly version of English for knowledge authoring.
We refer to this English as \textit{factual English} and its sentences as \textit{factual sentences}, which can be easily learned without requiring extensive training.
Then, we modify the \Stanza toolkit and propose a neural parser multi-\Stanza (\MS) that generates several ranked parses for input sentences.
Finally, we describe \KALMF, a product of adaptation of the KALM framework to factual English sentences---a language that is significantly less restricted than any known CNL and is much easier to learn.

\input{3_kalmfl/1_factual}
\input{3_kalmfl/2_mstanza}
\input{3_kalmfl/3_kalmfl}
\input{3_kalmfl/4_eval}

%% file: 3_kalmfl/1_factual.tex
\section{Factual Sentence}\label{sec:factual}
\label{chap:3:factual}

In knowledge authoring, we are not interested in fine letters but rather in sentences that express or query knowledge, such as facts, queries, rules, and actions.
So, one should expect that our input language is simpler and more structured.
In this chapter, we limit ourselves to facts and queries and more advanced types of knowledge, such as rules and actions, are left to Chapter~\ref{chap:5:kalmra}.
Consequently, here we focus on sentences for specifying and querying sets of facts, which we call \emph{factual  sentences}.
Non-factual sentences, like ``\textit{Go fetch more water},'' do not express any factual information and can thus be excluded from consideration.

\subsection{Definition}

Before defining factual sentences, we first remind some key grammatical concepts. A \emph{clause} is a unit of grammatical organization that contains a verb and usually other components.
A \emph{main clause}\footnote{\url{https://www.lexico.com/en/definition/main_clause}} is a clause that can form a complete sentence standing alone and having a subject and a predicate.
A \emph{subordinate clause} depends on a main clause for its meaning.
Together with the main clause, a subordinate clause forms part of a \emph{complex sentence}.
There are four types of subordinate clauses including adnominal clauses, adverbial clauses, clausal complements, and clausal subjects.
A \emph{coordination} is a syntactic structure that links together two or more elements with connectives such as ``\textit{and}'' and ``\textit{or}'' (e.g., \textit{a car and a watch}).
When the elements are main clauses, a \emph{compound sentence} is formed (e.g., ``\textit{Mary wants the car and the car is available}'').

Examination of various datasets shows that main clauses, compound sentences, and sentences with adnominal clauses (e.g. ``\textit{Mary bought a car made in USA}'') are by far the most common constructs in datasets that contain data and queries. In contrast, clausal complements and other types of subordinate clauses are typically non-factual or they are used to describe other kinds of logical statements, such as rules, which will be handled in Chapter~\ref{chap:5:kalmra}. For the same reason, connectives other than ``\textit{and}" and ``\emph{or}" are also eliminated. We then define \emph{factual sentence} for knowledge authoring as follows:

\begin{definition}
A \emph{factual sentence} is
\begin{enumerate}
    \item a factual main clause with subordinate adnominal clauses (if any), and no other subordinate clauses; or
    \item a compound sentence where the connectives connect only the clauses of the kind described in 1.
\end{enumerate}
\end{definition}

\begin{definition}
A \emph{main clause is factual}\footnote{According to \url{https://www.lexico.com/en/definition/main_clause}, all main clauses are factual.} if
\begin{itemize}
    \item it has a verb with a subject (e.g., ``\textit{Mary bought a car}''); or
    \item it has a nominal word (or an adjective) with a subject and a linking verb (e.g., ``\textit{Mary is rich},'' where an adjective \textit{rich} has a subject \textit{Mary} and linking verb \textit{is})
\end{itemize}
\end{definition}

\subsection{Necessary Properties}

We now use POS tags (part of speech) and universal dependencies to describe six \emph{grammatical properties for factual sentences} that follow from the aforesaid factual restriction on the English sentences, and thus are \emph{necessary conditions} for sentences to be factual.
We then use these properties to discover and correct errors made by the \Stanza parser.
We use the superscripts \texttt{U} and \texttt{X} to refer to universal POS\footnote{\url{https://universaldependencies.org/u/pos/}} (UPOS) tags, Penn Treebank extended POS\footnote{\url{https://www.ling.upenn.edu/courses/Fall_2003/ling001/penn_treebank_pos.html}} tags (XPOS), and \texttt{UD} will refer to universal dependency\footnote{\url{https://universaldependencies.org/u/dep/}} labels.

\begin{property}
\label{prop:1}
If the main clause is factual, then
    \begin{itemize}
        \item the main clause has a verb with a subject. That is, the clause has a word with an incoming \ud{root} edge tagged with \upos{VERB} and an outgoing \ud{nsubj} edge); or
        \item the main clause has a nominal word (or an adjective) with a subject and a linking verb. Thus, the clause has a word with an incoming \ud{root} edge that is (i) tagged with \upos{NOUN}, \upos{PRON}, \upos{PROPN},  or \upos{ADJ}; (ii) has an outgoing \ud{nsubj} edge; and (iii) has an outgoing \ud{cop} edge (copula).
    \end{itemize}
\end{property}

\begin{property}
\label{prop:2}
If a word $W$ is the last element of a coordination (e.g., ``\textit{watch}" in ``\textit{a car or a watch}"), then this coordination must be an \textit{and}- or an \textit{or}-coordination. That is, $W$ has an incoming \ud{conj} edge and a outgoing \ud{cc} edge pointing to ``\textit{and}'' or ``\textit{or}.''
\end{property}

\begin{property}
\label{prop:3}
If a verb $V$ has one or more auxiliary verbs ${V_1^a,..., V_n^a}$, and $V_n^a$ (tagged with \upos{AUX}) is the closest auxiliary verb to $V$ (e.g., in the sentence ``\textit{A car has been bought by Mary}," $V_1^a=$ \textit{has}, $V_n^a=V_2^a=$ \textit{been}, $V=$ \textit{bought}), then
    \begin{itemize}
        \item continuous tense ($V_n^a$ is \textit{be} -- $V$ is a present participle): $V_n^a$ has an incoming \ud{aux} edge starting from $V$, and $V$ is tagged with \xpos{VBG}; or
        \item perfect tense ($V_n^a$ is \textit{have} -- $V$ is a past participle): $V_n^a$ has an incoming \ud{aux} edge starting from $V$, and $V$ is tagged with \xpos{VBN}; or
        \item past, present, and future tense ($V_n^a$ is \textit{can / do / may / must / ought / should / will} -- $V$ in base form): $V_n^a$ has an incoming \ud{aux} starting at $V$, and $V$ is tagged with \xpos{VB}; or
        \item passive voice ($V_n^a$ is \textit{be/get} -- $V$ is a past participle): $V_n^a$ has an incoming \ud{aux:pass} edge starting from $V$, and $V$ is tagged with \xpos{VBN}
    \end{itemize}
\end{property}

\begin{property}
\label{prop:4}
For a verb $V$ without auxiliary verbs (no outgoing \ud{aux} / \ud{aux:pass} edges):
    \begin{enumerate}
    
        \item if $V$ is a present or past participle (i.e., tagged with \xpos{VBG} / \xpos{VBN}), then
        \begin{itemize}
            \item $V$ occurs in a coordination (i.e., has an incoming \ud{conj} edge); or
            \item $V$ occurs in adnominal clauses (i.e., has an incoming \ud{acl} edge)
        \end{itemize}

        \item if $V$ is in present or past tense (i.e., tagged with \xpos{VBP} / \xpos{VBZ} / \xpos{VBD}), then
            \begin{itemize}
                \item $V$ occurs in a coordination (i.e., has an incoming \ud{conj} edge); or
                \item $V$ occurs in main/adnominal clauses (i.e., has an incoming \ud{root} / \ud{acl} /  \ud{acl:relcl} edge) and have a subject (i.e., an outgoing \ud{nsubj} edge)
            \end{itemize}
        
        \item if $V$ is in the base form (i.e., tagged with \xpos{VB}), then
            \begin{itemize}
                \item $V$ occurs in a coordination (i.e., has an incoming \ud{conj} edge); or
                \item $V$ occurs in adnominal clauses with infinitive form (i.e., has an incoming \ud{acl} edge and an outgoing \ud{mark} edge pointing to ``\textit{to}'')
            \end{itemize}
        
    \end{enumerate}
\end{property}

\begin{property}
\label{prop:5}
If a non-verb word $W$ has one or more auxiliary verbs ${V_1^a,..., V_n^a}$, where $V_n^a$ is the closest auxiliary verb to $W$ (e.g., in the sentence ``\textit{Mary has been rich}," $V_1^a=$ \textit{has}, $V_2^a=$ \textit{been}," $W=$ \textit{rich}, and \textit{been} is the closest auxiliary verb to \textit{rich}), then $V_n^a$ and $W$ must satisfy these properties:
    \begin{enumerate}
        \item $W$ is a nominal word or an adjective (i.e., tagged with \upos{NOUN} / \upos{PRON} / \upos{PROPN} / \upos{ADJ})
        \item $V_n^a$ is the copula of $W$ (i.e., $V_n^a$ has an incoming \ud{cop} edge starting from $W$)
        \item $W$ has a subject (i.e., has an outgoing \ud{nsubj} edge)
    \end{enumerate}
\end{property}

\begin{property}
\label{prop:6}
The sentence must be \emph{projective}. Given a parse, if there are crossing edges (e.g., the incoming edges for ``\textit{of}'' and ``\textit{Winston}'' in Fig.~\ref{fig:chap3:non-proj}), the sentence is called \emph{non-projective}, otherwise it is \emph{projective}. Property~\ref{prop:6} expresses the belief held by linguists that well-constructed English sentences are typically projective, and so are factual sentences.

\begin{figure}[htbp!]
    \centering
    \includegraphics[scale=0.5]{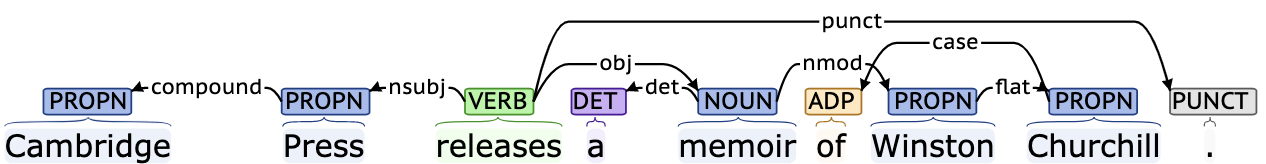}
    \caption{A example of a non-projective parse}
    \label{fig:chap3:non-proj}
\end{figure}
\end{property}

%% file: 3_kalmfl/2_mstanza.tex
\section{Multi-\Stanza}\label{sec:mstanza}

As introduced in Chapter \ref{chap:2:relwork}, \Stanza~\cite{qi2020stanza} is a pipelined neural parser with state-of-the-art performance, which was designed to return only the top parse for each sentence.
Unfortunately, we found that it frequently errs in POS tagging, and these errors then propagate to dependency parsing.
We then noticed that nearly top parses often give correct POS tags where the top parses err, so we modified \Stanza to return also some non-top-ranked parses.
We called the modified version multi-\Stanza, or, \MS.

\subsection{The Multi-\Stanza Pipeline}
\begin{figure}[htbp!]
    \centering
    \includegraphics[scale=0.32]{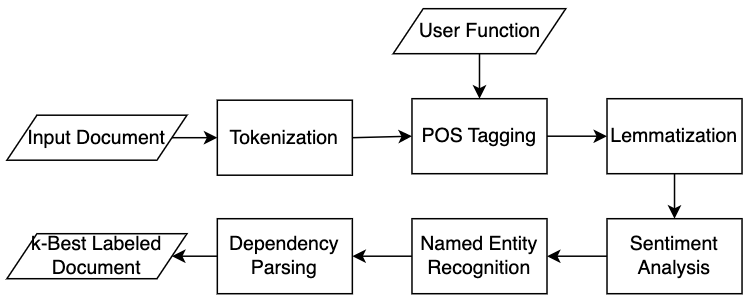}
    \caption{The pipeline of \MS}
    \label{fig:ms-arch}
\end{figure}

Figure~\ref{fig:ms-arch} shows the pipeline.
Each of these stages operates on \texttt{Document} objects. 
A \texttt{Document} object contains document-level annotations along with a list of \textsc{Sentence} objects, a \texttt{Sentence} object contains sentence-level annotations along with a list of \texttt{Word} objects, and a \texttt{Word} object contains the raw text of the word along with word-level annotations. 
Each stage adds multiple sets of annotations, creating a new \texttt{Document} object for each set.
These \texttt{Document} objects are then passed downstream.
Unlike \Stanza, the output of \MS~is a list of annotated \texttt{Document} objects ranked in the order of decreasing confidence.
The first item in the list is simply the result that \Stanza~would have returned, as \Stanza~always returns the highest-confidence result.
If $k$ results are generated, we refer to these results as $k$-best \texttt{Document}s.

\subsection{POS Tagger}
The Part-of-Speech (POS) tagger adds POS tags to each word. 
As each POS tag has a finite number of categories, it is straightforward to extract the $k$-best POS tags for each word, along with their confidences.
\MS allows users to provide a \textit{user function} to dynamically modify the list of $k$-best POS tags according to their needs.

\begin{example}
\label{exmp:ms-pos}
When parsing the sentence ``\textit{A student protests against the government}," the word ``\textit{protests}" is POS-tagged as \{\texttt{NOUN}: 0.732, \texttt{VERB}: 0.257, ...\}.
The real numbers represent confidence scores that determine the misassignment of \texttt{NOUN} to ``\textit{protests}," leading to incorrect dependency parsing for this sentence.
\end{example}
To address such error propagation, \MS enables users to intervene in POS-tagging with customized functions.
In Example~\ref{exmp:ms-pos}, the POS tag of ``\textit{protests}" can be determined by the following function:
$$
POS =
\begin{cases}
    POS_2, & \text{if \xspace} POS_1<0.8\\
    POS_1, & \text{otherwise}\\
\end{cases}
$$
\noindent
where $POS_1$ is the most confident POS tag, \texttt{NOUN}, and $POS_2$ is the second most confident one, \texttt{VERB}.
As a result, the POS tag of ``\textit{protests}" is finally corrected as \texttt{VERB}.

\subsection{Dependency Parser}
\MS generates a dependency parse by generating a fully connected directed graph, and generating the weights of the edges using a neural network.
The vertices of the graph represent the tokens of the document.
Between every pair of vertices, there are multiple directed edges, with a directed edge between vertex A and vertex B for every possible universal dependency edge. 
The neural network learns to assign the weight of the edge based on the type of edge and the relationship between the vertices.
Then, the minimum spanning arborescence is found and used as the dependency parse.
Multiple possible dependency parses for each sentence are combined in order to generate the next-best dependency parse for the entire document.
As a result, users can choose a less confident dependency parse when a noticeable error is found in a more confident one.

%% file: 3_kalmfl/3_kalmfl.tex
\section{\texorpdfstring{\KALMF}\xspace: KALM for Factual (English) Language}\label{sec:kalmf}

\begin{figure}[htbp!]
    \centering
    \includegraphics[scale=0.27]{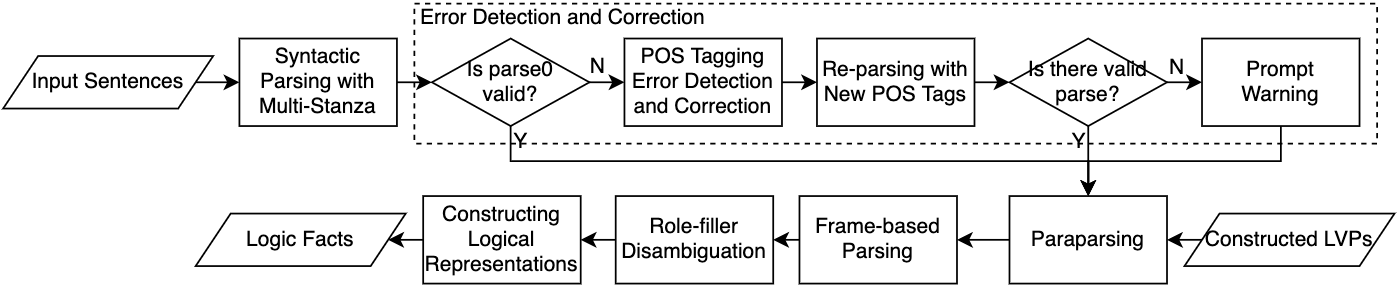}
    \caption{The \KALMF~framework}
    \label{fig:kalmf}
\end{figure}

Now we propose an extension of KALM to factual language, called \KALMF.
Compared to the original KALM framework shown in Fig.~\ref{fig:kalm}, \KALMF uses \MS instead of APE as the syntactic parser and includes two additional key steps (Error Detection and Correction, and Paraparsing) to tackle a slew of problems caused by the use of \MS.
The new \KALMF framework is shown in Fig.~\ref{fig:kalmf}.

\subsection{Error Detection and Correction}

\begin{figure}[htbp!]
    \centering
    \includegraphics[scale=0.5]{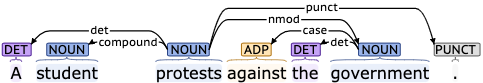}
    \caption{\Stanza error case}
    \label{fig:mstanzaerror}
\end{figure}

\KALMF checks \Stanza parses for being factual using the necessary conditions in Chapter \ref{chap:3:factual}.
If any of the checks don't pass, \KALMF attempts to correct the parse by conjecturing that some of the POS tags are wrong (a fairly common problem with \Stanza in our experience).
This is done by using other nearly top parses provided by \MS.
If the correction attempt fails, the user is asked to rephrase the sentence.
Fig.~\ref{fig:mstanzaerror} illustrates one of the errors in \Stanza POS tagging.
Here, the word ``\textit{protests}" is wrongly tagged as a noun and the dependencies related to ``\textit{protests}" are also wrong.

Fortunately, the six properties that stem from the factual sentence requirement
can detect and help correct some of these mis-taggings.
Denote the \MS parse with the highest confidence score as $parse_0=[w_1,...,w_n]$ where $n$ is the number of the words, $w_i$ contains all parsing information such as POS tag, dependency relation of the $i$-th word in the sentence. Let 
$Upos=[upos_1,...,upos_n]$ and $Xpos=[xpos_1,...,xpos_n]$ be the UPOS and XPOS taggings of the sentence. Note the corrected UPOS and XPOS tags for $w_i$ are $\overline{upos_i}$ and $\overline{xpos_i}$, respectively.

\begin{algorithm}[htbp!]
\caption{POS Tagging Error Detection and Correction}
\label{algo:pos}
\begin{algorithmic}[1]
\Require $upos_i$, $xpos_i$, the second best UPOS tag $upos_i^\prime$ for $w_i$, and the second best XPOS tag $upos_i^\prime$ for $w_i$.
\State $\overline{upos_i},\overline{xpos_i} \leftarrow upos_i, xpos_i$
\If{$upos_i.score<0.9$}
    \If {$upos_i==$ \upos{NOUN} and $upos_i^\prime==$ \upos{VERB}}
        \State $\overline{upos_i} \leftarrow$ \upos{VERB}
        \State $\overline{xpos_i} \leftarrow$ \xpos{VBP}/\xpos{VBZ}/\xpos{VBD}
    \ElsIf {$upos_i==$ \upos{VERB} and $upos_i^\prime==$ \upos{AUX}}
        \State $\overline{upos_i} \leftarrow$ \upos{AUX}
        \State $\overline{xpos_i} \leftarrow$ \xpos{VBP}/\xpos{VBZ}/\xpos{VBD}
    \ElsIf {$upos_i==$ \upos{PRON} and $upos_i^\prime==$ \upos{DET}}
        \State $\overline{upos_i} \leftarrow$ \upos{DET}
        \State $\overline{xpos_i} \leftarrow$ \xpos{WDT}/\xpos{PDT}/\xpos{DT}
    \ElsIf {$upos_i==$ \upos{SCONJ} and $upos_i^\prime==$ \upos{ADV}}
        \State $\overline{upos_i} \leftarrow$ \upos{ADV}
        \State $\overline{xpos_i} \leftarrow$ \xpos{WRB}/\xpos{IN}
    \EndIf
\ElsIf{$xpos_i.score<0.9$}
    \If {$xpos_i==$ \xpos{VBD} and $xpos_i^\prime==$ \xpos{VBN}}
        \State $\overline{xpos_i} \leftarrow$ \xpos{VBN}
    \ElsIf{$xpos_i==$ \xpos{VBN} and $xpos_i^\prime==$ \xpos{VBD}}
        \State $\overline{xpos_i} \leftarrow$ \xpos{VBD}
    \ElsIf{$xpos_{1j}==$ \xpos{VBP} and $xpos_i^\prime==$ \xpos{VB}}
        \State $\overline{xpos_i} \leftarrow$ \xpos{VB}
    \EndIf
\EndIf
\State \Return $\overline{upos_i}, \overline{xpos_i}$
\end{algorithmic}
\end{algorithm}

\subsubsection{POS Tagging Error Detection and Correction}
If $parse_0$ satisfies all the above-mentioned factual properties, $parse_0$ is assumed to be error-free and is directly sent to the Paraparsing step.
Otherwise, following Algorithm~\ref{algo:pos}, if a possibly wrong (with confidence $<$ 0.9) POS tag belongs to a certain type of frequent POS tagging errors (e.g., \MS relatively frequently mis-tags verbs as nouns, and this is what happened with ``\textit{protests}" in Fig.~\ref{fig:mstanzaerror}), \KALMF then asserts the tag is wrong and corrects it.
Lines 2 and 3 in Algorithm~\ref{algo:pos} capture such type of errors and assign corrected POS tags. Note that in Lines 5, 8, 11, and 14, the algorithm faces multiple options like \xpos{VBP}/\xpos{VBZ}/\xpos{VBD} and chooses the one with the highest confidence score.

\subsubsection{Re-parsing with New POS Tags}
Having re-tagged the words in the above step, the new POS tags,
$\overline{Upos}=[\overline{upos_1},...,\overline{upos_n}]$ and $\overline{Xpos}=[\overline{xpos_1},...,\overline{xpos_n}]$, are fed to the \MS dependency parser to re-generate a new list of dependency parses $\overline{Parse}$, ranked by confidence scores.
Then, \KALMF goes through the parses in $\overline{Parse}$, from the highest confidence score to lowest, looking for a parse, $parse'$, that satisfies all the properties of factual sentences.
If $parse'$ is found, it is taken as a corrected parse, $\overline{parse}=parse'$.
If $parse'$ is not found, the algorithm assumes the sentence in question is not factual, so it asks the user to paraphrase the sentence.

\subsection{Paraparsing}

Paraparsing is a set of corrective steps that modify the original $\overline{parse}$.
The aim here is to
eliminate possible semantic mismatches that were the original motivation for KALM, as explained in the introduction.
The mismatches handled here arise from the possibility that the same information may be described via passive or active voice, via a different order of elements in a coordination, via the different ways to attach adnominal clauses, and more.
Note that all these corrections became possible in \KALMF due to
the use of dependency parsing and were not possible in the original CNL-based KALM.

\subsubsection{Passive Voice}
\begin{figure}[htbp!]
     \centering
     \subfigure[Active voice]
         {\includegraphics[width=0.44\textwidth]{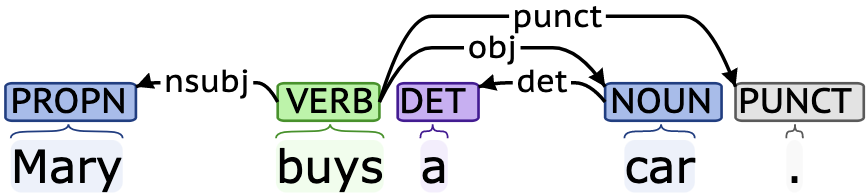}}
     \subfigure[Passive voice]
         {\includegraphics[width=0.63\textwidth]{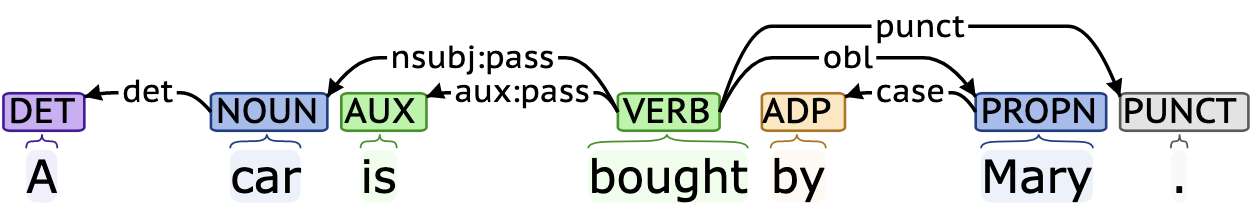}}
     \caption{Semantic mismatch caused by passive voice}
     \label{fig:paraparsingpassive}
\end{figure}

\MS handles the active and passive voices separately. 
For a pair of active/passive voice sentences with the same meaning, such as ``\textit{Mary buys a car}'' and ``\textit{A car is bought by Mary},''
\Stanza~gives two completely different parses shown in Fig.~\ref{fig:paraparsingpassive};
it does not attempt to reconcile the semantic mismatch between them so that they would yield the same logical representation.
To address this problem, \KALMF first recognizes passive voice by the \ud{aux:pass} edge in the parse, 
then modifies the edges of passive voice parses to make the parses equivalent to their active voice counterparts.
If the sentence is in active voice, keep the parse unchanged.
Otherwise, convert it into $n$ parses in active voice ($n$ is the number of \textit{by}-phrases in the clause,
since every \textit{by}-phrase could be the subject of the real active voice counterpart of this passive voice sentence) by
modifying (i) \ud{nsubj:pass} to \texttt{obl:by} and
(ii) $n$ \texttt{obl:by} edges to \ud{nsubj} one by one.

\subsubsection{Coordination}
\begin{figure}[htbp!]
     \centering
     \subfigure[]{\includegraphics[width=0.85\textwidth]{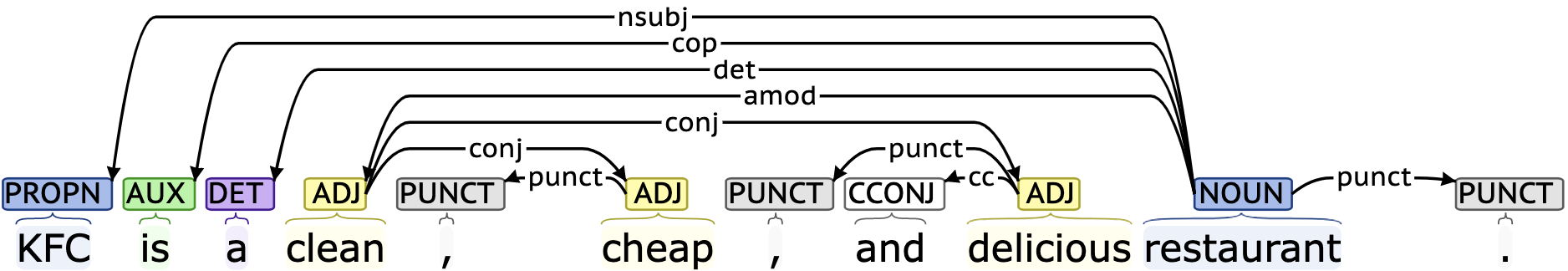}\label{fig:paraparsingcoord1}}
     \subfigure[]{\includegraphics[width=0.85\textwidth]{figs/paraparsingcoord2.png}\label{fig:paraparsingcoord2}}
     \caption{Semantic mismatch caused by coordination}
\end{figure}

Elements in \Stanza~coordinations are not treated equally.
For example, in the parse of ``\textit{KFC is a cheap, clean, and delicious restaurant}'' shown in Fig.~\ref{fig:paraparsingcoord1},
``\textit{cheap}" directly depends on ``\textit{restaurant}," 
but ``\textit{clean}" and ``\textit{delicious}" mutually depend on ``\textit{cheap}" instead of ``\textit{restaurant}".
In this case, if ``\textit{cheap}" and ``\textit{clean}" are swapped, the meaning of the sentence stays unchanged, 
but the parse will be different as shown in Fig.~\ref{fig:paraparsingcoord2}.
This phenomenon will lead to a semantic mismatch.

\KALMF treats coordination elements equally by modifying their edges. The procedure is shown below, and all examples refer to Fig.~\ref{fig:paraparsingcoord1}.

\begin{enumerate}
    \item Locate the root element $el_{root}$ of the coordination. It is a word that has outgoing \ud{conj} edges to other elements, which have incoming \ud{conj} edges (e.g. $el_{root}$ is ``\textit{cheap}," while ``\textit{clean}" and ``\textit{delicious}" are the other two elements);
    
    \item Copy the incoming edge of $el_{root}$ to each non-root element and delete the edge \ud{conj} (e.g. copy \ud{amod} to replace \ud{conj} that goes to ``\textit{clean}" and ``\textit{delicious}");
    
    \item Copy the outgoing edges of $el_{root}$ (other than the deleted \ud{conj}) to each non-root element (in our example, ``\textit{cheap}" has no outgoing edges, so no need to copy anything).
\end{enumerate}

\subsubsection{Adnominal Clause}
\begin{figure}[htbp!]
     \centering
     \subfigure[]{\includegraphics[width=0.76\textwidth]{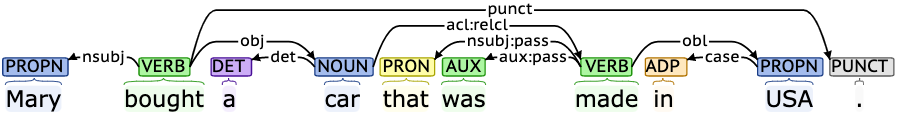}\label{fig:paraparsingnominal1}}
     \subfigure[]{\includegraphics[width=0.53\textwidth]{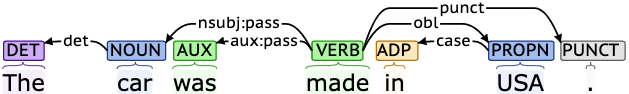}\label{fig:paraparsingnominal2}}
     \caption{Semantic mismatch caused by adnominal clause}
\end{figure}

An adnominal clause describes a fact about the nominal word it modifies.
For example, ``\textit{Mary bought a car that was made in USA}'' represents two facts: ``Mary bought a car,'' and ``The car was made in USA''.
However, the second fact is parsed differently when it is in an adnominal clause than when it is in a sentence by itself, and such phenomena lead to semantic mismatch.
As shown in Fig.~\ref{fig:paraparsingnominal1}, the subject of ``\textit{a car that was made in USA}'' is ``\textit{that}'' whereas the real subject should be ``\textit{car}'' like the parse in Fig.~\ref{fig:paraparsingnominal2}.

\KALMF recognizes the real roles (e.g., subject, object, etc.) that the modified word plays in the adnominal clause,
so that the adnominal clause can be seen as a complete sentence all by itself.
This is done via the following transformation.

\begin{itemize}
    \item if a word $V_1$ has an incoming \ud{acl} edge $e_1$ that starts at $V_2$ and has no outgoing \ud{nsubj} or \ud{nsubj:pass} edges, then 
        \begin{itemize}
            \item if $V_1$ is a present participle or a base-form verb tagged with \xpos{VBG} or \xpos{VB}, flip the direction of $e_1$ and change the label to \ud{nsubj};
            
            \item if $V_1$ is a past participle tagged with \xpos{VBN}, flip the direction of $e_1$ and change the label to \ud{nsubj:pass}.
        \end{itemize}

    \item if a word $V_1$ has an incoming \ud{acl:relcl} edge $e_1$ that starts at $V_2$, then
        \begin{itemize}
            \item if $V_1$ has an outgoing \ud{nsubj}, \ud{nsubj:pass} or \ud{obj} edge pointing to an introductory word $W_{intro}$ ``\textit{that}/\textit{who}/\textit{which},'' replace $W_{intro}$ with $V_2$;
            
            \item if $V_1$ has an outgoing \ud{mark} edge pointing to an introductory word $W_{intro}$ \\ ``\textit{where}/\textit{when}/\textit{why}/\textit{which}," replace $W_{intro}$ with $V_2$ and modify \ud{mark} to \ud{obl}.
        \end{itemize}
\end{itemize}

\subsubsection{Other Semantic Mismatches}
Besides the most frequent semantic mismatch issues solved above,
\KALMF also tackles other types of semantic mismatch caused by lemmatization, particle verbs, prepositional phrases, named entities, indirect objects, and so forth.
As shown in Fig.~\ref{fig:kalmf}, after the Paraparsing step,
the ultimate parse is delivered to Frame-based Parsing and undergoes further processing to ultimately yield a unique disambiguated logical representation.

\subsection{Representing Dependency Relations}

The original KALM used DRS to represent extracted information. In this paper, the parses are represented by much more general graphs, so we introduce an appropriate logical representation for them, following the syntax of XSB. Here is an example:

\begin{example}
\label{exmp:representation}
The \KALMF representation for the sentence ``\textit{Mary buys a car}'':

\begin{verbatim}
word(index(1,1,1),mary,
     [edge(index(1,2),jbusn)],
      edge(index(1,2),nsubj),propn,nnp,
      index(1,2),s_person,accepted).
word(index(1,2,1),buy,
     [edge(index(1,1),nsubj),edge(index(1,4),obj)],
     edge(index(1,0),root),verb,vbz,
     index(1,2),o,accepted).
word(index(1,3,1),a,
     [edge(index(1,4),ted)],
     edge(index(1,4),det),det,dt,
     index(1,2),o,accepted).
word(index(1,4,1),car,
     [edge(index(1,3),det),edge(index(1,2),jbo)],
     edge(index(1,2),obj),noun,nn,
     index(1,2),o,accepted).

\end{verbatim}
\end{example}
\noindent
where a sentence is represented by a set of \texttt{word/9} predicates and each \texttt{word/9} predicate represents a word $t$. The 1st argument in a \texttt{word/9} predicate includes sentence ID, candidate parse ID, and $t$'s ID. The 2nd argument is the lemma of $t$. The 3rd argument is a list of edges that connect $t$ to other words, and an \texttt{edge/2} predicate representing a specific edge $e$ includes the index of the other word on $e$, and the edge type (reversed if it is an in-coming one). The 4th argument is the one and only incoming edge that $t$ has. The 5th, 6th are $t$'s UPOS and XPOS tags, respectively. The 7th argument is the index of the root word in the whole sentence, namely, ``\textit{buys}" in Example \ref{exmp:representation}. Finally, the 8th and 9th arguments are the named entity and validation tags, where the latter indicates if the parse is factual (\texttt{accepted}) or not.

\subsection{Role-Filler Disambiguation and Unique Logical Representations}

In KALM, a clause represents a complete fact so each clause has only one parse, the one with the highest semantic score after disambiguation.
In \KALMF, coordinations and adnominal clauses are introduced and their meanings can be captured accordingly, which will be further explained in this section.

For logical representations, \KALMF~uses \texttt{ulr/2} and \texttt{role/3} for representing instances of final parses after disambiguation. Consider theses sentences: ``\textit{Mary bought a car for John}'' and ``\textit{Mary made a purchase of a car for John}''. Although they have different syntactic structures, they are ultimately converted into exactly the same parse and their role-fillers are assigned exactly the same synsets. Therefore, they must be translated into a unique logical representation (ULR). And indeed, the ULR for sentences ``\textit{Mary bought a car for John}'' and ``\textit{Mary made a purchase of a car for John}'' is the same:

\begin{verbatim}
ulr("Commerce_buy",
    [role("Buyer","Mary","bn:00046516n"),
     role("Goods",car,"bn:00007309n"),
     role("Recipient","John","bn:00046516n")]).
\end{verbatim}

\noindent
where the first argument, \texttt{"Commerce\_buy"}, is the frame name, and the second argument is a list of role descriptors.

\subsubsection{ULR for Factual Sentences with Coordination}

Generally, a sentence can have a mixture of ``\textit{and}"- and ``\textit{or}"-coordinations, whose meaning is quite hard to describe. For simplicity and less ambiguity, we focus on the case where $C$ has only one type of coordination, i.e., all connectives are ``\textit{and}" or all are ``\textit{or}".

Let $[C_1,...,C_n]$ be the list of all coordinations in a sentence $S$. A \emph{coordinated choice} is a list $\sigma=[el_1, ..., el_n]$ of coordination elements such that $el_i\in C_i$ for all $i=1,...,n$.
Let $S_\sigma$ be $S$ where each coordination, $C_i$, and its elements is replaced with the corresponding element $el_i$ from $\sigma$. Thus, for each coordinated choice $\sigma$ for $S$, the above replacement operation constructs another sentence, $S_\sigma$, which has no coordinations. Next, we collect $S_\sigma$ for all the different $\sigma$'s and organize these sentences as elements of a new homogeneous coordination of the same type as each of the original coordination $C_i$. The result is a sentence $S^\prime$ with only one coordination, found at the root of the parse for the sentence.

For example, for the sentence ``\textit{Mary bought and sold a car and a watch},'' the coordinated choices includes $\sigma_1=[bought,car]$, $\sigma_2=[bought,watch]$, $\sigma_3=[sold,car]$, $\sigma_4=[sold,watch]$. Based on the coordinated choices, we can construct 4 sentences without coordinations: ``\textit{Mary bought a car},'' ``\textit{Mary bought a watch},'' ``\textit{Mary sold a car},'' and ``\textit{Mary sold a watch},'' which are organized into a new ``\textit{and}"-coordination. Thus, we have the final ULR for this ``\textit{and}"-coordination shown below:

\begin{verbatim}
ulr("Commerce_buy",[role("Buyer","Mary","bn:00046516n"),
                    role("Goods",car,"bn:00007309n")]).
ulr("Commerce_buy",[role("Buyer","Mary","bn:00046516n"),
                    role("Goods',watch,"bn:00077172n")]).
ulr("Commerce_sell",[role("Seller","Mary","bn:00046516n"),
                     role("Goods",car,"bn:00007309n")]).
ulr("Commerce_sell",[role("Seller",mary,"bn:00046516n"),
                     role("Goods",watch,"bn:00077172n")]).
\end{verbatim}

\subsubsection{ULR for Factual Sentences with Adnominal Clauses}
An adnominal clause always describes the nominal word it modifies. This means that an adnominal clause expresses a complete fact about the nominal word. In other words, the facts represented by adnominal clauses and by the main clause must both hold. Thus, the ULRs for clauses must be in conjunction.

For example, the sentence ``\textit{[Mary bought a car]\textsubscript{main} [made in the count-
ry]\textsubscript{adnominal} [that John lives in]\textsubscript{adnominal}}''
has a main clause and two adnominal clauses, one modifying the word ``\textit{car}'' and the other the word
``\textit{country}.''
The ULR then is given below:

\begin{verbatim}
ulr("Commerce_buy",[role("Buyer","Mary","bn:00046516n"),
                    role("Goods",car,"bn:00007309n")]).
ulr(Manufacturing",[role("Product",car,"bn:00007309n"),
                    role("Place",country,"bn:00023236n")]).
ulr(Residence",[role("Resident","John","bn:00046516n"),
                role("Location",country,"bn:00023236n")]).
\end{verbatim}

%% file: 3_kalmfl/4_eval.tex
\section{Evaluation of \texorpdfstring{\KALMF}\xspace}\label{sec:kalmfl-eval}

We use four datasets to demonstrate the high performance of \KALMF as a knowledge authoring machine for factual English.

\subsection{Datasets}
\label{sec:kalmfl-datasets}
\begin{itemize}
\item
\textbf{CNLD.} \cite{gao2018knowledge} uses CNL sentences, largely inspired by FrameNet, to evaluate the original KALM. We call this dataset the CNL Dataset (CNLD). CNLD contains 250 short CNL sentences in present tense, such as ``\textit{Kate purchases a house}.'' This dataset is captured via 50 logical frames and 317 LVPs constructed from 213 training sentences.

\item
\textbf{CNLDM.} This dataset is obtained from CNLD by changing the voice of some sentences from active to passive and vice versa. In addition, some sentences are changed to past or future tense.
Thus, CNLDM contains sentences like ``\textit{A house was purchased by Kate}," with mixed voice and tense. Our evaluation uses the same LVPs as in CNLD.
  
\item
\textbf{MetaQA.} This dataset~\cite{zhang2017variational} has queries that use complex adnominal clauses. These queries neatly fall into several different templates. Within each template, the queries differ only in the entity names.
Also, all named entities are pre-annotated. For example, the queries ``\textit{who directed the movies written by [Thomas Ian Griffith]}'' and ``\textit{who directed the movies written by [Frank De Felitta]}'' belong to the same template ``\textit{who directed the movies written by [MASK]},'' where \textit{[MASK]} is a placeholder for pre-annotated named entities.
In this evaluation, we use 2- and 3-hop templates directly instead of the original queries, because with named entities annotated, different queries that fall into a same template have exactly the same \MS~parse.  Only 3 frames are needed to represent the semantics of all such queries: \texttt{Movie}, \texttt{Inequality}, and \texttt{Coop}(eration). Acting as knowledge engineers, we designed 85 training sentences and used the approach of~\cite{gao2019querying} to understand 2- and 3-hop queries.

\item
\textbf{NLD.} NLD uses part of the dataset from FrameNet. NLD includes 250 sentences which look like: ``\textit{GDA has purchased the site from Laing Homes and plans are being prepared for an 80 million dollar mixed development for business, media and leisure activities}'', which is the original sentence of the CNLD sentence ``\textit{Kate purchases a house}". NLD sentences have much more complicated structures that go beyond factual sentences. Besides factual parts, most of the NLD sentences have non-factual parts that are not usable for knowledge acquisition. In view of this, we ignore all the non-factual parts in NLD sentences. Another approach could be highlighting the non-factual parts and letting the user correct them or eliminate them.
\end{itemize}

For the original KALM, all these datasets have to be manually modified to eliminate future/past tense, 
to put adnominal clauses in a certain canonical form, 
restrict the vocabulary for the controlled natural language parser, 
particle verbs, appositives, and compound nouns also had to be manually modified.
In \KALMF, all this is done automatically, and therefore, it can handle a much bigger share of natural language sentences.

\subsection{Comparison Systems}\label{sec:kalmfl-compsys}
We compare \KALMF with the original KALM as well as three other
frame-based parsers: SEMAFOR \cite{das2014frame}, SLING~\cite{ringgaard2017sling}, and OpenSesame~\cite{swayamdipta2017frame}.
SEMAFOR and SLING have been previously shown to be inaccurate in~\cite{gao2018high}, so we will not repeat these findings and instead focus on other comparison systems listed below.

\begin{itemize}
    \item OpenSesame: a three-staged pipeline involving target (i.e., LU) identification, frame identification, and argument (i.e., role-filler) identification---each stage is essentially a neural network trained independently of the others.

    \item DrKIT \cite{dhingra2020differentiable}: a neural network model trained to answer MetaQA questions. DrKIT has state-of-the-art performance on the MetaQA dataset among neural models.
    Thus, in the following comparison, we only apply DrKIT to MetaQA and report the result based on our metrics described later.

    \item GPT \cite{brown2020language}: a large language model with excellent performance in a wide range of NLP tasks, which we have introduced in Section~\ref{sec:gpt}.
    GPT's few-shot learning enables models to learn from small amounts of data, making it useful for tasks without fine-tuning.
    By employing few-shot learning, we can harness GPT to help us with frame-based parsing and in-context word definition generation.
    Jumping ahead, we will see that GPT by itself is inferior to structured learning done by \KALMF.
    However, combined with embeddings, GPT can be used to speed up and improve accuracy of the role-filler disambiguation step in the KALM pipeline.
\end{itemize}

In our GPT experiments, we employ a two-step pipelined approach that involves both frame-based parsing and role-filler disambiguation.
We use the sentence ``\textit{Bob works for Apple}" to illustrate this pipeline.

To begin, we prompt GPT to obtain frame parses (defined in Section~\ref{sec:deploy}) for ``\textit{Bob works for Apple.}"
The prompt has three distinct components:

\begin{enumerate}
    \item Task description: guides GPT to carry out the \textit{frame-based parsing} task using the provided frame templates.
    \item Few-shot demonstration: exemplifies the task and formats the output. Including 2-3 examples of frame-based parsing suffices.
    \item Prompting: directs GPT to perform frame-based parsing on a specific sentence, such as ``\textit{Bob works for Apple.}"
\end{enumerate}

\noindent
Prompts~\ref{prompt:parsing} and \ref{prompt:parsing2} from Appendix~\ref{apdx:prompts} are example prompts dealing with two different ontologies.
For the CNLD, CNLDM, and NLD datasets, we use the example Prompt~\ref{prompt:parsing} to instruct GPT to perform frame-based parsing.
(GPT appears to have been trained on frame-based parses and such parsing works out of the box in most cases.)
For these three datasets, we employ the 50 logical frame templates mentioned in Section~\ref{sec:kalmfl-datasets} to substitute the \texttt{FRAME\_TEMPLATES} placeholder in Prompt~\ref{prompt:parsing}.
For the MetaQA dataset, we employ the example Prompt~\ref{prompt:parsing2}.
Compared to Prompt~\ref{prompt:parsing}, Prompt~\ref{prompt:parsing2} for MetaQA requires only 3 frame templates to substitute the \texttt{FRAME\_TEMPLATES} placeholder in Prompt~\ref{prompt:parsing2}, which results in 3 few-shot examples.
To perform frame-based parsing on the running example, we use Prompt~\ref{prompt:parsing} and substitute \texttt{SENTENCE} with ``\textit{Bob works for Apple.}"
In response, GPT generates this frame parse:
\begin{verbatim}
[p("Being_employed",[role("Employee","Bob"),
                     role("Employer","Apple")])]
\end{verbatim}

Once the frame-based parsing step is completed, we proceed with the \textit{role-filler disambiguation} step.
To this end, we prompt GPT to generate in-context definitions of the role-fillers extracted from the previous frame-based parsing step.
We define \textit{in-context definition} of a word as follows: when provided with a sentence and a specific word from that sentence, the in-context definition of that word refers to its meaning within that particular sentence. 
For the running example ``\textit{Bob works for Apple}," the in-context definition of the word "\textit{Apple}" should denote a technology company rather than a type of fruit.
The prompt for in-context definition generation, referred to as Prompt~\ref{prompt:disamb} in Appendix~\ref{apdx:prompts}, follows the same structure as Prompts~\ref{prompt:parsing}~and~\ref{prompt:parsing2}:

\begin{enumerate}
    \item Task description: guides GPT to carry out the \textit{in-context definition generation} task.
    \item Few-shot demonstration: exemplifies the task and formats the output.
    \item Prompting: directs GPT to generate the in-context definition of a role-filler with a specific (role-filler, sentence) pair,
    such as (``\textit{Apple}, ``\textit{Bob works for Apple}").
\end{enumerate}

\noindent
To perform role-filler disambiguation for the two role-fillers presented in the frame parse obtained from the GPT frame-based parsing, namely ``\textit{Bob}" and ``\textit{Apple}," we utilize the following approach.
First, we employ Prompt~\ref{prompt:disamb} by replacing \texttt{SENTENCE} with ``\textit{Bob works for Apple.}"
Next, we substitute \texttt{WORD} with ``\textit{Bob}" and ``\textit{Apple}" individually.
As a result, GPT provides the in-context definition of ``\textit{Bob}" as ``\textit{A male's name}," and the definition of ``\textit{Apple}" as ``\textit{An international technology company.}"

This two-step pipelined approach enables GPT to perform knowledge authoring tasks just like KALM and \KALMF and facilitates the comparisons across various accuracy metrics.
These metrics will be introduced next.

\subsection{Results}
\label{sec:kalmfl-results}

The evaluation is based on the following metrics:

\begin{enumerate}
    \item \textbf{Frame-level Accuracy}: 
    the ratio of sentences that (i) correctly trigger all the applicable frames, and (ii) do not trigger wrong frames.
    
    \item \textbf{Role-level Accuracy}: the ratio of sentences that (i) correctly trigger all the applicable frames with all roles correctly identified, and (ii) do not trigger wrong frames.
    
    \item \textbf{Synset-level Accuracy}: 
    the ratio of sentences that (i) correctly trigger all the applicable frames with all roles correctly identified and disambiguated, and (ii) do not trigger wrong frames.
    Note this metric applies only to KALM, \KALMF, and GPT; other systems do not attempt to give semantics with this level of accuracy.
\end{enumerate}

\begin{table}[htbp!]
\caption{\KALMF Result Comparisons}
\centering
\vspace{1mm}
\begin{tabular}{ccccccc}
\hline\hline
          & \multicolumn{3}{c}{CNLD}                     & \multicolumn{3}{c}{CNLDM}                    \\
          & F             & R             & S             & F             & R             & S             \\ 
\hline 
KALM       & 0.99          & 0.99          & 0.97          & --           & --           & --           \\
OpenSesame & 0.61          & 0.17          & --           & 0.59          & 0.11          & --           \\
\KALMF      & \textbf{0.99} & \textbf{0.99} & \textbf{0.97} & \textbf{0.99} & \textbf{0.99} & \textbf{0.97} \\
DrKIT      & -- & -- & -- & -- & -- & -- \\
GPT       & 0.95 & 0.81 & 0.81 & 0.93 & 0.80 & 0.80 \\
\hline\hline 
          &\multicolumn{3}{c}{MetaQA}          & \multicolumn{3}{c}{NLD}                       \\
          & F             & R             & S             & F             & R             & S  \\ 
\hline 
KALM       & \textbf{1.00} & \textbf{1.00} & \textbf{1.00} & --           & --           & --           \\
OpenSesame & 0.49          & 0.00          & --  & 0.56          & 0.12          & --           \\
\KALMF      & 0.95          & 0.95           &  0.95  & \textbf{0.99} & \textbf{0.98} & \textbf{0.95} \\
DrKIT      & --          & 0.88          & -- & -- & -- & -- \\
GPT       & 0.60 & 0.57 & 0.57 & 0.75 & 0.53 & 0.53 \\
\hline\hline
\end{tabular}
\label{tab:kalmfl-results}
\end{table}

Results for the different levels of accuracy values are presented in Table~\ref{tab:kalmfl-results}. In the table, F, R, and S refer to the frame, role, and synset-level accuracy values, respectively.
The results reported in the literature for DrKIT \cite{dhingra2020differentiable} can be interpreted as pertaining to the role-level accuracy metric.
The results in the table are summarized below.

\begin{itemize}
    \item CNLD: This dataset has only CNL sentences and \KALMF~achieves the same high accuracy values as the original KALM in all metric levels.
    OpenSesame's 0.61 frame-level accuracy value shows that it has difficulty even recognizing correct frames.
    GPT does reasonably well at finding correct frame names from provided lists of frames.
    However, when it comes to role-level accuracy, GPT achieves the score of only 0.81.
    This level of accuracy leads to a larger than before amount of incorrect factual knowledge, which leads to errors in subsequent reasoning and question answering tasks.
    
    \item CNLDM: Perturbation of the tenses and voices of CNLD sentences took this dataset
    outside of Attempto's APE CNL, thus the original
    KALM cannot handle some of the CNLDM sentences even though the meaning of these sentences did not change.
    Both OpenSesame and GPT show similar levels of accuracy here as they do on the CNLD dataset.
    
    \item MetaQA: KALM performs perfectly, but only after changing the sentences so they comply with the ACE CNL. In contrast, \KALMF gets
    the 0.95 synset accuracy value even without any preprocessing.
    OpenSesame fails on MetaQA with 0 role-level accuracy value---probably because it was never trained on the movie domain.
    In the comparison between DrKIT and \KALMF, DrKIT~\cite{dhingra2020differentiable} achieved 0.871 and 0.876 accuracy values on 2- and 3-hop query answering respectively.
    For \KALMF, 333 out of 350 correctly parsed templates cover 128,784 2-hop queries and 119,923 3-hop queries, which results in 0.962 and 0.933 accuracy values on 2- and 3-hop query answering and outperforms DrKIT.
    \KALMF also outperforms GPT few-shot learning on the MetaQA dataset by a large margin.
    The difficulty for GPT was to accurately identify frames in a MetaQA query, often resulting in a notable decrease in frame-level accuracy compared to \KALMF.
    
    \item NLD: The original KALM fails since this dataset breaks the CNL restrictions on the input language. 
    In contrast, \KALMF does well and easily outperforms OpenSesame, especially when it comes to the handling of adnominal clauses.
    \KALMF also outperforms GPT even though the sentences in NLD are much less restricted.
\end{itemize}

It is surprising that OpenSesame's role-level accuracy values are so low on the three FrameNet-related datasets. Error analysis shows that even for simple CNLD sentences like ``\textit{Mary buys a laptop},'' OpenSesame has hard time extracting all roles. For instance, ``\textit{laptop}" is not extracted as a role-filler for the role \texttt{Goods}.
Additionally, when considering all datasets, structured learning in \KALMF proves superior to neural-network-based few-shot learning used with GPT.

%% file: 4_kalmra/0_main.tex
\chapter{\texorpdfstring{\KALMR:}\xspace\xspace From Facts to Rules and Actions}\label{chap:5:kalmra}

The proposed \KALMF system in Chapter \ref{chap:4:kalmfl} successfully lifted the CNL restrictions to a great extent.
However, \KALMF still has limitations in representing certain types of knowledge, such as authoring rules for multi-step reasoning or understanding actions with timestamps.
Thus, in this chapter, our focus shifts towards utilizing \emph{factual sentences} to author more complex knowledge including rules and actions.
Since we want to be able to handle disjunctive information required by some of the bAbI tasks, we made a decision to switch from XSB which was used in KALM to an ASP-based system DLV \cite{leone2006dlv} that can handle disjunction in the rule heads.
Thus, the syntax of the ULR, i.e., the logical statements produced by \KALMR, follows that of DLV.
A number of examples inspired by the UTI guidelines and bAbI Tasks are used in this section to illustrate the workings of \KALMR.

\input{4_kalmra/1_rules}

\input{4_kalmra/2_actions}

\input{4_kalmra/3_eval}

%% file: 4_kalmra/1_rules.tex
\section{Authoring of \texorpdfstring{\KALMR}\xspace Rules}

Rules are important to KRR systems because they enable multi-step logical inferences needed for real-world tasks, such as diagnosis, planning, and decision-making.
Here we address the problem of rule authoring.

\subsection{Enhancements for Representation of Facts}

First, we discuss the representation of disjunction, conjunction, negation, and coreference, which is not covered in \KALMF.

\subsubsection{Conjunction and Disjunction}
The \KALMR system prohibits the use of a mixture of conjunction and disjunction within a single factual sentence to prevent ambiguous expressions such as ``\textit{Mary wants to have a sandwich or a salad and a drink.}"
To represent a factual sentence with homogeneous conjunction or disjunction,
the system first parses the sentence into a set of component ULRs.

For conjunction, 
\KALMR uses this set of ULRs as the final representation.
For disjunction, the component ULRs are assembled into a single disjunctive ULR using DLV's disjunction \texttt{v} as shown in
Example \ref{exmp:fact-coord}.

\begin{example}
\label{exmp:fact-coord}
The factual sentence with conjunction ``\textit{Daniel administers a parenteral and an oral antimicrobial therapy for Mary}'' is represented as the following set of ULRs:

\begin{verbatim}
ulr("Cure",[role("Doctor","Daniel"),role("Patient","Mary"),
            role("Therapy",antimicrobial),
            role("Method",parenteral)]).
ulr("Cure",[role("Doctor","Daniel"),role("Patient","Mary"),
            role("Therapy",antimicrobial),role("Method",oral)]).
doctor("Daniel"). patient("Mary"). therapy(antimicrobial).
method(parenteral). method(oral).
\end{verbatim}

\noindent
where the predicates \texttt{doctor}, \texttt{patient}, \texttt{therapy}, and \texttt{method} define the domains for the roles.
These domain predicates will be omitted in the rest of the paper, for brevity.

The disjunctive factual sentence 
``\textit{Daniel administers a parenteral or an oral antimicrobial therapy for Mary}" is represented as the following ULR:

\begin{verbatim}
ulr("Cure",[role("Doctor","Daniel"),role("Patient","Mary"),
            role("Therapy",antimicrobial),
            role("Route",parenteral)])
v
ulr("Cure",[role("Doctor","Daniel"),role("Patient","Mary"),
            role("Therapy",antimicrobial),role("Route",oral)]).
\end{verbatim}
\end{example}

\subsubsection{Negation}
The \KALMR system supports \textit{explicit negation} through the use of the negative words ``\textit{not}" and ``\textit{no}".
Such sentences are captured by appending the suffix ``\texttt{\_not}" to the name of the frame triggered by this sentence.

\begin{example}
\label{exmp:fact-negation}
The explicitly negated factual sentence ``\textit{Daniel's patient Mary does not have UTI}'' is represented by

\begin{verbatim}
ulr("Medical_issue_not",[role("Doctor","Daniel"),
                         role("Patient","Mary"),
                         role("Ailment","UTI")]).
\end{verbatim}
\end{example}

\subsubsection{Coreference}
Coreference occurs when a word or a phrase refers to something that is mentioned earlier in the text.
Without coreference resolution, one gets unresolved references to unknown entities in ULRs.
To address this issue, \KALMR uses 
a coreference resolution tool
neuralcoref,\footnote{\url{https://github.com/huggingface/neuralcoref}}
which
identifies and replaces coreferences with the corresponding entities from the preceding text.

\begin{example}
\label{exmp:4}
The factual sentences ``\textit{Daniel's patient Mary has UTI. He administers an antimicrobial therapy for her.}"
are turned into

\begin{verbatim}
ulr("Medical_issue",[role("Doctor","Daniel"),
                     role("Patient","Mary"),
                     role("Ailment","UTI")]).
ulr("Cure",[role("Doctor","Daniel"),role("Patient","Mary"),
            role("Therapy",antimicrobial)]).
\end{verbatim}
\end{example}
\noindent
where the second ULR uses entities \texttt{"Daniel"} and \texttt{"Mary"}  instead of the pronouns ``\textit{he}" and ``\textit{she}."

\subsection{\texorpdfstring{\KALMR}\xspace Rules}

Rules in \KALMR are expressed in a much more restricted syntax
compared to facts since, for knowledge authoring purposes,
humans have little difficulty learning and complying with the restrictions. Moreover, since variables play such a key role in rules, complex coreferences must be specified unambiguously.
All this makes writing rules in a natural language into a very cumbersome, error-prone, and ambiguity-prone task
compared to the restricted syntax below.

\begin{definition}
\label{def:rules}
A \textit{rule} in \KALMR is an if-then statement of the form ``If $P_1$, $P_2$, ..., and $P_n$, then $C_1$, $C_2$, ..., or $C_m$", where
\begin{enumerate}
    \item each $P_i$ ($i=1..n$) is a factual sentence without disjunction;
    \item each $C_j$ ($j=1..m$) is a factual sentence without conjunction;
    \item variables in $C_j$ ($j=1..m$) must use the \textit{explicitly typed syntax} \cite{gao2018high} and must appear in at least one of the $P_i$ ($i=1..n$). E.g., in the rule ``\textit{If Mary goes to the hospital, then \$doctor sees Mary}", the explicitly typed variable \textit{\$doctor} appears in the conclusion without appearing in the premise, which is prohibited.
    Instead, the rule author must provide some information about the doctor in a rule premise (e.g., ``and she has an appointment with \textit{\$doctor}"). This corresponds to the well-known ``rule safety" rule in logic programming.
    \item variables that refer to the same thing must have the same name. E.g., in the rule ``\textit{If \$patient is sick, then \$patient goes to see a doctor}", the two \textit{\$patient} variables are intended to refer to the same person and thus have the same name.
\end{enumerate}
\end{definition}
\noindent

Here are some examples of rules in \KALMR.

\begin{example}
\label{exmp:6}
The \KALMR rule ``If \textit{\$doctor's \$patient is a young child and has an unexplained fever}, then \textit{\$doctor assesses \$patient's degree of toxicity or dehydration}" is represented as follows:

\begin{verbatim}
ulr("Assessing",[role("Doctor",Doctor),role("Patient",Patient),
                   role("Item",toxicity)])
v
ulr("Assessing",[role("Doctor",Doctor),role("Patient",Patient),
                 role("Item",dehydration)]) :-
                 
    ulr("People_by_age",[role("Person",Patient),
                         role("Type",child)]),
    ulr("Medical_issues",[role("Doctor",Doctor),
                          role("Patient",Patient),
                          role("Ailment",fever),
                          role("Cause",unexplained)]).
\end{verbatim}
\end{example}

\KALMR supports two types of negation in rules: \textit{explicit negation} \cite{gelfond1991classical} and \textit{negation as failure} \cite{gelfond1988stable} (with the stable model semantics). 
The former allows users to specify explicitly known negative factual information while the latter lets one derive negative information from the lack of positive information.
Explicit negation in rules is handled the same way as in fact representation.
Negation as failure must be indicated by the rule author through the idiom ``\textit{not provable}", which is then converted into the predicate \texttt{not/1}.
The idiom ``\textit{not provable}" is prohibited in rule heads.

\begin{example}
\label{exmp:10}
The \KALMR rule ``If \textit{not provable \$doctor does not administer \$therapy for \$patient}, then \textit{\$patient undergoes \$therapy from \$doctor}"is represented as follows:

\begin{verbatim}
ulr("Undergoing",[role("Doctor",Doctor),role("Patient",Patient),
                  role("Therapy",Therapy)]) :-
    not frame("Cure_not",[role("Doctor",Doctor),
                          role("Patient",Patient),
                          role("Therapy",Therapy)]),
    patient(Patient), doctor(Doctor), therapy(Therapy).
\end{verbatim}
\end{example}
\noindent
where \texttt{patient/1}, \texttt{doctor/1}, and \texttt{therapy/1} are domain predicates that ensure that variables that appear under negation have well-defined domains.

\subsection{Queries and Answers}

Queries in \KALMR must be in factual English and end with a question mark.
\KALMR translates both \textit{Wh}-variables and explicitly typed variables into the corresponding DLV variables.
Example \ref{exmp:11} shows how \KALMR represents a query with variables.

\begin{example}
\label{exmp:11}
The query ``\textit{Who undergoes \$therapy?}" has the following ULR:
\begin{verbatim}
ulr("Undergoing",[role("Patient",Who),
                  role("Therapy",Therapy)])?
\end{verbatim}
\end{example}

\KALMR then invokes the DLV reasoner to compute query answers.
DLV has two inference modes: \textit{brave reasoning} and
\textit{cautious reasoning}.
In brave reasoning, a query returns answers that are true in \textit{at least one} model of the program and 
cautious reasoning returns the answers that are true in \textit{all} models. Users are free to choose either mode.

\begin{example}
\label{exmp:12}
For instance, if the underlying information contains only this single fact
\begin{verbatim}
{ulr("Undergoing",[role("Patient","Mary"),
                   role("Therapy",mental)]).}
\end{verbatim}
\noindent
then there is only one model and both modes return the same result:
\begin{verbatim}
{Who="Mary",Therapy=mental}
\end{verbatim}

\noindent
In case of
``\textit{Mary or Bob undergoes a mental therapy}", two models are are computed:
\begin{verbatim}
{ulr("Undergoing",[role("Patient","Mary"),
                   role("Therapy",antimicrobial)]).}
{ulr("Undergoing",[role("Patient","Bob"),
                   role("Therapy",antimicrobial)]).}
\end{verbatim}
\noindent
In the cautious mode there would be no answers while the brave mode yields two:
\begin{verbatim}
{Who="Mary",Therapy=mental}
{Who="Bob",Therapy=mental}
\end{verbatim}
\end{example}

%% file: 4_kalmra/2_actions.tex
\section{Authoring and Reasoning with Actions}

Time-independent facts and rules discussed earlier are knowledge that persists over time.
In contrast, actions are momentary occurrences of events that change the underlying knowledge, so actions are associated with timestamps.
Dealing with actions and their effects, also known as \textit{fluents}, requires an understanding of the passage of time.
The following discussion of authoring and reasoning with actions will be in the SEC framework.

\subsection{Actions}

\KALMR allows users to state actions using factual English sentences and then formalizes actions as temporal database facts using SEC discussed in Example~\ref{exmp:ec-fact} of Section \ref{sec:ec}.

\subsection{Fluent Initiation and Termination}

Reasoning based on SEC requires the knowledge of fluent initiation and termination.
This information is part of the commonsense and domain knowledge supplied by knowledge engineers and domain experts via high-level fluent initiation and termination statements (Definition \ref{def:rules}) and \KALMR translates them into facts and rules that involve the predicates \texttt{initiates/2} and \texttt{terminates/2} used by Event Calculus. 
Knowledge engineers supply the commonsense part of these statements and domain experts supply the domain-specific part. 

\begin{definition}\label{def:iandt}
    A \emph{fluent initiation statement} in \KALMR has the form ``\textit{$A$/$F_{obs}$ initiates $F_{init}$}" and a \emph{fluent termination statement} in \KALMR has the form ``\textit{$A$/$F_{obs}$ terminates $F_{term}$}," where

\begin{enumerate}
    \item action $A$, observed fluent $F_{obs}$, initiated fluent $F_{init}$, and terminated fluent $F_{term}$ are factual sentences without conjunction or disjunction;
    \item variables in $F_{init}$ use explicitly typed syntax and must appear in $A$ (or in $F_{obs}$ when a fluent is observed) to avoid unbound variables in initiated fluents;
    \item variables that refer to the same thing must have the same name.
\end{enumerate}
\end{definition}

Example \ref{exmp:5} shows how \KALMR represents fluent initiation and termination.

\begin{example}
\label{exmp:5}
The commonsense initiation statement ``\textit{\$person travels to \$place initiates \$person is located in \$place}" would be created by a knowledge engineer and translated by \KALMR as the following rule:
\begin{verbatim}
initiates(ulr("Travel",[role("Person",Person),
                        role("Place",Place)]),
          ulr("Located",[role("Entity",Person),
                         role("Location",Place)])) :-
    person(Person), place(Place).
\end{verbatim}
\noindent
Here \texttt{person/1} and \texttt{place/1} are used to guarantee rule safety.
Since, any object can be in one place only at any given time, we have a commonsense termination statement ``\textit{\$person travels to \$place1 terminates \$person is located in \$place2.}"
This statement would also be created by knowledge engineers
and translated by \KALMR as follows:

\begin{verbatim}
terminates(ulr("Travel",[role("Person",Person),
                         role("Place",Place)]),
           ulr("Located",[role("Entity",Person),
                          role("Location",Place2)])) :-
    person(Person), place(Place),
    entity(Person), location(Place2), Place != Place2.
\end{verbatim}
\end{example}

\subsection{Time-Related Rules and Queries}

\KALMR also enhances rules by incorporating temporal information, allowing the inference of new knowledge under the SEC framework.
The process begins by requiring users to specify their domain knowledge on fluents in the form of rules described in Definition \ref{def:rules}.
Then \KALMR translates these rules into ULRs, with each premise and conclusion linked to a timestamp via the \texttt{holdsAt/2} predicate.
We call these rules \textit{time-related} because they enable reasoning with fluents containing temporal information.
Here is the form of time-related rules.
\begin{verbatim}
holdsAt(ULRC1,T) v ... v holdsAt(ULRCm,T) :-
    holdsAt(ULRP1,T), ..., holdsAt(ULRPn,T).
\end{verbatim}
\noindent
where all \texttt{holdsAt/2} terms share the same timestamp \texttt{T}, since the disjunction of conclusion ULRs \texttt{ULRC1}, ..., \texttt{ULRCm} holds immediately if all premise ULRs \texttt{ULRP1}, ..., \texttt{ULRPn} hold simultaneously at \texttt{T}.
Here is an example of time-related rules.

\begin{example}
\label{exmp:actionrules}
The fluent ``\textit{\$object is located in \$place}" holds at timestamp \texttt{t}, if both fluents ``\textit{\$person is holding \$object}" and ``\textit{\$person is located in \$place}" hold at \texttt{t}.
In this case, the user should represent it as a \KALMR rule ``If \textit{\$person is holding \$object}, and \textit{\$person is located in \$place}, then \textit{\$object is located in \$place}", which is further represented as a time-related rule by \KALMR as follows.
\begin{verbatim}
holdsAt(ulr("Located",[role("Entity",Object),
                       role("Place",Place)]),T) :-
    holdsAt(ulr("Holding",[role("Holder",Person),
                           role("Object",Object)]),T),
    holdsAt(ulr("Located",[role("Entity",Person),
                           role("Place",Place)]),T).
\end{verbatim}
\end{example}

In terms of queries, \KALMR also incorporates temporal information using \texttt{holdsAt/2}.
In this representation, the second argument of \texttt{holdsAt/2} is set to the highest value in the temporal domain extracted from the narrative.
For Example \ref{exmp:fact-coord}, a time-related query can be represented as
\begin{verbatim}
holdsAt(ULRQ,3)?
\end{verbatim}
where \texttt{ULRQ} is the ULR of the query and \texttt{3} is the timestamp that exceeds all the explicitly given timestamps.

%% file: 4_kalmra/3_eval.tex
\section{Evaluation of \texorpdfstring{\KALMR}\xspace}

In this section, we assess the effectiveness of \KALMR-based knowledge authoring using two test suites, the clinical UTI guidelines \cite{committee1999practice} and the bAbI Tasks \cite{weston2015towards}.

\subsection{Evaluation of Authoring of Rules}

The clinical UTI guidelines is a set of therapeutic recommendations for the initial Urinary Tract Infection (UTI) in febrile infants and young children.
The original version in English was rewritten into the ACE 
CNL \cite{shiffman2009writing} for the assessment of ACE's expressiveness.
We rewrite the original English version into factual English, as shown in Appendix \ref{apdx:uti}.
This new version has a significant number of rules with disjunctive heads, as is common in the real-world medical domain.

The experimental results show that \KALMR is able to convert the UTI guidelines document into ULRs with 100\% accuracy.

\subsection{Evaluation of Authoring and Reasoning with Actions}

\begin{figure}[htbp!]
    \centering
    \includegraphics[scale=0.32]{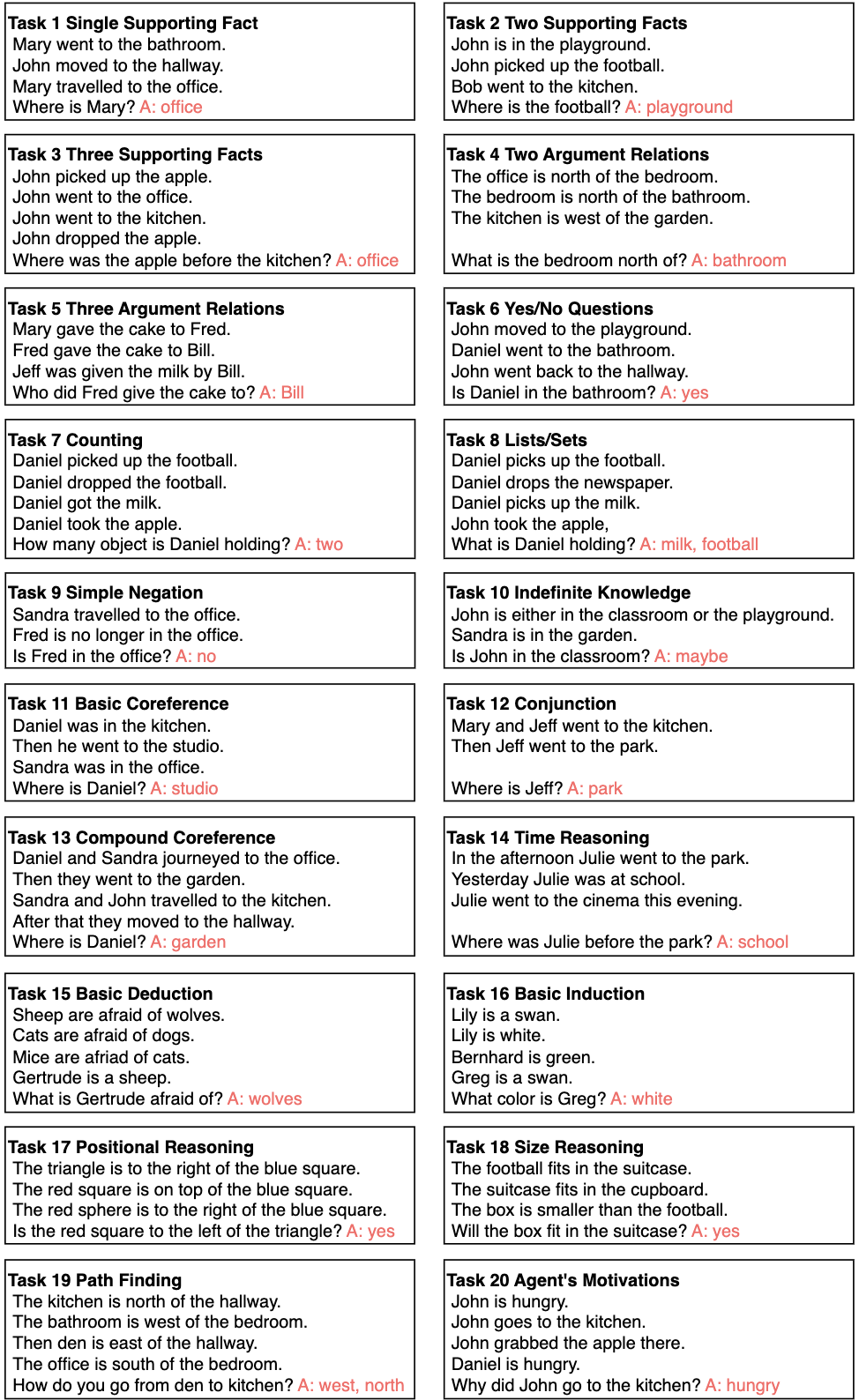}
    \caption{Sample narratives and questions from 20 bAbI tasks}
    \label{fig:babi}
\end{figure}

In this section, we evaluate \KALMR's capability of authoring and reasoning with actions, using the bAbI Tasks \cite{weston2015towards}.
The 20 bAbI tasks were designed to evaluate a system's capacity for natural language understanding, especially when it comes to actions.
They cover a range of aspects, such as 
moving objects (tasks 1-6),
counting (task 7),
positional reasoning (task 17), and 
path finding (task 19). 
Each task provides a set of training and test data, where each data point consists of a textual narrative, a question about the narrative, and the correct answer.
Fig.~\ref{fig:babi} presents 20 data points from 20 bAbI tasks respectively.
We used the test data for evaluation. Each task in the test data has 1,000 data points.

The comparison systems in this evaluation include a state-of-the-art neural model on bAbI Tasks, STM \cite{le2020self};
an approach based on inductive learning and logic programming \cite{mitra2016addressing} that we call LPA here;
and a recent sensation, ChatGPT.
The comparison with STM and ILA results are displayed in Table \ref{tab:kalmra-results}, where ``\#I\&T" denotes the number of user-given initiation and termination statements (Definition \ref{def:iandt}) used to specify each particular task in \KALMR. 
The table shows that \KALMR achieves accuracy comparable to
STM and ILA.

\begin{table}[htbp]
\caption{Result Comparisons}
\vspace{2mm}
\begin{tabular}{l|c|c|ccc}
\hline\hline
        & STM  & ILA & \multicolumn{3}{c}{\KALMR}          \\ \hline
TASK   & Acc.  & Acc.                & \#I\&T & \#Rules & Acc.  \\ \hline
1 Single Supporting Fact       & 100  & 100                & 2          & 0              & 100  \\
2 Two Supporting Facts      & 99.79 & 100                & 4          & 1              & 100  \\
3 Three Supporting Facts      & 97.87 & 100                & 4          & 1              & 100  \\
4 Two Argument Relations      & 100  & 100                & 4          & 0              & 100  \\
5 Three Argument Relations      & 99.43 & 100                & 4          & 0              & 100 \\
6 Yes/No Questions      & 100  & 100                & 2          & 0              & 100  \\
7 Counting      & 99.19 & 100                & 4          & 0              & 100  \\
8 Lists/Sets      & 99.88 & 100                & 4          & 0              & 100  \\
9 Simple Negation      & 100  & 100                & 4          & 0              & 100  \\
10 Indefinite Knowledge     & 99.97 & 100                & 2          & 0              & 100    \\
11 Basic Coreference     & 99.99 & 100                & 2          & 0              & 100  \\
12 Conjunction     & 99.96  & 100                & 2          & 0              & 100  \\
13 Compound Coreference     & 99.99 & 100                & 2          & 0              & 93.1 \\
14 Time Reasoning     & 99.84  & 100                & 2          & 0              & 100  \\
15 Basic Deduction     & 100  & 100                & 0          & 1              & 100  \\
16 Basic Induction     & 99.71 & 93.6               & 2          & 1              & 93.6 \\
17 Positional Reasoning     & 98.82 & 100                & 8          & 20             & 100  \\
18 Size Reasoning     & 99.73 & 100                & 0          & 1              & 100  \\
19 Path Finding     & 97.94 & 100                & 12          & 4              & 100  \\
20 Agent's Motivations     & 100  & 100                & 5          & 6              & 100  \\ \hline
Average & 99.61 & 99.68              & 3.45        & 1.75            & 99.34\\ \hline\hline
\end{tabular}
\label{tab:kalmra-results}
\end{table}

ChatGPT has shown impressive ability to give correct answers
for some manually-entered bAbI tasks even though (we assume) it was not trained on that data set.
However, it quickly became clear that it has no robust semantic model behind its impressive performance and it makes many mistakes on bAbI Tasks. 
The recent (Jan 30, 2023) update of ChatGPT fixed some of the cases, while still not being able to handle slight perturbations of those cases.
Three such errors are shown in Table \ref{tab:chatgpt},
which highlights the need for authoring approaches, like \KALMR, which are based on robust semantic models.

\begin{table}[hbtp]
\centering
\caption{ChatGPT Error Cases}
\vspace{2mm}
\begin{tabular}{l|l}
\hline\hline
\multicolumn{1}{c|}{\textbf{Task 2}}             & \multicolumn{1}{c}{\textbf{Task 17}}              \\
\multicolumn{1}{c|}{\textbf{2 Supporting Facts}} & \multicolumn{1}{c}{\textbf{Positional Reasoning}} \\ \hline
Mary went to the kitchen.                        & The red square is below the blue square.             \\
Mary got the apple.                              & The red square is left of the pink rectangle.        \\
Mary got the ball.                               &                                                      \\
Mary got the book.                               &                                                     \\
Mary went to the bedroom.                        &                                                    \\
Mary went to the garden.                         &                                                   \\ 
Mary dropped the book.                           &                                                   \\ \hline
Q: Where is the apple?                           & Q: Is the blue square below                       \\
                                                 & \hspace{5.7mm}the pink rectangle?                 \\ \hline
ChatGPT: ... not specified                       & ChatGPT: ... not specified                        \\
Correct: \hspace{3.6mm}garden                    & Correct: \hspace{3.6mm}no                         \\ \hline\hline
\multicolumn{2}{c}{\textbf{Task 19}}      \\
\multicolumn{2}{c}{\textbf{Path Finding}} \\ \hline
\multicolumn{2}{l}{The garden is west of the hallway.}        \\
\multicolumn{2}{l}{The kitchen is west of the garden.}        \\
\multicolumn{2}{l}{The garden is north of the bathroom.}       \\
\multicolumn{2}{l}{The bedroom is east of the bathroom.}      \\
\multicolumn{2}{l}{The hallway is west of the office.}        \\ \hline
\multicolumn{2}{l}{Q: How do you go from the bathroom to the hallway?}                  \\ \hline
\multicolumn{2}{l}{ChatGPT: ... east..., ... south...}        \\
\multicolumn{2}{l}{Correct: \hspace{3.6mm}east, north}         \\ \hline\hline
\end{tabular}
\label{tab:chatgpt}
\end{table}

As to \KALMR, it does not achieve 100\% correctness on Tasks 13 (Compound Coreference) and 16 (Basic Induction).
In Task 13, the quality of \KALMR's coreference resolution is entirely dependent on the output of neuralcoref, the coreference resolver we used. As this technology improves, so will \KALMR.
Task 16 requires the use of the induction principles adopted by bAbI tasks, some of which are questionable.
For instance,
in Case 2 of Table~\ref{tab:errors}, the color is determined by the maximum frequency of that type, whereas in Case 3, the latest evidence determines the color.
Both of these principles are too simplistic and, worse, contradict each other.

\begin{table}[hbtp]
\centering
\caption{\KALMR Error Cases}
\vspace{2mm}
\begin{tabular}{l|l}
\hline\hline
\multicolumn{2}{c}{\textbf{Task 13 Compound Coreferences}}                        \\\hline
\multicolumn{2}{c}{\textbf{Case1}} \\\hline
\multicolumn{2}{l}{Mary and Sandra went back to the bedroom.}                                    \\
\multicolumn{2}{l}{Then they moved to the kitchen.}                                              \\
\multicolumn{2}{l}{Sandra and Daniel went back to the bathroom.}                                  \\
\multicolumn{2}{l}{Then they went to the office.}                                                  \\\hline
\multicolumn{2}{l}{Q: Where is Daniel?}                                                  \\ \hline
\multicolumn{2}{l}{\KALMR: bathroom}                                                             \\
\multicolumn{2}{l}{bAbI Correct: office}                                                          \\ \hline\hline
\multicolumn{2}{c}{\textbf{Task 16 Basic Induction}}                                             \\\hline
\multicolumn{1}{c|}{\textbf{Case 2}}        & \multicolumn{1}{c}{\textbf{Case 3}}               \\ \hline
Brian is a swan.       & Berhnard is a rhino.                \\
Greg is a swan.        & Brian is a rhino.                    \\
Julius is a swan.      & Bernhard is white.                   \\
{Greg is gray.}          & Brian is white.                      \\
Julius is gray.        & Lily is a lion.                      \\
Bernhard is a lion.    & Lily is yellow.                      \\
Lily is a swan.        & Greg is a rhino.                     \\
Bernhard is green.     & Greg is green.                       \\
Brian is white.        & Julius is a rhino.                   \\\hline
Q: What color is Lily? & Q: What color is Julius?            \\ \hline
\KALMR: gray, white      & \KALMR: green, white               \\
bAbI Correct: gray          & bAbI Correct: green           \\\hline\hline
\end{tabular}
\label{tab:errors}
\end{table}

%% file: 5_disambiguation/0_main.tex
\chapter{Role-Filler Disambiguation Revisited}\label{chap:x:disam}
So far, we have introduced two novel extensions to the KALM framework, namely \KALMF and \KALMR, which have greatly expanded the range of the English sentences that can be used to express knowledge.
We have also identified a major limitation in the KALM-based approaches (the original KALM, \KALMF, and \KALMR)---runtime inefficiency that is a serious usability barrier for interactive applications. Our tests show that the main culprit here is the \emph{disambiguation} step, discussed in Section \ref{sec:deploy}.

Based on Formulas (\ref{formula:disamb}) and (\ref{formula:disamb-2}), we can infer that the calculation of $score^{rf}$ involves finding shortest paths in the BabelNet graph, which
necessitates a substantial amount of BabelNet querying.
As the search goes deeper, the amount of querying increases dramatically, which slows down the computation of $score^{cp}$ in Formula (\ref{formula:geo}).
Despite the various engineering tricks, such as bi-directional breadth-first search and multi-threading~\cite{gao2018high}, to speed up this step,
$score^{rf}$ calculations for a whole sentence still take up to 20 seconds (with Intel(R) Core(TM) i7-2600 CPU @ 3.40 GHz and 32 GB memory).
Moreover, BabelNet has abandoned some semantic features (like relevance weights on edges), preventing KALM from using the latest versions of BabelNet.
Considering all the above problems, we started looking into neural model approaches to computing $score^{rf}$ at a more reasonable cost.

\input{5_disambiguation/1_relbert}
\input{5_disambiguation/2_sbert}
\input{5_disambiguation/3_eval}

%% file: 5_disambiguation/1_relbert.tex
\section{RelBERT-Based Disambiguation}\label{sec:rbd}

In Section~\ref{sec:rep-learning}, we introduced RelBERT, a powerful relation encoder for relation-aware models.
In this section, we propose a novel approach that utilizes the relationship embeddings generated by RelBERT to improve the speed of role-filler disambiguation for KALM at the \emph{level of words}, without identifying the exact sense of each word.
In Section \ref{sec:rbd-word} we also show that GPT can be used to further refine this disambiguation to the level of word \emph{senses}.
We refer to the refined disambiguation approach as \textit{RelBERT-based disambiguation} (RBD).

\subsection{Disambiguation at Word Level}
We use Example~\ref{exmp:relbert} to illustrate RBD, the RelBERT-based disambiguation at word level.

\begin{example}
\label{exmp:relbert}
Suppose we have  the following training sentences
\begin{verbatim}
train("Mary buys a car for Jack","Commerce_buy","LU"=2,[],
      ["Buyer"=1+required,"Goods"=4+required,"Recipient"=6+optnl]).
train("Mary buys a car for her family","Commerce_buy","LU"=2,[],
      ["Buyer"=1+required,"Goods"=4+required,"Recipient"=7+optnl]).
train("Mary buys a car for 3,000 dollars","Commerce_buy","LU"=2,[],
      ["Buyer"=1+required,"Goods"=4+required,"Money"=7+optnl]).
\end{verbatim}
\noindent
and one test sentence ``\textit{Bob bought a piano for John.}"
We define \textit{training role-fillers} to refer to the filler words for the roles annotated in training sentences.
In our example, the role \texttt{Recipient} has two training role-fillers, ``\textit{Jack}" and ``\textit{family}", and the role \texttt{Money} has one, ``\textit{dollars}".
For instance, ``\emph{Jack}" is a training role-filler for \texttt{Recipient} because the first training sentence has ``\emph{Jack}" as word \#6 and the annotation of that sentence says that the \texttt{Recipient} role's filler is also word \#6.
Since training sentences are curated by knowledge engineers, training role-fillers are considered ``ideal'' examples of the fillers for the corresponding roles.

LVPs are learned from training sentences and test sentences
are parsed using these learned LVPs.
In our example, we get
two candidate parses (\ref{parse:relbert-1}) and (\ref{parse:relbert-2}), one interpreting the word ``\textit{John}" as a \texttt{Recipient} and the other as an amount of \texttt{Money}.

\end{example}
\begin{equation}
\label{parse:relbert-1}
\hspace*{-35mm}
\begin{aligned}
\verb|p("Commerce_buy",[role("Buyer","Bob"),|\\
\verb|role("Goods","piano"|&\verb|),|\\
\verb|role("Recipient","Jo|&\verb|hn")]).|
\end{aligned}
\end{equation}
\begin{equation}
\label{parse:relbert-2}
\hspace*{-42.5mm}
\begin{aligned}
\verb|p("Commerce_buy",[role("Buyer","Bob"),|\\
\verb|role("Goods","piano"|&\verb|),|\\
\verb|role("Money","John")|&\verb|]).|
\end{aligned}
\end{equation}

\noindent
The RBD approach employs two steps, outlined below, to determine the likeliest candidate parse for a given sentence.

\begin{enumerate}
    \item $score^{R}$ computation. RBD first uses Formula~(\ref{formula:relbert}) to calculate $score^{R}(rn,rf)$, where $rn$ is a role name, $rf$ is the role-filler for $rn$, $d$ is the number of all training role-fillers for $rn$, and $rf_i^{train}$ is the $i$-th training role-filler of $rn$.
    Function $emb_{rel}(\cdot)$ returns the RelBERT relationship embedding of a given word pair, and function $sim(\cdot)$ calculates cosine similarity of
    two relationship embeddings.
    Therefore, based on the the maximal similarity between $emb_{rel}(rn,rf)$ and $emb_{rel}(rn,rf_i^{train})$ ($i\in \{1,..,d\}$), we can reasonably estimate the suitability of $rf$ as a filler for role $rn$.
    \begin{equation}\label{formula:relbert}
        \begin{split}
            score^{R}(rn,rf)=\max_{i \in \{1, .., d\}}\Bigl(sim\bigl(&emb_{rel}(rn,rf),\\
            &emb_{rel}(rn,rf_i^{train})\bigl)\Bigl)
        \end{split}
    \end{equation}
    For example, considering the candidate parse (\ref{parse:relbert-1}), $score^{R}(\text{``Recipient"},\text{``John"})$ is determined by Formula~(\ref{formula:relbert}) as follows:
    $$
    \begin{aligned}
    &\max_{i \in \{1,2\}}\Bigl(\bigl\{sim\bigl(emb_{rel}(\text{``Recipient"},\text{``John"}),emb_{rel}(\text{``Recipient"},rf_i^{train})\bigl)\bigl\}\Bigl)\\
    &=\max\Bigl(sim\bigl(emb_{rel}(\text{``Recipient"},\text{``John"}),emb_{rel}(\text{``Recipient",``Jack"})\bigl),\\
    &\quad\quad\quad\;\;\; sim\bigl(emb_{rel}(\text{``Recipient"},\text{``John"}),emb_{rel}(\text{``Recipient",``family"})\bigl)\Bigl)\\
    &=max\big(\{0.909, 0.524\}\big)\\
    &=0.909
    \end{aligned}
    $$
    \noindent
    where ``\textit{Jack}" and ``\textit{family}" are the training role-fillers of the role \texttt{Recipient}.
    The high score of $0.909$ indicates that the relationship between \texttt{Recipient} and ``\textit{John}" is highly similar to the relationship between \texttt{Recipient} and ``\textit{Jack}". Thus, ``\textit{John}" is a good role-filler for the role \texttt{Recipient}.
    
    Now consider candidate parse (\ref{parse:relbert-2}).
    $score^{R}(\text{``Money"},\text{``John"})$ is determined 
    using Formula~(\ref{formula:relbert}) as follows:
    $$
    \begin{aligned}
    &\: \max_{i \in \{1\}}\Bigl(\bigl\{sim\bigl(emb_{rel}(\text{``Money"},\text{``John"}),emb_{rel}(\text{``Place"},rf_i^{train})\bigl)\bigl\}\Bigl)\\
    =&\: \max\Bigl(sim\bigl(emb_{rel}(\text{``Money"},\text{``John"}),emb_{rel}(\text{``Money",``dollar"})\bigl)\Bigl)\\
    =&\; sim\bigl(emb_{rel}(\text{``Money"},\text{``John"}),emb_{rel}(\text{``Money",``dollar"})\bigl)\\
    =&\; 0.292 < 0.909
    \end{aligned}
    $$
    \noindent
    This indicates that ``\textit{John}" is less suited
    as a filler for the role \texttt{Money} than for the role \texttt{Recipient}.

    \item $score^{P}$ computation. Formula~(\ref{formula:geo}) 
    on page \pageref{formula:geo}
    is used to calculate the corresponding $score^{P}$ for each candidate parse.
    As a result, candidate parses with low $score^{P}$ values can be eliminated, thus disambiguating the entire sentence.
    In the running example, candidate parse (\ref{parse:relbert-2}) is eliminated as $score^{R}(\text{``Money"},\text{``John"})$ makes crucial contribution to the overall $score^{P}$ value of the parse.
    Despite the fact that the two parses share the same $score^{R}(\text{``Buyer"},\text{``Bob"})$ and $score^{R}(\text{``Goods"},\text{``piano"})$,
    the lower $score^{R}(\text{``Money"},\text{``John"})$ in candidate parse (\ref{parse:relbert-2}) leads to the lower overall $score^{P}$ of candidate parse (\ref{parse:relbert-2}).
    Therefore, the test sentence ``\textit{Bob bought a piano for John}" is disambiguated as parse (\ref{parse:relbert-1}), which we call the \textit{likeliest} parse for the sentence.

\end{enumerate}

\subsection{Disambiguation at Word Sense Level}\label{sec:rbd-word}

In contrast to KALMs, achieving disambiguation at word level in RBD does not require word-sense level information.
Thus, RBD needs to incorporate a separate step for disambiguation at word sense level.
This involves generation of in-context definitions for all role-fillers present in the \textit{likeliest} parse (see the end of the previous subsection).
To accomplish this, RBD queries GPT with Prompt~\ref{prompt:disamb}, which can be found in Appendix~\ref{apdx:prompts}, page \pageref{prompt:disamb}.
In our example, to perform word-sense disambiguation for the role-filler ``\textit{piano}" in the likeliest parse \ref{parse:relbert-1}, we instantiate Prompt~\ref{prompt:disamb} as follows:

\begin{promptinstantiation}
\label{prompt-inst:1}
\begin{verbatim}
Given a SENTENCE, and a WORD in the SENTENCE,
generate the WORD's DEFINITION with only one short sentence.
============================================================
Example 1:
SENTENCE: "Kate charters a car."
WORD: "car"
DEFINITION of "car": A motor vehicle with four wheels.
Example 2:
SENTENCE: "Jackson protested against the government."
WORD: "Jackson"
DEFINITION of "Jackson": A male's name
Example 3:
SENTENCE: "Antoinette is staying with friends
           in Paris at present."
WORD: "friend"
DEFINITION of "friend": A person with whom one has a bond
                        of mutual affection and trust.
============================================================
Try:
SENTENCE: "Bob bought a piano for John."
WORD: "piano"
DEFINITION of "piano":
\end{verbatim}
\end{promptinstantiation}
\noindent
to which GPT would respond with ``\textit{A keyboard instrument that produces sound by striking strings with hammers.}"

In this manner, RBD proceeds to separately instantiate Prompt~\ref{prompt:disamb} with the remaining two role-fillers in parse~\ref{parse:relbert-1}, namely, ``\textit{Bob}" and ``\textit{John}".
The prompt instantiations for ``\textit{Bob}" and ``\textit{John}" are shown as follows in order:

\begin{promptinstantiation}
\label{prompt-inst:2}
\begin{verbatim}
Given a SENTENCE, and a WORD in the SENTENCE,
generate the WORD's DEFINITION with only one short sentence.
============================================================
Example 1:
SENTENCE: "Kate charters a car."
WORD: "car"
DEFINITION of "car": A motor vehicle with four wheels.
Example 2:
SENTENCE: "Jackson protested against the government."
WORD: "Jackson"
DEFINITION of "Jackson": A male's name
Example 3:
SENTENCE: "Antoinette is staying with friends
           in Paris at present."
WORD: "friend"
DEFINITION of "friend": A person with whom one has a bond
                        of mutual affection and trust.
============================================================
Try:
SENTENCE: "Bob bought a piano for John."
WORD: "Bob"
DEFINITION of "Bob":
\end{verbatim}
\end{promptinstantiation}
\noindent
to which GPT would respond with ``\textit{A male's name.}"

\begin{promptinstantiation}
\label{prompt-inst:3}
\begin{verbatim}
Given a SENTENCE, and a WORD in the SENTENCE,
generate the WORD's DEFINITION with only one short sentence.
============================================================
Example 1:
SENTENCE: "Kate charters a car."
WORD: "car"
DEFINITION of "car": A motor vehicle with four wheels.
Example 2:
SENTENCE: "Jackson protested against the government."
WORD: "Jackson"
DEFINITION of "Jackson": A male's name
Example 3:
SENTENCE: "Antoinette is staying with friends
           in Paris at present."
WORD: "friend"
DEFINITION of "friend": A person with whom one has a bond
                        of mutual affection and trust.
============================================================
Try:
SENTENCE: "Bob bought a piano for John."
WORD: "John"
DEFINITION of "John":
\end{verbatim}
\end{promptinstantiation}
\noindent
to which GPT would respond the same as in Instantiation of Prompt 3, No.~\ref{prompt-inst:2}, ``\textit{A male's name.}"


The big advantage of RBD over the earlier BabelNet-based disambiguation is that
RBD never queries BabelNet, which results in significantly shorter runtime for the disambiguation step.
On the other hand, BabelNet-based disambiguation is more precise in the sense that the meaning of words is determined down to the level of synsets,
while RBD uses textual definitions for word senses.
These definitions may change slightly between GPT sessions and also are not invariant when words are replaced by their synonyms.
For instance, 
in the sentences ``\textit{Bob bought a piano for John}" and ``\textit{Bob will buy a piano for John}," GPT might provide very different definitions for \emph{piano}: ``\textit{A keyboard instrument that produces sound by striking strings with hammers}" and ``\textit{A musical instrument that produces sound}," respectively.
These two definitions convey the same meaning but it is not easy to establish this automatically with high fidelity and efficiency.
Therefore, there is a need for a 
dedicated synset-like system for clustering the various in-context definitions produced by GPT.
Such a system would then enable a more direct comparison between RBD and the original KALM disambiguation process outlined in Section~\ref{sec:kalm}.
We leave this for future work.


%% file: 5_disambiguation/2_sbert.tex
\section{SBERT-Based Disambiguation}\label{sec:sbd}

In the previous section, we introduced an alternative approach to BabelNet-based disambiguation, with the primary goal of enhancing the speed of KALMs.
In this section, we propose a similar technique that can be applied to effectively achieve the same goal.
This technique uses GPT-generated in-context definitions and SBERT sentence embeddings, as discussed in Section~\ref{sec:sbert}.
We refer to this approach as \textit{SBERT-based disambiguation} (SBD).

We illustrate SBD through the same
Example~\ref{exmp:relbert} and the test sentence ``Bob bought a piano for John" that has two candidate frame parses, (\ref{parse:relbert-1}) and (\ref{parse:relbert-2}).
To disambiguate a test sentence, i.e., to find the likeliest parse,
SBD performs three steps:

\begin{enumerate}
    \item \emph{GPT in-context definition generation.}
    SBD starts by
    querying GPT with the prompts described in Appendix~\ref{apdx:prompts} (e.g., Instantiation of Prompt 3, No.~\ref{prompt-inst:1} above),
    to generate in-context definitions for \textit{all} role-fillers present in the candidate parses.
    In Example~\ref{exmp:relbert}, we have three role-fillers in total for both candidate parses~(\ref{parse:relbert-1}) and (\ref{parse:relbert-2}), namely, ``\textit{Bob}," ``\textit{piano}," and ``\textit{John}."
    Then, GPT produces in-context definitions for these role-fillers, just like in Section~\ref{sec:rbd-word}.
    Note that RBD performs word-sense disambiguation at the final step, SBD does it as the first step.

    \item \emph{The computation of} $score^{R}$.
    SBD uses Formula~(\ref{formula:sbert}), shown below, to calculate $score^{R}(rn,rf)$,
    where $rn$ is a role name, $rf$ is the role-filler for $rn$, $d$ is the total number of definitions of the role name $rn$.
    Recall that role name definitions in KALMs are synsets of BabelNet.
    For SBD, we could either use those same BabelNet synsets or sentences curated by knowledge engineers and used as role name definitions.
    In Formula~(\ref{formula:sbert}), $def_i(rn)$ 
    is the $i$-th definition of $R$ and $def_{gpt}(rf)$ is the GPT-generated in-context definition for $rf$.
    Also, $emb_{snt}(\cdot)$ encodes a given sentence into an SBERT sentence embedding and $sim(\cdot)$ calculates cosine similarity of two sentence embeddings.
    \begin{equation}\label{formula:sbert}
        \begin{split}
            score^{R}(rn,rf)=\max_{i \in \{1, .., d\}}\biggl(sim\Bigl(&emb_{snt}\bigl(def_i(rn)\bigl),\\
            &emb_{snt}\bigl(def_{gpt}(rf)\bigl)\Bigl)\biggl)
        \end{split}
    \end{equation}
    For instance, for  candidate parse (\ref{parse:relbert-1}), $score^{R}(\text{``Recipient"},
    \linebreak[2]
    \text{``John"})$ is computed using Formula (\ref{formula:sbert}):
    $$
    \begin{aligned}
    &\max_{i \in \{1,2\}}\Bigl(\bigl\{sim\bigl(emb_{snt}(def_i(\text{``Recipient"})),emb_{snt}(def_{gpt}(\text{``John"}))\bigl)\bigl\}\Bigl)\\
    &=\max\Bigl(sim\bigl(emb_{snt}(\text{``A person's name"}),emb_{snt}(\text{``A male's name"})\bigl),\\
    &\quad\quad\quad\;\;\; sim\bigl(emb_{snt}(\text{``An organization's name"}),emb_{snt}(\text{``A male's name"})\bigl)\Bigl)\\
    &=\max\big(\{0.721,0.209\}\big) \\
    &=0.721
    \end{aligned}
    $$
    which indicates that, between the two definitions of \texttt{Recipient}, ``\textit{a person's name}" and ``\textit{an organization's name}", ``\textit{John}" as ``\textit{a male's name}" is more similar
    to ``\textit{a person's name}".
    
    Regarding candidate parse (\ref{parse:relbert-2}), $score^{R}(\text{``Money"},\text{``John"})$ is determined using Formula (\ref{formula:sbert}) as follows:
    $$
    \begin{aligned}
    &\max_{i \in \{1\}}\Bigl(\bigl\{sim\bigl(emb_{snt}(def_i(\text{``Money"})),emb_{snt}(def_{gpt}(\text{``John"}))\bigl)\bigl\}\Bigl)\\
    &=\max\Bigl(sim\bigl(emb_{snt}(\text{``Medium of exchange; currency"}),emb_{snt}(\text{``A male's name"})\bigl)\Bigl)\\
    &=sim\bigl(emb_{snt}(\text{``Medium of exchange; currency"}),emb_{snt}(\text{``A male's name"})\bigl)\\
    &=0.082 < 0.721
    \end{aligned}
    $$
    which indicates that ``\textit{John}" is less suited as a filler for the role \texttt{Money} than the role \texttt{Recipient}.

    \item \emph{The computation of} $score^{P}$.
    Similar to Step 2 of the RBD approach, Formula (\ref{formula:geo}) on page 30 is used to calculate the likelihood score for each candidate parse as a whole.
    Candidate parses with low scores are subsequently eliminated.
    In the running example, the lower value of $score^{R}(\text{``Money"},\text{``John"})$ results in a lower overall value of $score^{P}$ for candidate parse (\ref{parse:relbert-2}), so the parse (\ref{parse:relbert-2}) is dropped.
    Ultimately, the test sentence ``\textit{Bob bought a piano for John}" is disambiguated via the parse (\ref{parse:relbert-1}).

\end{enumerate}

Like the RBD approach, SBD does not issue queries to BabelNet, which greatly improves the speed of disambiguation.
Additionally, as mentioned at the end of Section~\ref{sec:rbd-word}, an analog of BabelNet's synset system could be developed for SBD word senses, which would make SBD more comparable to KALM.


%% file: 5_disambiguation/3_eval.tex
\section{Evaluation of Revised Disambiguation}\label{sec:disam-eval}

In this section, we evaluate the revised disambiguation approaches to demonstrate the improvements in accuracy and speed made by RBD and SBD.

\subsection{Settings}

We use the role-filler disambiguation approach outlined in Section~\ref{sec:kalm} as the baseline for our evaluation.
In terms of test datasets, we utilize the ones introduced in Section~\ref{sec:kalmfl-datasets}, with the exception of MetaQA.
This is because the semantics in the MetaQA dataset are relatively shallow compared to other datasets.
Specifically, the dataset can be adequately covered by only three frames, as discussed in Section~\ref{sec:kalmfl-datasets}.
As a result, ambiguous candidate parses are rare in MetaQA, and there is no need to perform disambiguation.

The evaluation of RBD and SBD involves assessing their performance in terms of speed and accuracy.
To measure speed, we conduct five independent runs of each disambiguation method on the same machine\footnote{
    Equipped with Intel(R) Core(TM) i7-2600 CPU @ 3.40 GHz and 32 GB memory. No GPUs installed.
} for each dataset.
We record the time taken for each run, from the start of the disambiguation process to its completion.
The final running time for each method on each dataset was obtained by averaging the results of the five runs.
Additionally, we calculated the average accuracy for each approach--dataset combination from the five runs.
The results of these evaluations are presented in the following section.
Note that in order to ensure fairness, we employ the candidate parses generated by \KALMF as the input for all approaches, including the baseline.
This guarantees that each approach is tested on the same set of candidate frame-based parses.

\subsection{Results}
Our evaluation results are presented in two tables.

Table~\ref{tab:disamb-results-time} compares the runtimes of different approaches on various datasets.
The runtimes (in seconds) for each approach--dataset setting are listed as \textit{Runtime} in the table, and are further divided into two parts.
For the baseline approach, the runtime consists of two components: 1) the time taken for BabelNet to query for role-fillers, and 2) the time taken for BabelNet connectivity search.
For the RBD approach, the runtime also consists of two components: 1) the time taken for RelBERT encoding, and 2) the time taken for GPT generation.
For the SBD approach, the runtime consists of: 1) the time taken for GPT generation, and 2) the time taken for SBERT enconding.
The percentage decrease in runtime compared to the baseline approach is denoted by ``$\downarrow\%$".

\begin{table}[htbp!]
\caption{Runtime Comparisons}
\centering
\vspace{1mm}
\begin{tabular}{ccccc}
\hline\hline
          & \multicolumn{4}{c}{CNLD}      \\
          & & Runtime & & $\downarrow\%$  \\
\hline 
Baseline     & & 4.40 (1.35+3.05) & & --  \\
\hline
RBD      & & 2.06 (0.96+1.10) & & 53\%   \\
SBD      & & \textbf{1.41} (1.15+0.26) & & \textbf{68\%}  \\
\hline\hline
          & \multicolumn{4}{c}{CNLDM}  \\
          & & Runtime & & $\downarrow\%$ \\
\hline
Baseline     & & 4.40 (1.35+3.05) & & --   \\
\hline
RBD      & & 2.08 (0.96+1.12) & & 53\%      \\
SBD      & & \textbf{1.43} (1.16+0.27) & & \textbf{68\%} \\
\hline\hline
          & \multicolumn{4}{c}{NLD}  \\
          & & Runtime & & $\downarrow\%$ \\
\hline
Baseline     & & 4.82 (1.71+3.11) & & --      \\
\hline
RBD      & & 2.19 (1.00+1.19) & & 55\%      \\
SBD      & & \textbf{1.55} (1.28+0.27) & & \textbf{68\%} \\
\hline\hline
\end{tabular}
\label{tab:disamb-results-time}
\end{table}

As shown in Table~\ref{tab:disamb-results-time}, the revised approaches demonstrated significantly faster performance. 
Specifically, RBD reduces the runtime by approximately 55\% compared to the baseline approach, while
SBD approximately triples the speed of the process.

Table~\ref{tab:disamb-results-acc} presents the accuracy for different combinations of settings: the datasets (CNLD, CNLDM, NLD) and the types of accuracies (``F" and ``R" stand for frame-level and role-level accuracies---see
Section~\ref{sec:kalmfl-results}).

In terms of the \textit{synset-level} accuracy metric,
it is important to note that RBD and SBD do not utilize BabelNet synsets or any similar synset-based systems for word sense disambiguation.
As a result, this particular metric cannot be directly applied to RBD and SBD.
However, as discussed in Section~\ref{sec:rbd-word}, fair comparisons at word sense level could be done if a synset-level system is developed for RBD and SBD in the future.
In this evaluation, we therefore limited our attention to frame- and role-level accuracies.


\begin{table}[htbp!]
\caption{Accuracy Comparisons}
\centering
\vspace{1mm}
\begin{tabular}{ccccccc}
\hline\hline
          & \multicolumn{2}{c}{CNLD}                  & \multicolumn{2}{c}{CNLDM}       & \multicolumn{2}{c}{NLD}    \\
          & F             & R                         & F             & R                & F     & R           \\ 
\hline 
Baseline      & 0.99 & 0.99 & 0.99 & 0.99 & 0.99 & 0.98 \\
\hline
RBD      & \textbf{1.00} & \textbf{1.00} & \textbf{1.00} & \textbf{1.00} & \textbf{1.00} & \textbf{0.99} \\
SBD      & \textbf{1.00} & 0.99 & 0.99 & 0.99 & 0.99 & \textbf{0.99} \\
\hline\hline
\end{tabular}
\label{tab:disamb-results-acc}
\end{table}

The accuracy results in Table~\ref{tab:disamb-results-acc} show that our revised role-filler disambiguation approaches achieve superior accuracy compared to the BabelNet-based baseline approach.
For the RBD approach, RelBERT computes semantic scores for each candidate frame and filters out unwanted parses and results in higher frame- and role-level accuracy than the baseline approach.
As for the SBD approach, GPT generates high-quality role-filler definitions first, which gives better frame- and role-level accuracy compared to the baseline approach.

In summary, our two proposed role-filler disambiguation approaches, particularly SBD, have shown significant improvement in speed and achieved correctness close to 100\% at frame- and role-levels.
This will make KALM-based knowledge authoring approaches both
more efficient and accurate.

%% file: 6_conclusion/0_main.tex
\chapter{Conclusion and Future Work}\label{chap:7:conclusion}

The original KALM \cite{gao2018high,gao2018knowledge} was proposed as a solution to the problem of semantic mismatch in knowledge authoring using natural languages, but this solution was limited to CNLs, which is a severe limitation both in expressiveness and human training. 
In this thesis, we first extended KALM to \KALMF,
which is no longer chained by CNL limitations.
The only restriction is that the sentences used for knowledge authoring must be factual, i.e., express factual information as opposed to, say, feelings, allegories, hyperbolas, etc.
We also proposed \KALMR based on \KALMF, which enables authoring of rules and actions after tackling a slew of problems using logic programming techniques including F-logic and Event Calculus.
In addition, we observed that the role-filler disambiguation step in KALM-based approaches can be too time-consuming for interactive users.
To address this issue, we proposed two faster and more accurate solutions that rely on representation learning techniques.
Benchmarking shows that our approach captures the meanings of facts and queries with very high accuracy:
the 0.95 F1 score.
On rules and actions, our approach achieves 100\% accuracy on rule authoring, and 99.34\% accuracy on authoring and reasoning with actions, demonstrating the effectiveness of our approach at capturing knowledge via facts, actions, rules, and queries.
In terms of execution efficiency, our revised role-filler disambiguation
approach
improves the efficiency by a factor of 3
and also leads to slightly better accuracy.


As future work, several things could be done to extend KALM and make it more practical.
For example, GPT-based disambiguation requires an efficient mechanism to replace BabelNet synsets.
Likewise, to make KALM more expressive, we might look into extending factual English to support defeasible reasoning~\cite{WanGKFL09} and other forms of common-sense reasoning.

%% file: 7_appendix/0_main.tex
\appendix

\input{7_appendix/1_prompts}
\input{7_appendix/2_rules}

%% file: 7_appendix/1_prompts.tex
\chapter{GPT Few-Shot Prompts}\label{apdx:prompts}

This appendix presents three GPT prompts for the tasks of frame-based parsing and role-filler definition generation.
The outputs of these prompts are used to evaluate the frame-, role-, and synset-level accuracy of GPT, as discussed in Sections~\ref{sec:kalmfl-compsys} and \ref{sec:kalmfl-results}.

\begin{prompt}
\label{prompt:parsing}
Frame-based parsing prompt for datasets other than MetaQA:
\begin{verbatim}
Given a list of FRAME_TEMPLATES, frame-parse a sentence 
SENTENCE using the provided FRAME_TEMPLATES, output a 
LIST_OF_PARSES that represents the meaning or semantics
of SENTENCE. Ignore the role if there is no appropriate
filler word for that role.
FRAME_TEMPLATES: {FRAME_TEMPLATES}
==============================================================
Example 1:
SENTENCE: "John features the role of Hamlet in a new film."
LIST_OF_PARSES: [p("Performing",[role("Performer","John"),
                                 role("Role","Hamlet"),
                                 role("Performance","film")])]
Example 2:
SENTENCE: "Mary buys a car."
LIST_OF_PARSES: [p("Commerce_buy",[role("Buyer","Mary"),
                                   role("Goods","car")])]
==============================================================
Try:
SENTENCE: "{SENTENCE}"
LIST_OF_PARSES: 
\end{verbatim}

\noindent
where \texttt{\{FRAME\_TEMPLATES\}} and \texttt{\{SENTENCE\}} are placeholders for 50 general-domain frame templates preset by knowledge engineers and the sentence to be parsed, respectively.

\end{prompt}

\begin{prompt}
\label{prompt:parsing2}
Frame-based parsing prompt for MetaQA:
\begin{verbatim}
Given a list of FRAME_TEMPLATES, frame-parse a sentence 
SENTENCE using the provided FRAME_TEMPLATES, output a 
LIST_OF_PARSES that represents the meaning or semantics
of SENTENCE. Ignore the role if there is no appropriate
filler word for that role.
FRAME_TEMPLATES: {FRAME_TEMPLATES}
==============================================================
Example 1:
SENTENCE: "The films directed by the director of [MASK]
           starred who?"
Return:
LIST_OF_PARSES: [p("Movie",[role("Movie_name","film"),
                            role("Actor","who")]),
                 p("Movie",[role("Movie_name","MASK"),
                            role("Director","director")]),
                 p("Movie",[role("Movie_name","film"),
                            role("Director","director")])]

Example 2:
SENTENCE: "The movies that share directors with the movie [MASK]
           were released in which years?"
Return:
LIST_OF_PARSES: [p("Movie",[role("Movie_name","movie"),
                            role("Release_year","which")]),
                 p("Movie",[role("Movie_name","movie"),
                            role("Director","director")]),
                 p("Movie",[role("Movie_name","MASK"),
                            role("Director","director")]),
                 p("Inequality",[role("Entity_1","movie"),
                                 role("movie_mask","MASK")])]

Example 3:
SENTENCE: "The actor [MASK] starred together with who?"
Return:
LIST_OF_PARSES: [p("Cooperation",[role("Actor","who"),
                                  role("Actor","MASK")])]
==============================================================
Try:
SENTENCE: "{SENTENCE}"
LIST_OF_PARSES: 
\end{verbatim}

\noindent
where \texttt{\{FRAME\_TEMPLATES\}} and \texttt{\{SENTENCE\}} are placeholders for 3 MetaQA frame templates preset by knowledge engineers and the sentence to be parsed, respectively.

\end{prompt}

\begin{prompt}
\label{prompt:disamb}
In-context definition generation prompt:
\begin{verbatim}
Given a SENTENCE, and a WORD in the SENTENCE,
generate the WORD's DEFINITION with only one short sentence.
============================================================
Example 1:
SENTENCE: "Kate charters a car."
WORD: "car"
DEFINITION of "car": A motor vehicle with four wheels.
Example 2:
SENTENCE: "Jackson protested against the government."
WORD: "Jackson"
DEFINITION of "Jackson": A male's name
Example 3:
SENTENCE: "Antoinette is staying with friends
           in Paris at present."
WORD: "friend"
DEFINITION of "friend": A person with whom one has a bond
                        of mutual affection and trust.
============================================================
Try:
SENTENCE: "{SENTENCE}"
WORD: "{WORD}"
DEFINITION of "{WORD}":
\end{verbatim}
\noindent
where \texttt{\{SENTENCE\}} and \texttt{\{WORD\}} are placeholders for the sentence and the role-filler, respectively.

\end{prompt}

%% file: 7_appendix/2_rules.tex
\chapter{Complete UTI Guidelines in Factual English}\label{apdx:uti}
\textbf{Background Rules}
\begin{enumerate}
    \item \textbf{If} \textit{\$doctor} administers \textit{\$therapy} for \textit{\$patient}, \textbf{then} \textit{\$patient} undergoes \textit{\$therapy} from \textit{\$doctor}.
    \item \textbf{If} \textit{\$patient} undergoes \textit{\$therapy} from \textit{\$doctor}, \textbf{then} \textit{\$patient}'s \textit{\$therapy} from \textit{\$doctor} is completed, or not completed.
    \item \textbf{If} \textit{\$doctor} performs \textit{\$imaging\_study} for \textit{\$patient}, \textbf{then} \textit{\$patient}'s \textit{\$imaging\_study} from \textit{\$doctor} is completed, or not completed.
\end{enumerate}

\noindent
\textbf{Recommendation 1}
\begin{enumerate}
    \item \textbf{If} \textit{\$doctor}'s \textit{\$patient} is a young child and has an unexplained fever, \textbf{then} \textit{\$doctor} considers UTI for \textit{\$patient}.
\end{enumerate}

\noindent
\textbf{Recommendation 2}
\begin{enumerate}
    \item \textbf{If} \textit{\$doctor}'s \textit{\$patient} is a young child and has an unexplained fever, \textbf{then} \textit{\$doctor} assesses \textit{\$patient}'s degree of toxicity.
    \item \textbf{If} \textit{\$doctor}'s \textit{\$patient} is a young child and has an unexplained fever, \textbf{then} \textit{\$doctor} assesses \textit{\$patient}'s degree of dehydration.
    \item \textbf{If} \textit{\$doctor}'s \textit{\$patient} is a young child and has an unexplained fever, \textbf{then} \textit{\$doctor} assesses \textit{\$patient}'s ability to retain oral intake.
    \item \textbf{If} \textit{\$doctor} assesses \textit{\$patient}'s ability to retain oral intake, \textbf{then} \textit{\$doctor}'s \textit{\$patient} is or is not able to retain oral intake.
\end{enumerate}

\noindent
\textbf{Recommendation 3}
\begin{enumerate}
    \item \textbf{If} \textit{\$doctor}'s \textit{\$patient} is a young child and has an unexplained fever, and \textit{\$doctor}'s \textit{\$patient} is sufficiently ill, \textbf{then} \textit{\$doctor}  analyzes the culture of \textit{\$patient}'s urine specimen obtained by SPA or transurethral catheterization.
\end{enumerate}

\noindent
\textbf{Recommendation 4}
\begin{enumerate}
    \item \textbf{If} \textit{\$doctor}'s \textit{\$patient} is a young child and has an unexplained fever, and \textit{\$doctor}'s \textit{\$patient} is not sufficiently ill, \textbf{then} \textit{\$doctor}  analyzes the culture of \textit{\$patient}'s urine specimen obtained by SPA, transurethral catheterization, or a convenient method.
    \item \textbf{If} \textit{\$doctor}'s \textit{\$patient} is a young child and has an unexplained fever, \textit{\$doctor}'s \textit{\$patient} is not sufficiently ill, \textit{\$doctor} analyzes the culture of \textit{\$patient}'s urine specimen obtained by a convenient method, and the analysis of \textit{\$patient}'s culture of a urine specimen suggests UTI, \textbf{then} \textit{\$doctor}  analyzes \textit{\$patient}'s culture of a urine specimen obtained by SPA or transurethral catheterization.
    \item \textbf{If} \textit{\$doctor} analyzes the culture of \textit{\$patient}'s urine specimen obtained by SPA or transurethral catheterization, \textbf{then} \textit{\$doctor}'s analysis of \textit{\$patient}'s culture confirms UTI or excludes UTI.
    \item \textbf{If} \textit{\$doctor} analyzes the culture of \textit{\$patient}'s urine specimen obtained by a convenient method, \textbf{then} \textit{\$doctor}'s analysis of \textit{\$patient}'s culture suggests UTI or does not suggest UTI.
    \item \textbf{If} \textit{\$doctor}'s analysis of the culture of \textit{\$patient}'s urine specimen confirms UTI, \textbf{then} \textit{\$doctor}'s \textit{\$patient} has UTI.
    \item \textbf{If} \textit{\$doctor}'s analysis of the culture of \textit{\$patient}'s urine specimen excludes UTI, \textbf{then} \textit{\$doctor}'s \textit{\$patient} does not have UTI.
\end{enumerate}

\noindent
\textbf{Recommendation 5} is integrated into 3 and 4.

\noindent
\textbf{Recommendation 6}
\begin{enumerate}
    \item \textbf{If} \textit{\$doctor}'s \textit{\$patient} is a young child and has an unexplained fever, and \textit{\$doctor}'s \textit{\$patient} is toxic, \textbf{then} \textit{\$doctor} administers an antimicrobial therapy for \textit{\$patient}.
    \item \textbf{If} \textit{\$doctor}'s \textit{\$patient} is a young child and has an unexplained fever, and \textit{\$doctor}'s \textit{\$patient} is toxic, \textbf{then} \textit{\$doctor} considers hospitalization for \textit{\$patient}.
    \item \textbf{If} \textit{\$doctor}'s \textit{\$patient} is a young child and has an unexplained fever, and \textit{\$doctor}'s \textit{\$patient} is dehydrated, \textbf{then} \textit{\$doctor} administers an antimicrobial therapy for \textit{\$patient}.
    \item \textbf{If} \textit{\$doctor}'s \textit{\$patient} is a young child and has an unexplained fever, and \textit{\$doctor}'s \textit{\$patient} is dehydrated, \textbf{then} \textit{\$doctor} considers hospitalization for \textit{\$patient}.
    \item \textbf{If} \textit{\$doctor}'s \textit{\$patient} is a young child and has an unexplained fever, and \textit{\$doctor}'s \textit{\$patient} is not able to retain oral intake, \textbf{then} \textit{\$doctor} administers an antimicrobial therapy for \textit{\$patient}.
    \item \textbf{If} \textit{\$doctor}'s \textit{\$patient} is a young child and has an unexplained fever, and \textit{\$doctor}'s \textit{\$patient} is not able to retain oral intake, \textbf{then} \textit{\$doctor} considers hospitalization for \textit{\$patient}.
\end{enumerate}

\noindent
\textbf{Recommendation 7}
\begin{enumerate}
    \item \textbf{If} \textit{\$doctor}'s \textit{\$patient} is a young child, and \textit{\$doctor}'s analysis of the culture of \textit{\$patient}'s urine specimen confirms UTI, \textbf{then} \textit{\$doctor}  administers a parenteral or oral antimicrobial therapy for \textit{\$patient}.
\end{enumerate}

\noindent
\textbf{Recommendation 8}
\begin{enumerate}
    \item \textbf{If} \textit{\$doctor}'s \textit{\$patient} is a young child and has UTI, \textit{\$patient} undergoes an antimicrobial therapy from \textit{\$doctor} for 2 days, and \textit{\$doctor}'s \textit{\$patient} does not show the expected response of the antimicrobial therapy, \textbf{then} \textit{\$doctor}  reevaluates \textit{\$patient} and analyze the culture of \textit{\$patient}'s second urine specimen.
    \item \textbf{If} \textit{\$doctor}'s \textit{\$patient} is a young child and has UTI, and \textit{\$patient} undergoes an antimicrobial therapy from \textit{\$doctor} for 2 days, \textbf{then} \textit{\$doctor}'s \textit{\$patient} shows or does not show the expected response of the antimicrobial therapy.
\end{enumerate}

\noindent
\textbf{Recommendation 9}
\begin{enumerate}
    \item \textbf{If} \textit{\$doctor}'s \textit{\$patient} is a young child and has UTI, \textbf{then} \textit{\$doctor} administers an oral antimicrobial therapy that lasts at least 7 days for \textit{\$patient}.
    \item \textbf{If} \textit{\$doctor}'s \textit{\$patient} is a young child and has UTI, \textbf{then} \textit{\$doctor} administers an oral antimicrobial therapy that lasts at most 14 days for \textit{\$patient}.
\end{enumerate}

\noindent
\textbf{Recommendation 10}
\begin{enumerate}
    \item \textbf{If} \textit{\$doctor}'s \textit{\$patient} is a young child and has UTI, the antimicrobial therapy of \textit{\$doctor}'s \textit{\$patient} is completed, and the imaging study of \textit{\$doctor}'s \textit{\$patient} is not completed, \textbf{then} \textit{\$doctor} administers \textit{\$patient} a therapeutically or prophylactically dosed antimicrobial.
\end{enumerate}

\noindent
\textbf{Recommendation 11}
\begin{enumerate}
    \item \textbf{If} \textit{\$doctor}'s \textit{\$patient} is a young child and has UTI, \textit{\$patient} undergoes an antimicrobial therapy for 2 days from \textit{\$doctor}, and \textit{\$doctor}'s \textit{\$patient} does not show the expected response of the antimicrobial therapy, \textbf{then} \textit{\$doctor} performs ultrasonography promptly for \textit{\$patient}.
    \item \textbf{If} \textit{\$doctor}'s \textit{\$patient} is a young child and has UTI, \textit{\$patient} undergoes an antimicrobial therapy for 2 days from \textit{\$doctor}, and \textit{\$doctor}'s \textit{\$patient} does not show the expected response of the antimicrobial therapy, \textbf{then} \textit{\$doctor} performs VCUG or RNC for \textit{\$patient}.
    \item \textbf{If} \textit{\$doctor}'s \textit{\$patient} is a young child and has UTI, \textit{\$patient} undergoes an antimicrobial therapy for 2 days from \textit{\$doctor}, and \textit{\$doctor}'s \textit{\$patient} shows the expected response of the antimicrobial therapy, \textbf{then} \textit{\$doctor} performs VCUG or RNC for \textit{\$patient}.
    \item \textbf{If} \textit{\$doctor}'s \textit{\$patient} is a young child and has UTI, \textit{\$patient} undergoes an antimicrobial therapy for 2 days from \textit{\$doctor}, and \textit{\$doctor}'s \textit{\$patient} shows the expected response of the antimicrobial therapy, \textbf{then} \textit{\$doctor} performs VCUG or RNC for \textit{\$patient}.
\end{enumerate}

%% file: main.bbl
\begin{thebibliography}{10}

\bibitem{arias2022modeling}
Joaqu{\'\i}n Arias, Manuel Carro, Zhuo Chen, and Gopal Gupta.
\newblock Modeling and reasoning in event calculus using goal-directed constraint answer set programming.
\newblock {\em Theory and Practice of Logic Programming}, 22(1):51--80, 2022.

\bibitem{arnab2021vivit}
Anurag Arnab, Mostafa Dehghani, Georg Heigold, Chen Sun, Mario Lu{\v{c}}i{\'c}, and Cordelia Schmid.
\newblock Vivit: A video vision transformer.
\newblock In {\em Proceedings of the IEEE/CVF international conference on computer vision}, pages 6836--6846, 2021.

\bibitem{baker1998berkeley}
Collin~F Baker, Charles~J Fillmore, and John~B Lowe.
\newblock The berkeley framenet project.
\newblock In {\em 36th Annual Meeting of the Association for Computational Linguistics and 17th International Conference on Computational Linguistics}, pages 86--90. COLING, 1998.

\bibitem{baker2003framenet}
Collin~F Baker and Hiroaki Sato.
\newblock The framenet data and software.
\newblock In {\em The Companion Volume to the Proceedings of 41st Annual Meeting of the Association for Computational Linguistics}, pages 161--164, 2003.

\bibitem{bojanowski2017enriching}
Piotr Bojanowski, Edouard Grave, Armand Joulin, and Tomas Mikolov.
\newblock Enriching word vectors with subword information.
\newblock {\em Transactions of the association for computational linguistics}, 5:135--146, 2017.

\bibitem{borca2022provable}
Giorgian Borca-Tasciuc, Xingzhi Guo, Stanley Bak, and Steven Skiena.
\newblock Provable fairness for neural network models using formal verification.
\newblock {\em arXiv preprint arXiv:2212.08578}, 2022.

\bibitem{brown2020language}
Tom Brown, Benjamin Mann, Nick Ryder, Melanie Subbiah, Jared~D Kaplan, Prafulla Dhariwal, Arvind Neelakantan, Pranav Shyam, Girish Sastry, Amanda Askell, et~al.
\newblock Language models are few-shot learners.
\newblock {\em Advances in neural information processing systems}, 33:1877--1901, 2020.

\bibitem{carreira2017quo}
Joao Carreira and Andrew Zisserman.
\newblock Quo vadis, action recognition? a new model and the kinetics dataset.
\newblock In {\em proceedings of the IEEE Conference on Computer Vision and Pattern Recognition}, pages 6299--6308, 2017.

\bibitem{chen2023accelerating}
Zhen Chen, Xingzhi Guo, Baojian Zhou, Deqing Yang, and Steven Skiena.
\newblock Accelerating personalized pagerank vector computation.
\newblock {\em arXiv preprint arXiv:2306.02102}, 2023.

\bibitem{chiang2014real}
Shu-Yin Chiang, Xingzhi Guo, and Hsien-Wen Hu.
\newblock Real time self-localization of omni-vision robot by pattern match system.
\newblock In {\em 2014 International Conference on Advanced Robotics and Intelligent Systems (ARIS)}, pages 46--50. IEEE, 2014.

\bibitem{committee1999practice}
Subcommittee on Urinary Tract~Infection Committee~on Quality~Improvement.
\newblock Practice parameter: the diagnosis, treatment, and evaluation of the initial urinary tract infection in febrile infants and young children.
\newblock {\em Pediatrics}, 103(4):843--852, 1999.

\bibitem{das2014frame}
Dipanjan Das, Desai Chen, Andr{\'e}~FT Martins, Nathan Schneider, and Noah~A Smith.
\newblock Frame-semantic parsing.
\newblock {\em Computational linguistics}, 40(1):9--56, 2014.

\bibitem{devlin2018bert}
Jacob Devlin, Ming-Wei Chang, Kenton Lee, and Kristina Toutanova.
\newblock Bert: Pre-training of deep bidirectional transformers for language understanding.
\newblock {\em arXiv preprint arXiv:1810.04805}, 2018.

\bibitem{dhingra2020differentiable}
Bhuwan Dhingra, Manzil Zaheer, Vidhisha Balachandran, Graham Neubig, Ruslan Salakhutdinov, and William~W Cohen.
\newblock Differentiable reasoning over a virtual knowledge base.
\newblock {\em arXiv preprint arXiv:2002.10640}, 1, 2020.

\bibitem{dosovitskiy2020image}
Alexey Dosovitskiy, Lucas Beyer, Alexander Kolesnikov, Dirk Weissenborn, Xiaohua Zhai, Thomas Unterthiner, Mostafa Dehghani, Matthias Minderer, Georg Heigold, Sylvain Gelly, et~al.
\newblock An image is worth 16x16 words: Transformers for image recognition at scale.
\newblock {\em arXiv preprint arXiv:2010.11929}, 2020.

\bibitem{dozat2016deep}
Timothy Dozat and Christopher~D Manning.
\newblock Deep biaffine attention for neural dependency parsing.
\newblock {\em arXiv preprint arXiv:1611.01734}, 2016.

\bibitem{feichtenhofer2019slowfast}
Christoph Feichtenhofer, Haoqi Fan, Jitendra Malik, and Kaiming He.
\newblock Slowfast networks for video recognition.
\newblock In {\em Proceedings of the IEEE/CVF international conference on computer vision}, pages 6202--6211, 2019.

\bibitem{fillmore2006frame}
Charles~J Fillmore et~al.
\newblock Frame semantics.
\newblock {\em Cognitive linguistics: Basic readings}, 34:373--400, 2006.

\bibitem{fuchs2005extended}
Norbert~E Fuchs, Stefan Hoefler, Kaarel Kaljurand, Gerold Schneider, and Uta Schwertel.
\newblock Extended discourse representation structures in attempto controlled english.
\newblock {\em ifi Technical Reports}, 2005.

\bibitem{fuchs1996attempto}
Norbert~E Fuchs and Rolf Schwitter.
\newblock Attempto controlled english (ace).
\newblock {\em arXiv preprint cmp-lg/9603003}, 1996.

\bibitem{gao2018high}
Tiantian Gao, Paul Fodor, and Michael Kifer.
\newblock High accuracy question answering via hybrid controlled natural language.
\newblock In {\em 2018 IEEE/WIC/ACM International Conference on Web Intelligence (WI)}, pages 17--24. IEEE, 2018.

\bibitem{gao2018knowledge}
Tiantian Gao, Paul Fodor, and Michael Kifer.
\newblock Knowledge authoring for rule-based reasoning.
\newblock In {\em OTM Confederated International Conferences" On the Move to Meaningful Internet Systems"}, pages 461--480. Springer, 2018.

\bibitem{gao2019querying}
Tiantian Gao, Paul Fodor, and Michael Kifer.
\newblock Querying knowledge via multi-hop english questions.
\newblock {\em Theory and Practice of Logic Programming}, 19(5-6):636--653, 2019.

\bibitem{gebser2019multi}
Martin Gebser, Roland Kaminski, Benjamin Kaufmann, and Torsten Schaub.
\newblock Multi-shot asp solving with clingo.
\newblock {\em Theory and Practice of Logic Programming}, 19(1):27--82, 2019.

\bibitem{gelfond1988stable}
Michael Gelfond and Vladimir Lifschitz.
\newblock The stable model semantics for logic programming.
\newblock In {\em ICLP/SLP}, volume~88, pages 1070--1080. Cambridge, MA, 1988.

\bibitem{gelfond1991classical}
Michael Gelfond and Vladimir Lifschitz.
\newblock Classical negation in logic programs and disjunctive databases.
\newblock {\em New generation computing}, 9:365--385, 1991.

\bibitem{gillespie2020improving}
Kellen Gillespie, Ioannis~C Konstantakopoulos, Xingzhi Guo, Vishal~Thanvantri Vasudevan, and Abhinav Sethy.
\newblock Improving device directedness classification of utterances with semantic lexical features.
\newblock In {\em ICASSP 2020-2020 IEEE International Conference on Acoustics, Speech and Signal Processing (ICASSP)}, pages 7859--7863. IEEE, 2020.

\bibitem{guo2023analyzing}
Xingzhi Guo.
\newblock {\em Analyzing the Evolution of Graphs and Texts}.
\newblock PhD thesis, State University of New York at Stony Brook, 2023.

\bibitem{guo2024evolution}
Xingzhi Guo, Dakota Handzlik, Jason~J Jones, and Steven~S Skiena.
\newblock The evolution of occupational identity in twitter biographies.
\newblock In {\em Proceedings of the International AAAI Conference on Web and Social Media}, volume~18, pages 502--514, 2024.

\bibitem{guo2019inferring}
Xingzhi Guo, Yu-Cian Huang, Edwinn Gamborino, Shih-Huan Tseng, Li-Chen Fu, and Su-Ling Yeh.
\newblock Inferring human feelings and desires for human-robot trust promotion.
\newblock In {\em International Conference on Human-Computer Interaction}, pages 365--375. Springer, 2019.

\bibitem{guo2022verba}
Xingzhi Guo, Brian Kondracki, Nick Nikiforakis, and Steven Skiena.
\newblock Verba volant, scripta volant: Understanding post-publication title changes in news outlets.
\newblock In {\em Proceedings of the ACM Web Conference 2022}, pages 588--598, 2022.

\bibitem{guo2022hierarchies}
Xingzhi Guo and Steven Skiena.
\newblock Hierarchies over vector space: Orienting word and graph embeddings.
\newblock {\em arXiv preprint arXiv:2211.01430}, 2022.

\bibitem{guo2021subset}
Xingzhi Guo, Baojian Zhou, and Steven Skiena.
\newblock Subset node representation learning over large dynamic graphs.
\newblock In {\em Proceedings of the 27th ACM SIGKDD Conference on Knowledge Discovery \& Data Mining}, pages 516--526, 2021.

\bibitem{guo2022subset}
Xingzhi Guo, Baojian Zhou, and Steven Skiena.
\newblock Subset node anomaly tracking over large dynamic graphs.
\newblock In {\em Proceedings of the 28th ACM SIGKDD Conference on Knowledge Discovery and Data Mining}, pages 475--485, 2022.

\bibitem{he2017mask}
Kaiming He, Georgia Gkioxari, Piotr Doll{\'a}r, and Ross Girshick.
\newblock Mask r-cnn.
\newblock In {\em Proceedings of the IEEE international conference on computer vision}, pages 2961--2969, 2017.

\bibitem{he2016deep}
Kaiming He, Xiangyu Zhang, Shaoqing Ren, and Jian Sun.
\newblock Deep residual learning for image recognition.
\newblock In {\em Proceedings of the IEEE conference on computer vision and pattern recognition}, pages 770--778, 2016.

\bibitem{kamp1993discourse}
Hans Kamp and Uwe Reyle.
\newblock From discourse to logic: Introduction to modeltheoretic semantics of natural language, formal logic and discourse representation, 1993.

\bibitem{katzouris2015incremental}
Nikos Katzouris, Alexander Artikis, and Georgios Paliouras.
\newblock Incremental learning of event definitions with inductive logic programming.
\newblock {\em Machine Learning}, 100(2-3):555--585, 2015.

\bibitem{flogic-95}
M.~Kifer, G.~Lausen, and J.~Wu.
\newblock Logical foundations of object-oriented and frame-based languages.
\newblock {\em Journal of the ACM}, 42:741--843, July 1995.

\bibitem{kifer1989f}
Michael Kifer and Georg Lausen.
\newblock F-logic: a higher-order language for reasoning about objects, inheritance, and scheme.
\newblock In {\em Proceedings of the 1989 ACM SIGMOD international conference on Management of data}, pages 134--146, 1989.

\bibitem{kowalski1989logic}
Robert Kowalski and Marek Sergot.
\newblock A logic-based calculus of events.
\newblock In {\em Foundations of knowledge base management}, pages 23--55. Springer, 1989.

\bibitem{krizhevsky2012imagenet}
Alex Krizhevsky, Ilya Sutskever, and Geoffrey~E Hinton.
\newblock Imagenet classification with deep convolutional neural networks.
\newblock {\em Advances in neural information processing systems}, 25, 2012.

\bibitem{le2020self}
Hung Le, Truyen Tran, and Svetha Venkatesh.
\newblock Self-attentive associative memory.
\newblock In {\em International Conference on Machine Learning}, pages 5682--5691. PMLR, 2020.

\bibitem{leone2006dlv}
Nicola Leone, Gerald Pfeifer, Wolfgang Faber, Thomas Eiter, Georg Gottlob, Simona Perri, and Francesco Scarcello.
\newblock The dlv system for knowledge representation and reasoning.
\newblock {\em ACM Transactions on Computational Logic (TOCL)}, 7(3):499--562, 2006.

\bibitem{lin2023comet}
Zhuoyi Lin, Lei Feng, Xingzhi Guo, Yu~Zhang, Rui Yin, Chee~Keong Kwoh, and Chi Xu.
\newblock Comet: Convolutional dimension interaction for collaborative filtering.
\newblock {\em ACM Transactions on Intelligent Systems and Technology}, 14(4):1--18, 2023.

\bibitem{mikolov2013efficient}
Tomas Mikolov, Kai Chen, Greg Corrado, and Jeffrey Dean.
\newblock Efficient estimation of word representations in vector space.
\newblock {\em arXiv preprint arXiv:1301.3781}, 2013.

\bibitem{mikolov2013distributed}
Tomas Mikolov, Ilya Sutskever, Kai Chen, Greg~S Corrado, and Jeff Dean.
\newblock Distributed representations of words and phrases and their compositionality.
\newblock {\em Advances in neural information processing systems}, 26, 2013.

\bibitem{miller1995wordnet}
George~A Miller.
\newblock Wordnet: a lexical database for english.
\newblock {\em Communications of the ACM}, 38(11):39--41, 1995.

\bibitem{mitra2016addressing}
Arindam Mitra and Chitta Baral.
\newblock Addressing a question answering challenge by combining statistical methods with inductive rule learning and reasoning.
\newblock In {\em AAAI}, 2016.

\bibitem{mueller2004event}
Erik~T Mueller.
\newblock Event calculus reasoning through satisfiability.
\newblock {\em Journal of Logic and Computation}, 14(5):703--730, 2004.

\bibitem{navigli2021ten}
Roberto Navigli, Michele Bevilacqua, Simone Conia, Dario Montagnini, and Francesco Cecconi.
\newblock Ten years of babelnet: A survey.
\newblock In {\em IJCAI}, pages 4559--4567, 2021.

\bibitem{navigli2012babelnet}
Roberto Navigli and Simone~Paolo Ponzetto.
\newblock Babelnet: The automatic construction, evaluation and application of a wide-coverage multilingual semantic network.
\newblock {\em Artificial Intelligence}, 193:217--250, 2012.

\bibitem{pennington2014glove}
Jeffrey Pennington, Richard Socher, and Christopher~D Manning.
\newblock Glove: Global vectors for word representation.
\newblock In {\em Proceedings of the 2014 conference on empirical methods in natural language processing (EMNLP)}, pages 1532--1543, 2014.

\bibitem{perozzi2014deepwalk}
Bryan Perozzi, Rami Al-Rfou, and Steven Skiena.
\newblock Deepwalk: Online learning of social representations.
\newblock In {\em Proceedings of the 20th ACM SIGKDD international conference on Knowledge discovery and data mining}, pages 701--710, 2014.

\bibitem{peters2018deep}
Matthew~E Peters, Mark Neumann, Mohit Iyyer, Matt Gardner, Christopher Clark, Kenton Lee, and Luke Zettlemoyer.
\newblock Deep contextualized word representations.
\newblock In {\em Proceedings of NAACL-HLT}, pages 2227--2237, 2018.

\bibitem{qi2020stanza}
Peng Qi, Yuhao Zhang, Yuhui Zhang, Jason Bolton, and Christopher~D Manning.
\newblock Stanza: A python natural language processing toolkit for many human languages.
\newblock {\em arXiv preprint arXiv:2003.07082}, 2020.

\bibitem{radford2018improving}
Alec Radford, Karthik Narasimhan, Tim Salimans, Ilya Sutskever, et~al.
\newblock Improving language understanding by generative pre-training.
\newblock {\em OpenAI}, 2018.

\bibitem{radford2019language}
Alec Radford, Jeffrey Wu, Rewon Child, David Luan, Dario Amodei, Ilya Sutskever, et~al.
\newblock Language models are unsupervised multitask learners.
\newblock {\em OpenAI blog}, 1(8):9, 2019.

\bibitem{reimers2019sentence}
Nils Reimers and Iryna Gurevych.
\newblock Sentence-bert: Sentence embeddings using siamese bert-networks.
\newblock {\em arXiv preprint arXiv:1908.10084}, 2019.

\bibitem{ringgaard2017sling}
Michael Ringgaard, Rahul Gupta, and Fernando~CN Pereira.
\newblock Sling: A framework for frame semantic parsing.
\newblock {\em arXiv preprint arXiv:1710.07032}, 2017.

\bibitem{rouces2015framebase}
Jacobo Rouces, Gerard De~Melo, and Katja Hose.
\newblock Framebase: Representing n-ary relations using semantic frames.
\newblock In {\em The Semantic Web. Latest Advances and New Domains: 12th European Semantic Web Conference, ESWC 2015, Portoroz, Slovenia, May 31--June 4, 2015. Proceedings 12}, pages 505--521. Springer, 2015.

\bibitem{sadri1987three}
Fariba Sadri.
\newblock Three recent approaches to temporal reasoning.
\newblock In {\em Temporal logics and their applications}, pages 121--168. Academic Press Professional, 1987.

\bibitem{sadri1995variants}
Fariba Sadri and Robert~A Kowalski.
\newblock Variants of the event calculus.
\newblock In {\em ICLP}, pages 67--81, 1995.

\bibitem{sagonas1994xsb}
Konstantinos Sagonas, Terrance Swift, and David~S Warren.
\newblock Xsb as an efficient deductive database engine.
\newblock {\em ACM SIGMOD Record}, 23(2):442--453, 1994.

\bibitem{shiffman2009writing}
Richard~N Shiffman, George Michel, Michael Krauthammer, Norbert~E Fuchs, Kaarel Kaljurand, and Tobias Kuhn.
\newblock Writing clinical practice guidelines in controlled natural language.
\newblock In {\em International Workshop on Controlled Natural Language}, pages 265--280. Springer, 2009.

\bibitem{sultan2022low}
Syed~Fahad Sultan, Xingzhi Guo, and Steven Skiena.
\newblock Low-dimensional genotype embeddings for predictive models.
\newblock In {\em Proceedings of the 13th ACM International Conference on Bioinformatics, Computational Biology and Health Informatics}, pages 1--4, 2022.

\bibitem{swayamdipta2017frame}
Swabha Swayamdipta, Sam Thomson, Chris Dyer, and Noah~A Smith.
\newblock Frame-semantic parsing with softmax-margin segmental rnns and a syntactic scaffold.
\newblock {\em arXiv preprint arXiv:1706.09528}, 2017.

\bibitem{swift2012xsb}
Terrance Swift and David~S Warren.
\newblock Xsb: Extending prolog with tabled logic programming.
\newblock {\em Theory and Practice of Logic Programming}, 12(1-2):157--187, 2012.

\bibitem{tran2015learning}
Du~Tran, Lubomir Bourdev, Rob Fergus, Lorenzo Torresani, and Manohar Paluri.
\newblock Learning spatiotemporal features with 3d convolutional networks.
\newblock In {\em Proceedings of the IEEE international conference on computer vision}, pages 4489--4497, 2015.

\bibitem{ushio2021distilling}
Asahi Ushio, Jose Camacho-Collados, and Steven Schockaert.
\newblock Distilling relation embeddings from pre-trained language models.
\newblock {\em arXiv preprint arXiv:2110.15705}, 2021.

\bibitem{WanGKFL09}
Hui Wan, Benjamin~N. Grosof, Michael Kifer, Paul Fodor, and Senlin Liang.
\newblock Logic programming with defaults and argumentation theories.
\newblock In Patricia~M. Hill and David~Scott Warren, editors, {\em Logic Programming, 25th International Conference, {ICLP} 2009, Pasadena, CA, USA, July 14-17, 2009. Proceedings}, volume 5649 of {\em Lecture Notes in Computer Science}, pages 432--448. Springer, 2009.

\bibitem{wang2022knowledge}
Yuheng Wang, Giorgian Borca-Tasciuc, Nikhil Goel, Paul Fodor, and Michael Kifer.
\newblock Knowledge authoring with factual english.
\newblock {\em arXiv preprint arXiv:2208.03094}, 2022.

\bibitem{wang2017time}
Zhiguang Wang, Weizhong Yan, and Tim Oates.
\newblock Time series classification from scratch with deep neural networks: A strong baseline.
\newblock In {\em 2017 International joint conference on neural networks (IJCNN)}, pages 1578--1585. IEEE, 2017.

\bibitem{weston2015towards}
Jason Weston, Antoine Bordes, Sumit Chopra, Alexander~M Rush, Bart van Merri{\"e}nboer, Armand Joulin, and Tomas Mikolov.
\newblock Towards ai-complete question answering: A set of prerequisite toy tasks.
\newblock {\em arXiv preprint arXiv:1502.05698}, 2015.

\bibitem{wu2018learning}
Benjamin Wu, Alessandra Russo, Mark Law, and Katsumi Inoue.
\newblock Learning commonsense knowledge through interactive dialogue.
\newblock In {\em Technical Communications of the 34th International Conference on Logic Programming (ICLP 2018)}. Schloss Dagstuhl-Leibniz-Zentrum fuer Informatik, 2018.

\bibitem{zhang2023subanom}
Chi Zhang, Wenkai Xiang, Xingzhi Guo, Baojian Zhou, and Deqing Yang.
\newblock Subanom: Efficient subgraph anomaly detection framework over dynamic graphs.
\newblock In {\em 2023 IEEE International Conference on Data Mining Workshops (ICDMW)}, pages 1178--1185. IEEE, 2023.

\bibitem{zhang2021dynehr}
Lida Zhang, Xiaohan Chen, Tianlong Chen, Zhangyang Wang, and Bobak~J Mortazavi.
\newblock Dynehr: Dynamic adaptation of models with data heterogeneity in electronic health records.
\newblock In {\em 2021 IEEE EMBS International Conference on Biomedical and Health Informatics (BHI)}, pages 1--4. IEEE, 2021.

\bibitem{zhang2020developing}
Lida Zhang, Nathan~C Hurley, Bassem Ibrahim, Erica Spatz, Harlan~M Krumholz, Roozbeh Jafari, and Mortazavi~J Bobak.
\newblock Developing personalized models of blood pressure estimation from wearable sensors data using minimally-trained domain adversarial neural networks.
\newblock In {\em Machine Learning for Healthcare Conference}, pages 97--120. PMLR, 2020.

\bibitem{zhang2018measuring}
Lida Zhang, Zachary King, Begum Egilmez, Jonathan Reeder, Roozbeh Ghaffari, John Rogers, Kristen Rosen, Michael Bass, Judith Moskowitz, Darius Tandon, et~al.
\newblock Measuring fine-grained heart-rate using a flexible wearable sensor in the presence of noise.
\newblock In {\em 2018 IEEE 15th International Conference on Wearable and Implantable Body Sensor Networks (BSN)}, pages 160--164. IEEE, 2018.

\bibitem{zhang2023semi}
Lida Zhang and Bobak~J Mortazavi.
\newblock Semi-supervised meta-learning for multi-source heterogeneity in time-series data.
\newblock {\em Machine}, 2023.

\bibitem{zhang2017variational}
Yuyu Zhang, Hanjun Dai, Zornitsa Kozareva, Alexander~J Smola, and Le~Song.
\newblock Variational reasoning for question answering with knowledge graph.
\newblock In {\em Thirty-second AAAI conference on artificial intelligence}, pages 6069--6076, 2018.

\bibitem{zhou2024iterative}
Baojian Zhou, Yifan Sun, Reza~Babanezhad Harikandeh, Xingzhi Guo, Deqing Yang, and Yanghua Xiao.
\newblock Iterative methods via locally evolving set process.
\newblock {\em arXiv preprint arXiv:2410.15020}, 2024.

\end{thebibliography}
